\documentclass[secnumarabic,amssymb, nobibnotes, aps, prd]{revtex4-2}
\usepackage{algorithmic}
\usepackage{algorithm}
\usepackage{amsmath}
\usepackage{txfonts}
\usepackage{color}
\usepackage{graphicx}  
\usepackage{here}
\usepackage{braket}
\usepackage{bm}
\usepackage[OT1]{fontenc} 
\usepackage[version=3]{mhchem}

\begin{document}
\title{Sequential Experimental Design for Spectral Measurement:\\ Active Learning Using a Parametric Model}

\author{Tomohiro Nabika$^1$, Kenji Nagata$^2$, Shun Katakami$^1$, Masaichiro Mizumaki$^3$,\\and Masato Okada$^1$}
\affiliation{$^1$Graduate School of Frontier Sciences, The University of Tokyo, Kashiwa, Chiba 277-8561, Japan, \\
$^2$Research and Services Division of Materials Data and Integrated System, National Institute for Materials Science,Tsukuba, Ibaraki 305-0047, Japan,\\
$^3$Faculty of Science, Course for Physical Sciences, Kumamoto University, Kumamoto, Kumamoto 860-8555, Japan.} 

\begin{abstract}
  In this study, we demonstrate a sequential experimental design for spectral measurements by active learning using parametric models as predictors. In spectral measurements, it is necessary to reduce the measurement time because of sample fragility and high energy costs. To improve the efficiency of experiments, sequential experimental designs are proposed, in which the subsequent measurement is designed by active learning using the data obtained before the measurement. Conventionally, parametric models are employed in data analysis; when employed for active learning, they are expected to afford a sequential experimental design that improves the accuracy of data analysis. However, due to the complexity of the formulas, a sequential experimental design using general parametric models has not been realized. Therefore, we applied Bayesian inference-based data analysis using the exchange Monte Carlo method to realize a sequential experimental design with general parametric models. In this study, we evaluated the effectiveness of the proposed method by applying it to Bayesian spectral deconvolution and Bayesian Hamiltonian selection in X-ray photoelectron spectroscopy. Using numerical experiments with artificial data, we demonstrated that the proposed method improves the accuracy of model selection and parameter estimation while reducing the measurement time compared with the results achieved without active learning or with active learning using the Gaussian process regression.
\end{abstract}
\maketitle
\section{Introduction}
In spectral measurements employed for the analysis of various physical properties \cite{IR, XAS}, a large volume of data is typically measured, and sufficient time is required to acquire a low-noise spectrum for high-precision analysis.
However, long measurement times are frequently not feasible because of the fragility of the measured samples \cite{Storp1985} and the associated experimental costs. 
Therefore, it is necessary to reduce the measurement time using high-throughput experiments. \par
\begin{figure}[h]
    \centering
    \includegraphics*[width = 8.0cm]{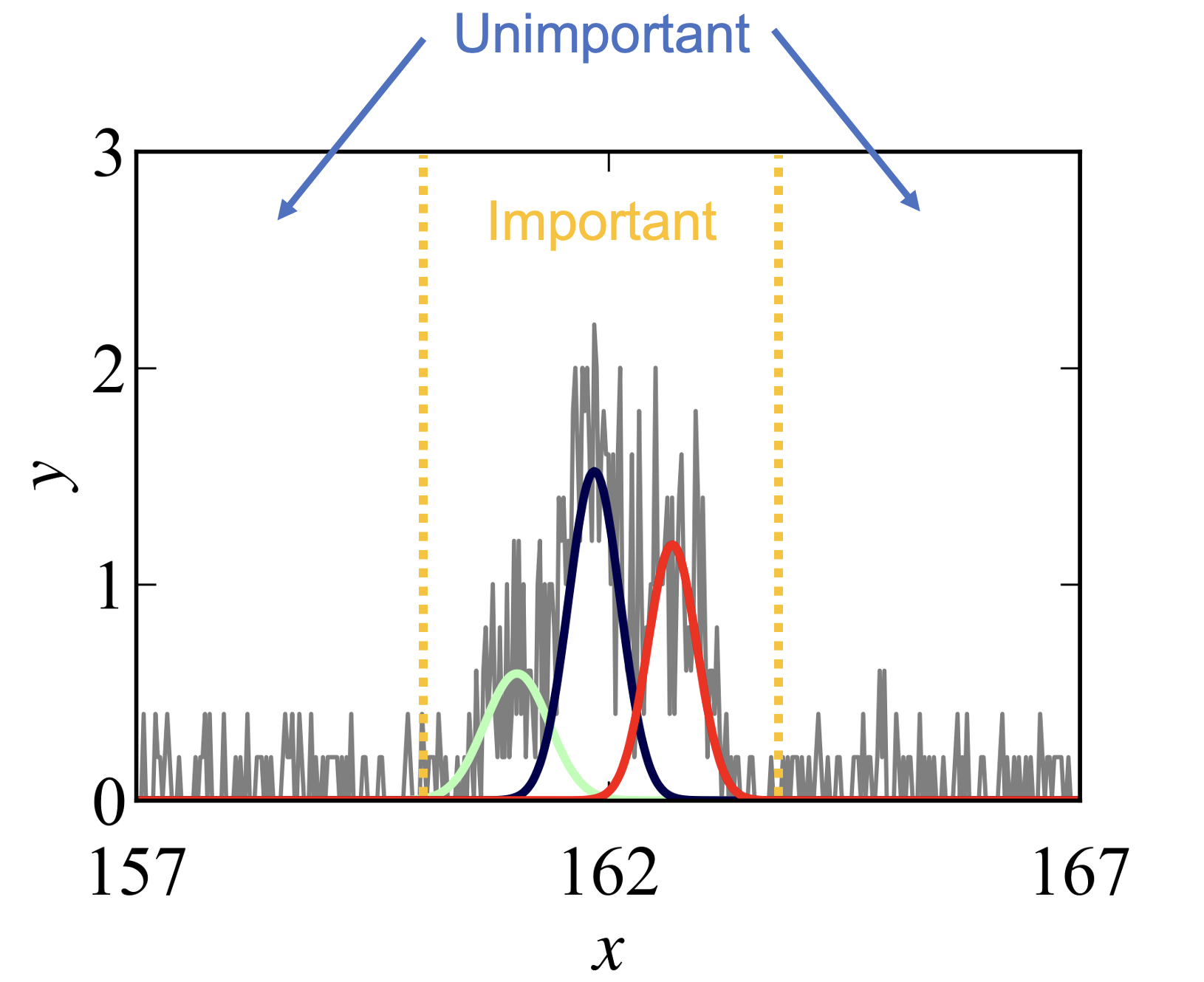}
    \caption{Example of spectral deconvolution. The grey line represents data points and the green, blue and red lines represent the three peak functions. When estimating the parameter of each spectrum and selecting the number of peaks, the area around the center surrounded by the dotted line is more important.}
    \label{spectral_deconvolution}
\end{figure}

One method for achieving this involves focusing on specific measurement points, as the importance of the measurement data differs from point to point. 
For example, in spectral deconvolution, the data near the peaks exert a comparatively significant impact on the analysis results, as shown in Fig. \ref{spectral_deconvolution}, and should thus be prioritized for measurement. 
However, as the true value is unknown before the measurement, a sequential experimental design framework is necessary, where the next measurement is designed based on the data obtained from the previous measurement \cite{Ford1980}. 
The method of sequentially designing experiments based on the results of data analysis is called active learning in the field of machine learning.
Specifically, active learning is described as the sequential selection of samples that will best improve the effectiveness of the predictor learned from a small volume of available data \cite{Hino2021}.
\par
Sequential experimental design based on the Gaussian process regression, which is a nonparametric model, is considered to improve the efficiency of spectral measurements.
The Gaussian process regression is widely employed as a predictor for active learning because its linearity allows us to obtain estimates and estimation accuracy analytically \cite{Rasmussen2006}.
Ueno et al. devised a sequential experimental design that selects the point with the highest estimation uncertainty as the next measurement point using active learning with the Gaussian process regression and succeeded in improving the efficiency of X-ray magnetic circular dichroism spectroscopy \cite{Ueno2018}.
However, active learning using the Gaussian process regression has certain limitations. No prior knowledge regarding the experiment can be introduced, and it is not applicable when the noise is large \cite{Hino2021}. \par
Conversely, in the analysis of spectral measurement data, several parametric models using theoretical formulas are proposed and we select the most plausible model and fit its model parameters from the data, 
even when the data are obtained by active learning using a nonparametric model \cite{Ueno2018,Ueno2021}.
Therefore, it is considered that sequential experimental design using the parametric model as a predictor of active learning improves the accuracy of parameter estimation and is applicable to various experiments.
However, in general, it is difficult to determine the estimation uncertainty analytically in active learning using parametric models.\par
In this study, we apply the Bayesian inference-based data analysis method proposed by Nagata et al. \cite{Nagata2012} to realize a sequential experimental design by active learning using any parametric model as a predictor.
Nagata et al.'s method, which numerically calculates the posterior distribution of models and their parameters by exchange Monte Carlo simulations, has been applied in various spectral measurements \cite{Nagata2019,Mototake2019,Katakami2022,Moriguchi2022,Kashiwamura2022,Ueda2023}.
We implemented active learning using the parametric model by computing the estimation uncertainty for each measurement point from the posterior distribution of the parameters obtained numerically using Nagata et al.'s method.
Our sequential experimental design using parametric models can be applied to various spectral measurements by selecting the appropriate model for the experiment.
Furthermore, since active learning is performed using the model for data analysis, the measurement points that improve the accuracy of model selection and parameter estimation are selected as the next measurement points.\par
Additionally, we applied our method to spectral deconvolution and Hamiltonian selection using Bayesian inference in X-ray photoelectron spectroscopy(XPS) and evaluated its effectiveness.
In the numerical experiment, our method succeeded in improving the signal-to-noise ratio of the measurement points that contribute to the model selection and parameter estimation. Therefore, our method improves the accuracy of model selection and parameter estimation while reducing the experiment time compared with the case without active learning or with active learning using nonparametric models.\par
The structure of this paper is as follows. In Sect. \ref{section2}, we describe a sequential experimental design framework for spectral measurement and its realization using active learning with parametric models.
In Sect. \ref{section3}, we describe the application of our method to Bayesian spectral deconvolution in XPS and the evaluation of its effectiveness. In Sect. \ref{section4}, we discuss the application of our method to the Bayesian Hamiltonian selection in XPS and the evaluation of its effectiveness.
In Sect. \ref{section5}, we conclude this paper and discuss future works. In Appendix \ref{AppendixA}, we disclose that with the problem settings employed in this study, it is difficult to conduct sequential experiments using the Gaussian process regression. 
In Appendix \ref{AppendixB}, we highlight the results of the estimation of other parameters that are omitted in Sects. \ref{section3}, \ref{section4}.
\par
\section{Proposed method}
\label{section2}
In this section, we describe the sequential experimental design framework for spectral measurement and its realization using active learning with parametric models.
In Sect. \ref{section2.1}, we describe a sequential experimental design considering the constraints of spectral measurements, and in Sect. \ref{section2.2}, we describe a method for selecting the next measurement point by active learning using parametric models.
\subsection{Sequential experimental design for spectral measurement}
\label{section2.1}
As shown in the Fig. \ref{spectral_deconvolution}, the measurement points have different levels of importance in the spectral measurement; however, the shape of the spectrum is unknown before the experiment is performed.
Therefore, it is better to focus on the important measurement points by selecting the next measurement points from the currently obtained data.
Here, we consider a sequential experimental design that considers two spectral measurement constraints.
The first constraint is that the resolution of the horizontal axis is fixed. This limits the measurement points to a finite set $\mathcal{X}$.
Thus, we first perform short measurements for all points on $\mathcal{X}$ and subsequently actively select points that need to be measured repeatedly.
The second constraint is that, due to the nature of the measurement instrument, the energy of the irradiated photons can only increase, and the number of times where the minimum energy increases to the maximum energy should be decreased.
Therefore, from the given data, we reduced the number of scans by selecting multiple $x\in\mathcal{X}$ as the next measurement points.\par
\begin{figure}[h]
    \centering
    \includegraphics*[width = 14.0cm]{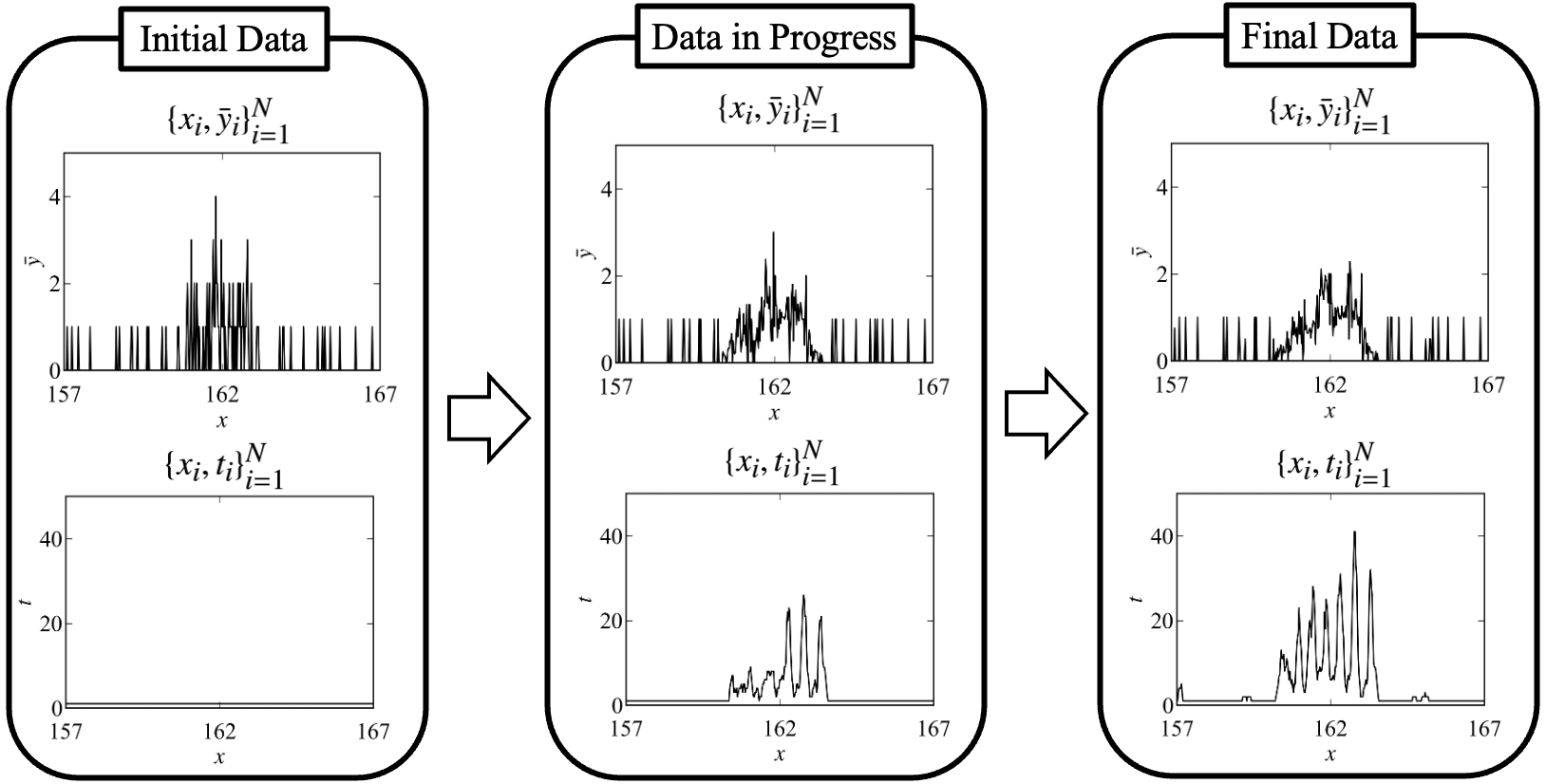}
    \caption{Flow of the sequential experimental design in spectral measurements. The upper figure shows the observed values per total measurement time $\bar{y}_i$, and the lower figure shows the total measurement time per measurement point $t_i$. Although the signal-to-noise ratio of the initial data is poor at all measurement points, it is expected that the signal-to-noise ratio of the data near the peak will be improved by repeating the sequential design of experiments.}
    \label{flow_spectral}
  \end{figure}

The formulation of the sequential experimental design is as follows. Suppose that $D = \{ (x_1,y_1),...,(x_N,y_N)\}$ is given as the initial data. 
Given data $D$, we choose $\{x_{next,1},...,x_{next,n}\}$ from a finite set $\mathcal{X}$ as the next measurement points and observe $\{y_{next,1},...,y_{next,n}\}$.
Thereafter, we update data $D$ by adding $\{(x_{next,1},y_{next,1}),...,(x_{next,n},y_{next,n})\}$. We repeat this update $k$ times.
For a measurement point $x_i$, we define the total measurement time per measurement time as $t_i = \#\{j|x_j = x_i\}\times (\textup{time per measurement})$ and the observed values per total measurement time as $\bar{y}_i = \frac{\sum_{x_j = x_i} y_j}{t_i}$.
The more measurement point $x_i$ is measured repeatedly, the larger the $t_i$ value and the better the signal-to-noise ratio of $\bar{y}_i$.
Therefore, we must consider an algorithm that measures the important measurement points intensively.
The flow of the sequential experimental design is shown in Fig. \ref{flow_spectral}.
As shown in Fig. \ref{flow_spectral}, in the sequential experimental design, the signal-to-noise ratio is poor at all points at first. However, as the experiment progresses, the signal-to-noise ratio near the peaks improves due to focused measurements.
\subsection{Active learning using a parametric model}
\label{section2.2}
\begin{figure}[h]
    \centering
    \includegraphics[width = 10cm]{"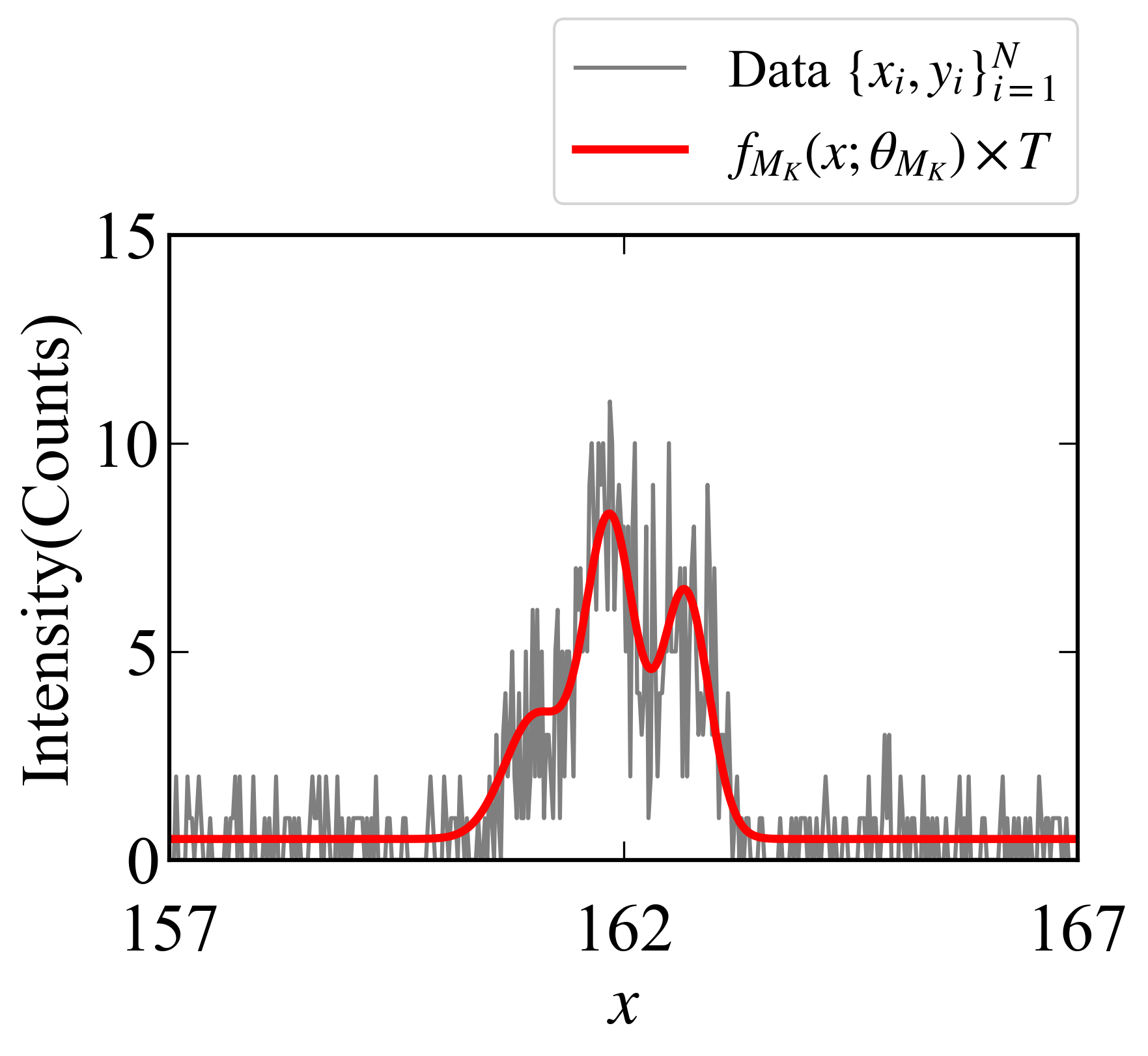"}
    \caption{
      Value of the modeling function $(f_{M}(x;\theta_M))$ and the example of the observed data $(y_1,.... ,y_N)$.
      We define the sum of three Gaussian functions as the modeling function and the Poisson distribution as the probability distribution
      (See Sect. \ref{section3} for the parameter values and the problem settings).
    } 
    \label{data_generation}
  \end{figure}

In this section, we describe a method to realize active learning by Bayesian inference using parametric models and to select the next measurement point.
First, we consider the parametric model $(M)$, in which the observed value $(y)$ for a measurement point $(x)$ is generated independently with a probability distribution:
\begin{align}
    y 	\sim p(y|f_{M}(x;\theta_M)),
\end{align}
where $\theta_M$ is the parameter of $M$, and $f_{M}(x;\theta_M)$ is the function used for modeling.
For example, if we define the sum of three Gaussian functions as the modeling function and the Poisson distribution as the probability distribution, we obtain Fig. \ref{data_generation}.
This assumes that the values of the modeling function with noise are observed.
 \par
  From the independence of probability distributions, the probability that data $D = \{x_i, y_i \}_{i=1}^L$ is given, assuming that a certain model $M$ and its parameters $(\theta_M)$ can be formulated as follows:
  \begin{align}
    p(D|\theta_M,M) = \prod_{i=1}^n p(y_i|f_M(x_i;\theta_M)),
  \end{align}
  where $\{x_i\}_{i=1}^n$ can be chosen by the sequential experimental design. Otherwise, $p(D|\theta_M) = 0$.\par
  From this probabilistic model, we consider estimating the parameters from the data.
  If $p(\theta_M|M)$ is the prior distribution of the model parameters, then the joint probability distribution, $p(D,\theta_M|M)$, is as follows:
\begin{align}
  p(D,\theta_M|M) = p(D|\theta_M,M)p(\theta_M|M).
\end{align}
From Bayes' theorem, the posterior distribution of parameter $\theta_M$, given data $D$ and model $M$, is given by
\begin{align}
  p(\theta_M|D,M) &= \frac{p(D,\theta_M|M)}{\int p(D,\theta_M|M)\textup{d}\theta_M}.
\end{align}
This posterior distribution can be employed to estimate these parameters.
Moreover, we consider selecting the most plausible model from model candidates, such as the selection of the number of peaks in the example shown in Fig. \ref{data_generation}.
Let $\mathcal{M}$ be the set of model candidates and $p(M)$ be the prior distribution of the model. Thus, the joint probability $(p(D,\theta_M,M))$ is given by
\begin{align}
    p(D,\theta_M,M) = p(D|\theta_M,M)p(\theta_M|M)p(M).
\end{align}
From Bayes' theorem, the posterior distribution of model $M$, given data $D$, is as follows:
\begin{align}
    p(M|D) &= \frac{\int p(D,\theta_M,M) \textup{d}\theta_M}{\sum_{M\in \mathcal{M}} \int p(D,\theta_M,M)\textup{d}\theta_M}.
\end{align}
This posterior distribution can be employed to select a plausible model.
For various cases of $p(y|f_{M}(x;\theta_M))$, the numerical sampling of $p(\theta_M|D,M)$ and the numerical computation of $p(M|D)$ can be realized by the exchange Monte Carlo method \cite{Nagata2012,Nagata2019,Mototake2019,Hukushima1996}. \par
Using the posterior distribution $(p(M|D),p(\theta_M|D,M))$, we consider the estimation uncertainty at the measurement points of each model to select the next measurement points.
We assume that the model candidate set $(\mathcal{M})$, a prior distribution of models $(p(M))$, and a prior distribution of parameters $(p(\theta_M|M))$ are given.
Let $D = \{x_i, y_i \}_{i=1}^L$ be the data currently available.
We define $\widehat{M} = \max_{M\in\mathcal{M}} p(M|D)$ as the best predicted model and $M^{\circ} = \max_{M \neq \widehat{M}}p(M|D)$ as the second-best predicted model.
Furthermore, we define the best model map estimator of parameter $\grave{\theta} = \max_{\theta_{\widehat{M}}}p(\theta_{\widehat{M}}|D,\widehat{M})$ and the estimator of the modeling function, $\grave{f}(x) = f_{\widehat{M}}(x;\grave{\theta})$.
Here, we formulate the estimated uncertainties, i.e., $G(x;\widehat{M})$ and $G(x;M^{\circ})$, for models $\widehat{M}$ and $M^{\circ}$, respectively, at the measurement point $x$, as follows:
\begin{align}
    \label{bestmodel}
    G(x;\widehat{M}) = \int &\textup{KL}(p(y|\grave{f}(x)),p(y|f(x;\theta_{\widehat{M}})))p(\theta_{\widehat{M}}|\widehat{M},D)\textup{d}\theta_{\widehat{M}} \\
    \label{secondmodel}
    G(x;M^{\circ}) = \int &\textup{KL}(p(y|\grave{f}(x)),p(y|f(x;\theta_{M^{\circ}})))p(\theta_{M^{\circ}}|M^{\circ},D)\textup{d}\theta_{M^{\circ}}.
\end{align}
These formulations calculate the distance between the probability distribution $(p(y|\grave{f}(x)))$ and $(p(y|f(x;\theta_{\widehat{M}})),p(y|f(x;\theta_{M^{\circ}})))$, obtained from the posterior distribution $(p(\theta_M|M,D))$ of each model.
These integrations can be computed by the exchange Monte Carlo sampling of $p(\theta_M|M,D)$.\par
Using this estimation uncertainty, we describe a method for selecting the next measurement point.
We choose $\{x_{next,1},...,x_{next,\frac{n}{2}}\}$ and $\{x_{next,\frac{n}{2} + 1},...,x_{next,n}\}$ from $\mathcal{X}$, where $G(x;\widehat{M})$ and $G(x;M^{\circ})$ are large, respectively.
The selection of points using $G(x;\widehat{M})$ corresponds to focusing on important points for the parameter estimation of the best predicted model,
and selecting points using $G(x;M^{\circ})$ corresponds to focusing on the difference between the best predicted model and the second-best predicted model. \par
In Eq. \ref{bestmodel} and \ref{secondmodel}, $\textup{KL}(x,y)$ is the Kullback -- Leibler divergence of $x$ and $y$.
For photon counting measurements, such as XPS, the probability distribution of the observed data $(p(y|f_{M}(x;\theta_M)))$ is considered to be 
\begin{align}
\textup{Poisson}(y;f_{M}(x;\theta_M)\times T) = \frac{\exp(-f_{M}(x;\theta_M)\times T)\left(f_{M}(x;\theta_M)\times T\right)^y}{y!},
\end{align}
with T representing the measurement time. Thus, the Kullback--Leibler divergence is expressed as follows:
\begin{align}
    &\textup{KL}(p(y|\grave{f}(x)),p(y|f(x;\theta_{M}))) \\ 
    &=  - T\grave{f}(x) + T \grave{f}(x)\log\left(T \grave{f}(x)\right) + Tf(x;\theta_{M}) - T\grave{f}(x)\log\left(T f(x;\theta_{M})\right) \\
    &= T\left(- \grave{f}(x) + f(x;\theta_{M}) +  \grave{f}(x)\log\left( \frac{\grave{f}(x)}{f(x;\theta_{M})}\right)  \right).
\end{align}
Thus, the value of $T$ does not change the selection result in this method.\par 
Although there are other possible selection methods, in this study, we focused on this method and evaluated its effectiveness through numerical experiments.
Furthermore, this method can also be applied in the problem setting where the model is fixed by setting $\widehat{M} = M^{\circ}$.
\section{Validation of our method for Bayesian spectral deconvolution}
\label{section3}
In this section, we consider the Bayesian spectral deconvolution in XPS, a problem setting, to estimate the number of peaks and parameters with prior distributions \cite{Nagata2019}.
In Sect. \ref{section3.1}, we describe the problem setting of the Bayesian spectral deconvolution in XPS. In Sect. \ref{section3.2}, we describe the detailed algorithm of the sequential experimental design in Bayesian spectral deconvolution.
In Sect. \ref{section3.3}, we discuss the results obtained with artificial data and evaluate its effectiveness.
\subsection{Problem setting of the Bayesian spectral deconvolution in XPS}
\label{section3.1}
We consider a model in which the expected value of the XPS observation per unit time is represented by a linear sum of multiple peaks and backgrounds.
Let $M_K$ be a model with $K$ peaks, the parameter set $(\theta_{M_K})$ be $\theta_{M_K} = \{\{a_k, \mu_k, \sigma_k\}_{k = 1}^K ,B\}$, and the modeling function $(f_{M_K}(x;\theta_K))$ be $f_{M_K}(x;\theta_K) = \sum_{k = 1}^{K} a_k\exp\left(-\frac{(x-\mu_k)^2}{\sigma_k^2}\right) + B$ (where $a_k,\mu_k,\sigma_k$, and $B$ correspond to the peak intensity, peak position, peak width, and background intensity, respectively).
Since the measurement is performed by photon counting in XPS, the probability distribution of the number of observed photons $(p(y|f_{M}(x;\theta_M)))$ is considered to be $\textup{Poisson}(y;f_{M}(x;\theta_M)\times T)$ with measurement time $T$.
The posterior distributions, $p(M_K|D),p(\theta_K|D,M_K)$, for data $D = \{(x_1,y_1),...,(x_N,y_N) \}$ can be obtained by the exchange Monte Carlo method \cite{Nagata2019}.
In this problem setting, the goal is to select the number of peaks and estimate the parameters of the peak positions with high accuracy from experiments with a short total measurement time.
\subsection{Detailed algorithms in the Bayesian spectral deconvolution}
\label{section3.2}
In Sect. \ref{section3.2}, the set of candidate models must be given in advance; however, in the Bayesian spectral deconvolution, the number of peaks $(K)$ can take any integer.
Therefore, we consider changing the candidate model set sequentially. \par
We define the initial model set as $\mathcal{M} = \{M_1,M_2\}$.
At each step, let $\widehat{K}$ be the number of peaks of the best predicted model ($\widehat{M}$ ($\widehat{M} = M_{\widehat{K}}$)), and we use the following model set in the next estimation:
\begin{align}
    \mathcal{M} = \left\{
        \begin{array}{ll}
            \{M_1,M_2\} &  (\widehat{K} = 1)\\
            \{M_{\widehat{K} - 1},M_{\widehat{K}},M_{\widehat{K}+ 1}\}. & (\textup{otherwise})
        \end{array}
    \right.
\end{align}
The specific algorithm is shown in Algorithm \ref{Algorithm_Deconvolution}.
\vspace{0.5cm}
\begin{figure}[h]
  \begin{algorithm}[H]
      \caption{Sequential experimental design for Bayesian spectral deconvolution}
      \label{Algorithm_Deconvolution}
      \begin{algorithmic}[1]
      \REQUIRE Number of measurement points per one experiment $n$, Number of experiments $k$, Measurement points set $\mathcal{X} = \{x_i\}_{i=1}^N$
      \ENSURE Data $D = \{(x_i,y_i)\}_{i = 1}^{N + n \times k} $
      \STATE Measure $y_1,...,y_N$ with $x_1,...,x_N$.
      \STATE Data $D = \{(x_i,y_i)\}_{i\in \{1,...,N\}} $
      \STATE Candidate model set $\mathcal{M} = \{M_1,M_2\}$
      \FOR{$i \in \{1,...,k\}$}
          \STATE $\widehat{M} = M_{\widehat{K}}= \text{max}_{M_K \in \mathcal{M}} P(M_K|D)$, $M^{\circ} = \text{max}_{M\neq \widehat{M}} P(M|D)$.
          \STATE Calculate estimation accuracies $\{G(x_i;\widehat{M})\}_{x_i \in \mathcal{X}}, \{G(x;M^{\circ})\}_{x_i \in \mathcal{X}}$.
          \STATE Select $\frac{n}{2}$ points $\{x_{next,1},...,x_{next,\frac{n}{2}}\}$ from $\mathcal{X}$ in descending order of $\{G(x_i;\widehat{M})\}_{x_i \in \mathcal{X}}$.
          \STATE Select $\frac{n}{2}$ points $\{x_{next,\frac{n}{2} + 1},...,x_{next,n}\}$ from $\mathcal{X}$ in descending order of $\{G(x_i;M^{\circ})\}_{x_i \in \mathcal{X}}$.
          \STATE Measure $\{y_{next,1},...,y_{next,n}\}$ in selected points $\{x_{next,1},...,x_{next,n}\}$.
          \STATE $D = D \cup \{(x_{next,1},y_{next,1}),...,(x_{next,n},y_{next,n}) \}$
          \IF{$\widehat{K} = 1$}
          \STATE $\mathcal{M} = \{M_1 ,M_2\}$
          \ELSE
          \STATE $\mathcal{M} = \{M_{\widehat{K}-1} ,M_{\widehat{K}}, M_{\widehat{K}+1}\}$
          \ENDIF
      \ENDFOR
      \end{algorithmic}
  \end{algorithm}
\end{figure}
\subsection{Results of the Bayesian spectral deconvolution}
\label{section3.3}
Let the true model be the model $M_3$ with $K=3$ peaks, and the true values of the parameters $(\theta_{M_3}^* = \{ \{a_k^*,\mu_k^*,\sigma_k^*\}_{k=1}^3,B^*\})$ be as follows:
\begin{align}
    \begin{pmatrix}
        a_1^*\\
        a_2^*\\
        a_3^*\\
    \end{pmatrix} 
    = 
    \begin{pmatrix}
        0. 587\\
        1.522\\
        1.183\\
    \end{pmatrix}  ,\ 
    \begin{pmatrix}
        \mu_1^*\\
        \mu_2^*\\
        \mu_3^*\\
    \end{pmatrix}
    = 
    \begin{pmatrix}
        161.032\\
        161.852\\
        162.677\\
    \end{pmatrix} ,\ 
    \begin{pmatrix}
        \sigma_1^*\\
        \sigma_2^*\\
        \sigma_3^*\\
    \end{pmatrix}
    = 
    \begin{pmatrix}
        0.341\\
        0.275\\
        0.260\\
    \end{pmatrix},\ 
    B = 0.1.
  \end{align}
  This true value is the same as that reported by Nagata et al. \cite{Nagata2019}, which is set so that the number of peaks is somewhat difficult to estimate.
  The modeling function $(f_{M_3}(x;\theta_{M_3}^*))$ is shown in Fig. \ref{data_generation}.
  In this situation, let $\eta_a = 2.0, \lambda_a = 1.0,$ $\nu_0 = 157.0,\xi_0 = 167.0, \eta_\sigma = 10.0,\lambda_\sigma = 2.5,$ $\nu_B = 0.1,\xi_B = 0.01$, and we set the prior distributions of $\{a_k,\mu_k,\sigma_k\}_{k=1}^K,B$ be set as follows:
  \begin{align}
    \varphi(a_k)& = \textup{Gamma} \left(a_k;\eta_a,\lambda_a\right)\\ 
    &= \frac{1}{\Gamma(\eta_a)}(\lambda_a)^{\eta_a}(a_k)^{\eta_a-1}\exp\left(-\lambda_a a_k\right) \\
    \varphi(\mu_k) &= U(\nu_0,\nu_1) \\
    \varphi(\sigma_k) &= \textup{Gamma} \left(\frac{1}{\sigma_k^2};\eta_{\sigma},\lambda_{\sigma}\right)\\
    \varphi(B) &= N(B;\nu_B,{\xi_B}^2), \\
  \end{align}
where $\Gamma(\eta_a)$ is the gamma function, $U(\nu_0,\nu_1)$ is the uniform distribution on $[\nu_0,\nu_1]$, and $N(B;\nu_B,{\xi_B}^2)$ is the Gaussian distribution of mean $\nu_B$ and variance ${\xi_B}^2$. \par
Let the prior distribution of the model set be $\varphi(M_k) = \frac{1}{|\mathcal{M}|}$, the time for one measurement in the sequential experiment $T$ be $T = 1$, the vertical resolution of the experiment be $0.025\ \textup{(eV)}$, the number of data $N$ be $N = 400$, and 
the candidate set of measurement points $\mathcal{X}$ be $\mathcal{X} = \{157 + 0.025(i-1)\ \textup{(eV)}\}_{i=1}^{400}$.
First, we measure all points on $\mathcal{X}$ with a measurement time $T=1$.
Thereafter, we repeat the experiment $k=160$ times by sequentially selecting $n=10$ points to be measured next 
so that the total measurement time $T_{sum} = N\times T + n\times k \times T = 2000$.
To evaluate the effectiveness of our method, we compare the result with another method in which the total measurement time $T_{sum} = 2000$ with $T = 5$ at all measurement points and one in which the total measurement time $T_{sum} = 6000$ with $T = 15$ at all measurement points. 
We denote these experiments in which the same measurement is conducted at all points as static experiments. \par
\begin{figure}[h]
    \centering
    \includegraphics*[width = 5.0cm]{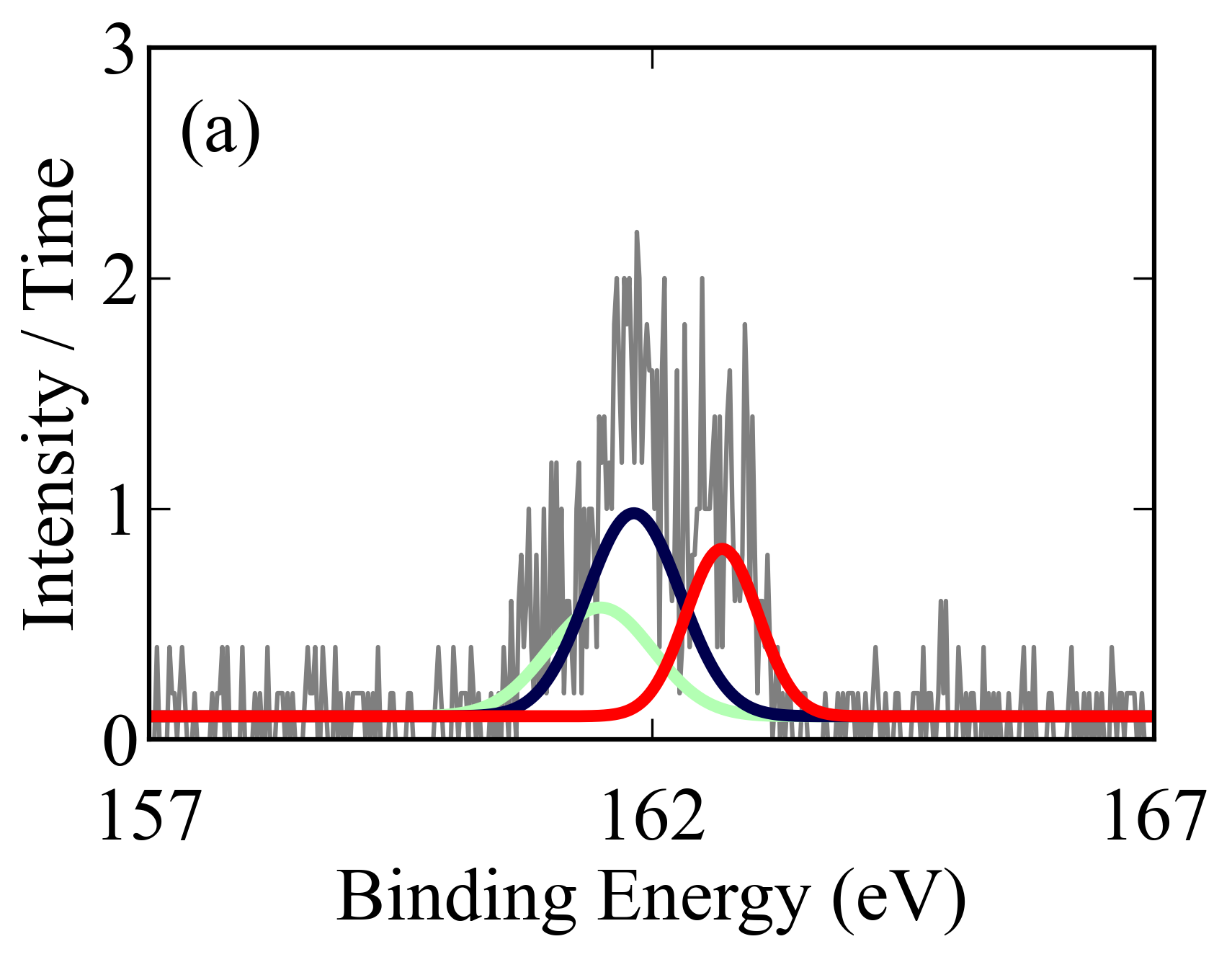}
    \includegraphics*[width = 5.0cm]{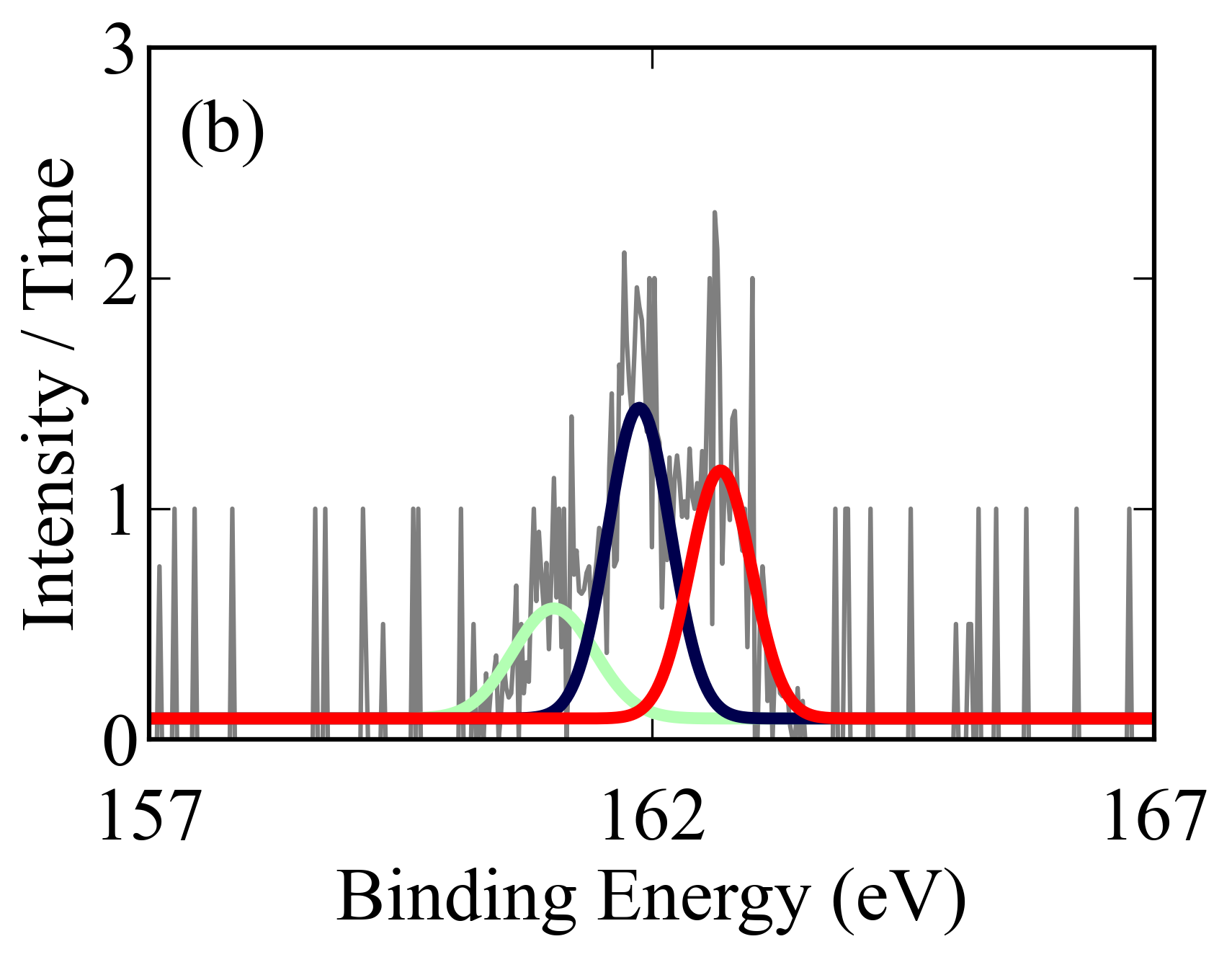}
    \includegraphics*[width = 5.0cm]{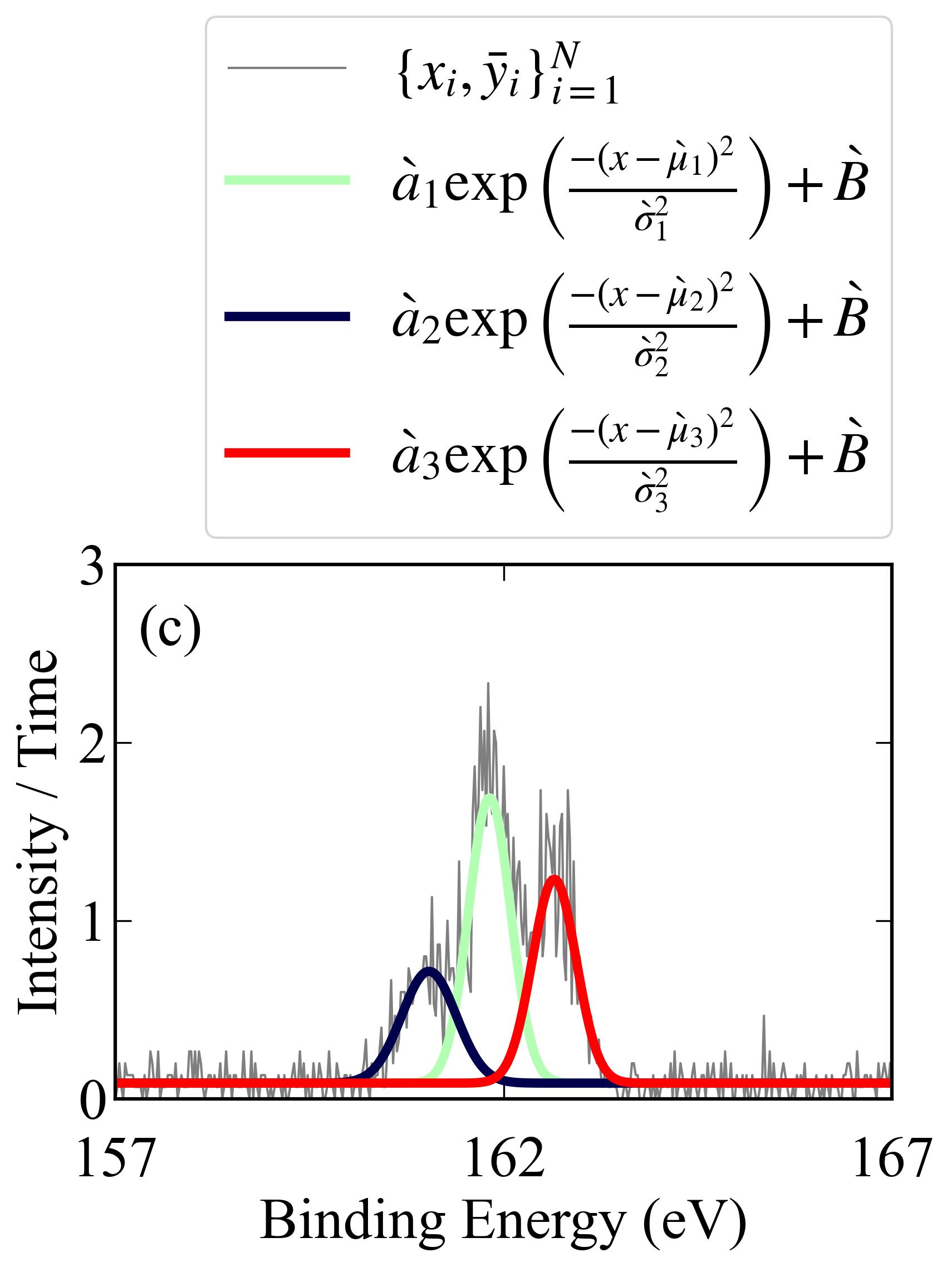} \\
    \includegraphics*[width = 5.0cm]{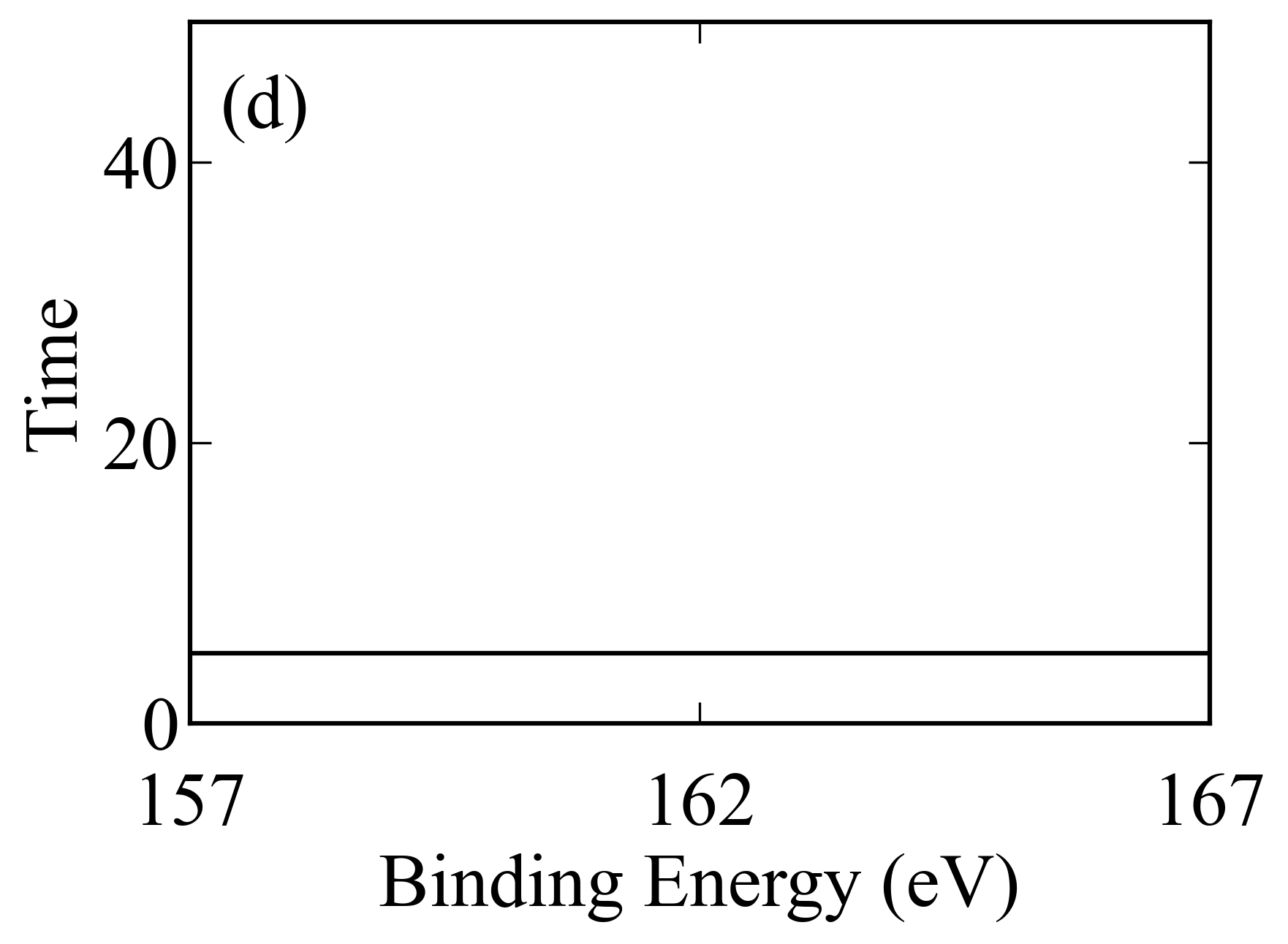}
    \includegraphics*[width = 5.0cm]{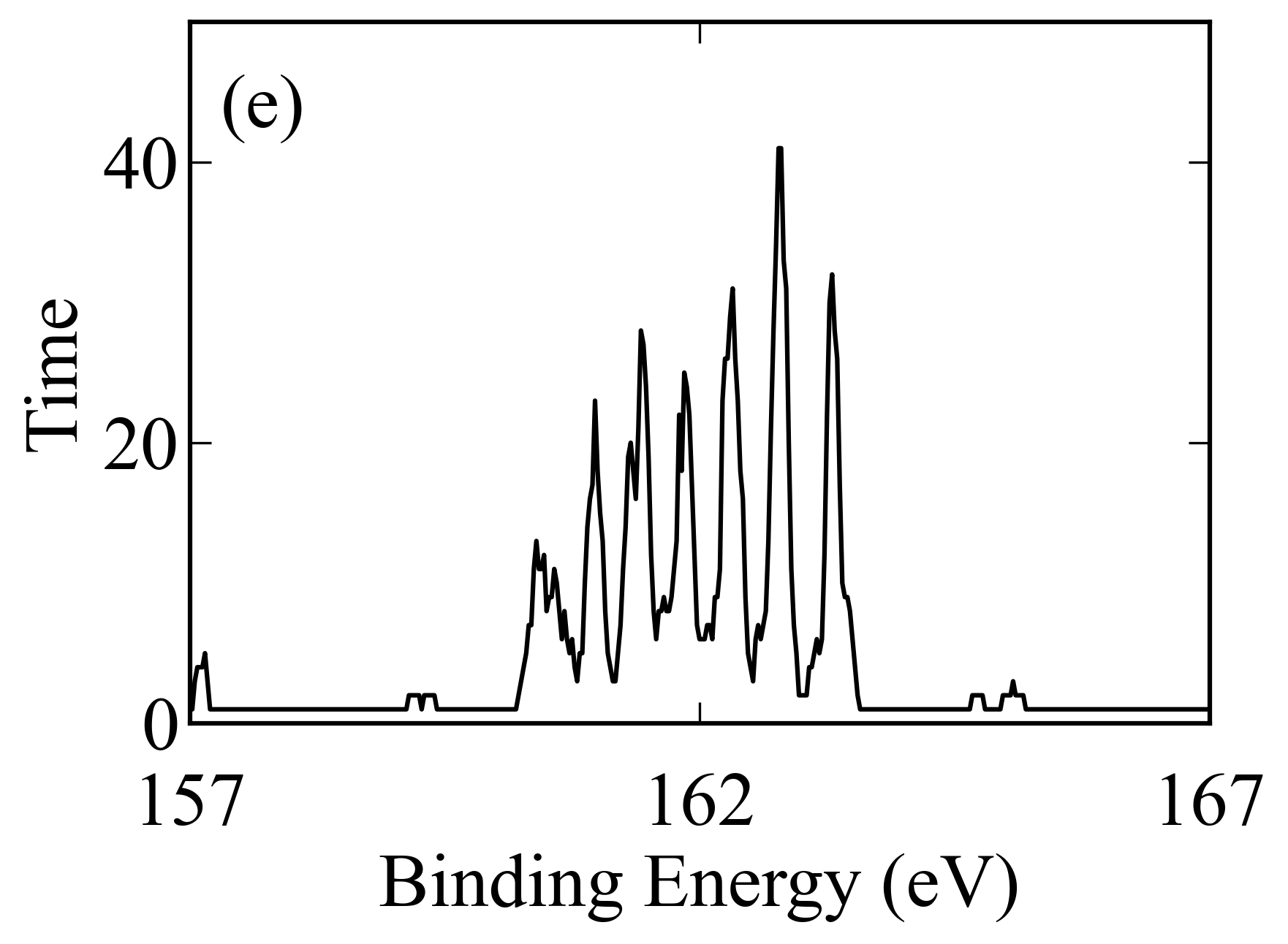}
    \includegraphics*[width = 5.0cm]{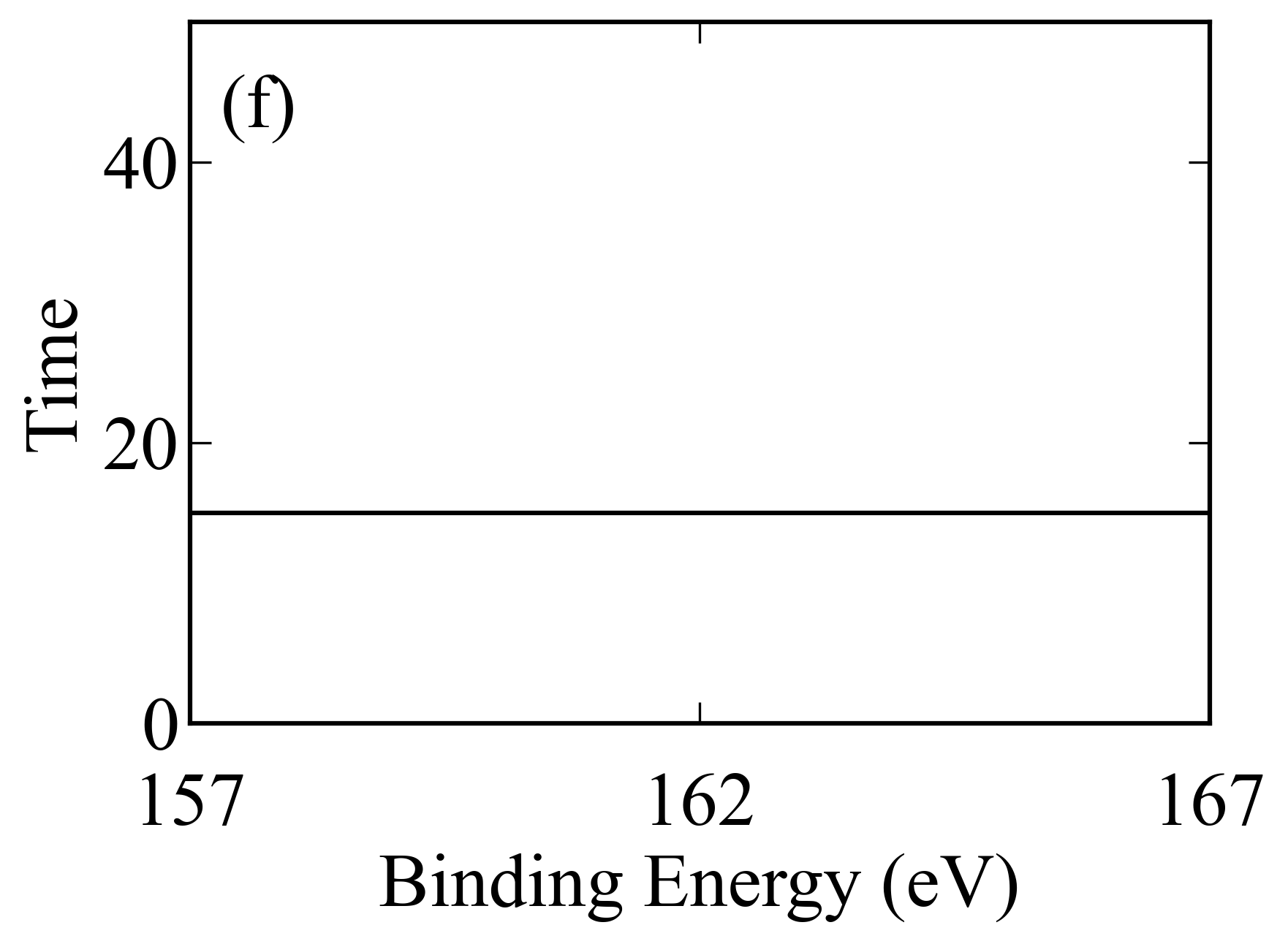}
    \caption{Data and fitting obtained by sequential experiments on the Bayesian spectral deconvolution. The upper figure shows that the number of photons observed per unit time $(\bar{y}_i)$ and the fitting by the map estimator $(\grave{\theta}_{M_3} = \underset{\theta}{\text{argmax}}p(\theta| D, M_3))$. The lower figure shows the total measurement time per measurement point. 
    Panels (a),(d) show the cases of a static experiment with a total measurement time of $T_{sum} = 2000$. Panels (b),(e) show the cases of a sequential experiment with a total measurement time of $T_{sum} = 2000$. Panels (c),(f) show the cases of a static experiment with a total measurement time of $T_{sum} = 6000$.
    }
    \label{fitting_deconvolution}
  \end{figure}

The data and the fitting by the map estimator ($\grave{\theta}_{M_3} = \{\{\grave{a}_k, \grave{\mu}_k, \grave{\sigma}_k\}_{k = 1}^K ,\grave{B}\} = \underset{\theta}{\text{argmax}}p(\theta| D, M_3))$ are shown in Fig. \ref{fitting_deconvolution} (The parameter indices are set so that $\mu_1<\mu_2<\mu_3$).
  This figure shows that the sequential experimental design focuses on the measurement points near the peaks that are considered to be important in the spectral deconvolution and improve the signal-to-noise ratio near the peaks. \par
\begin{figure}[h]
    \centering
    \includegraphics*[width = 5.0cm]{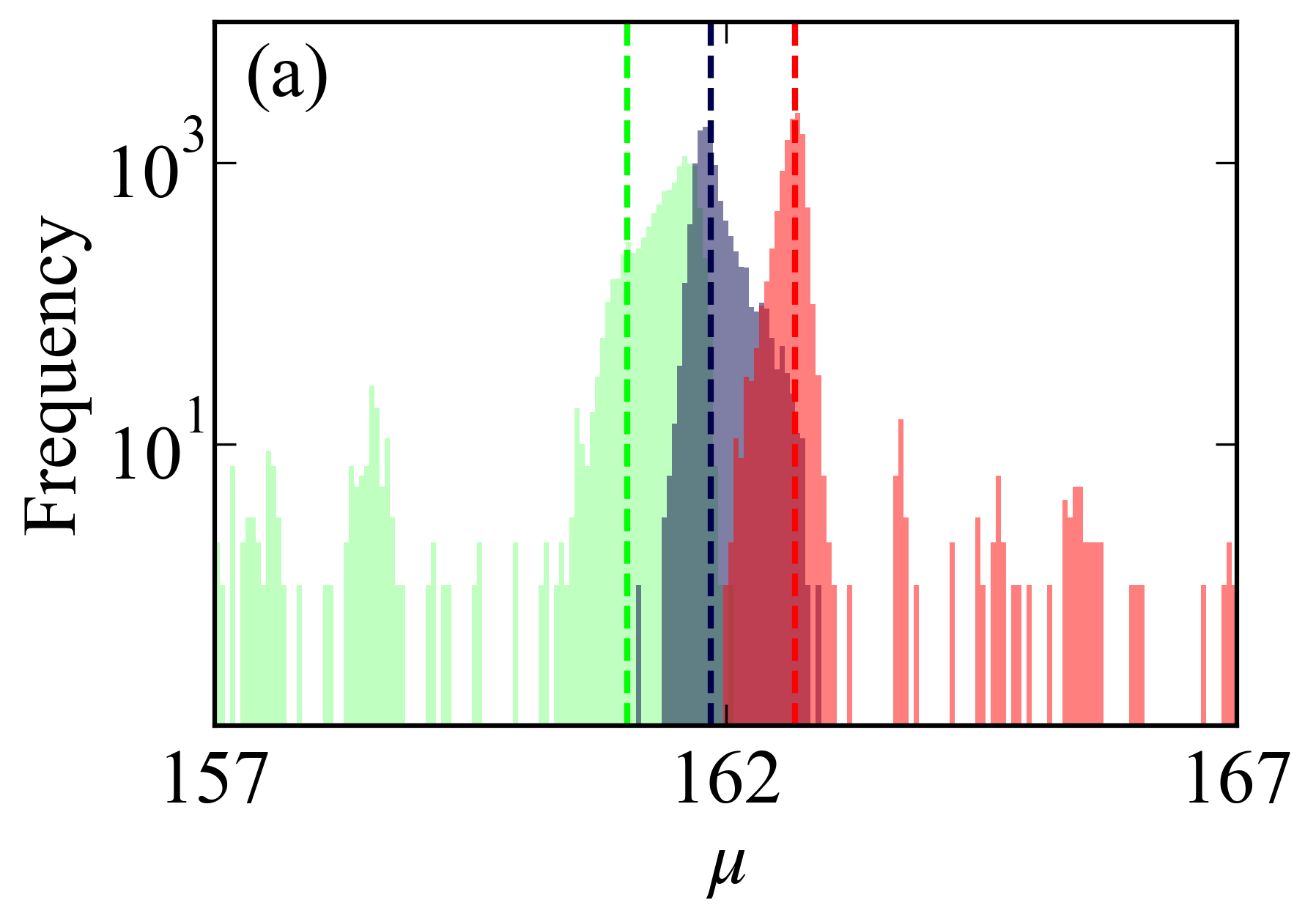}
    \includegraphics*[width = 5.0cm]{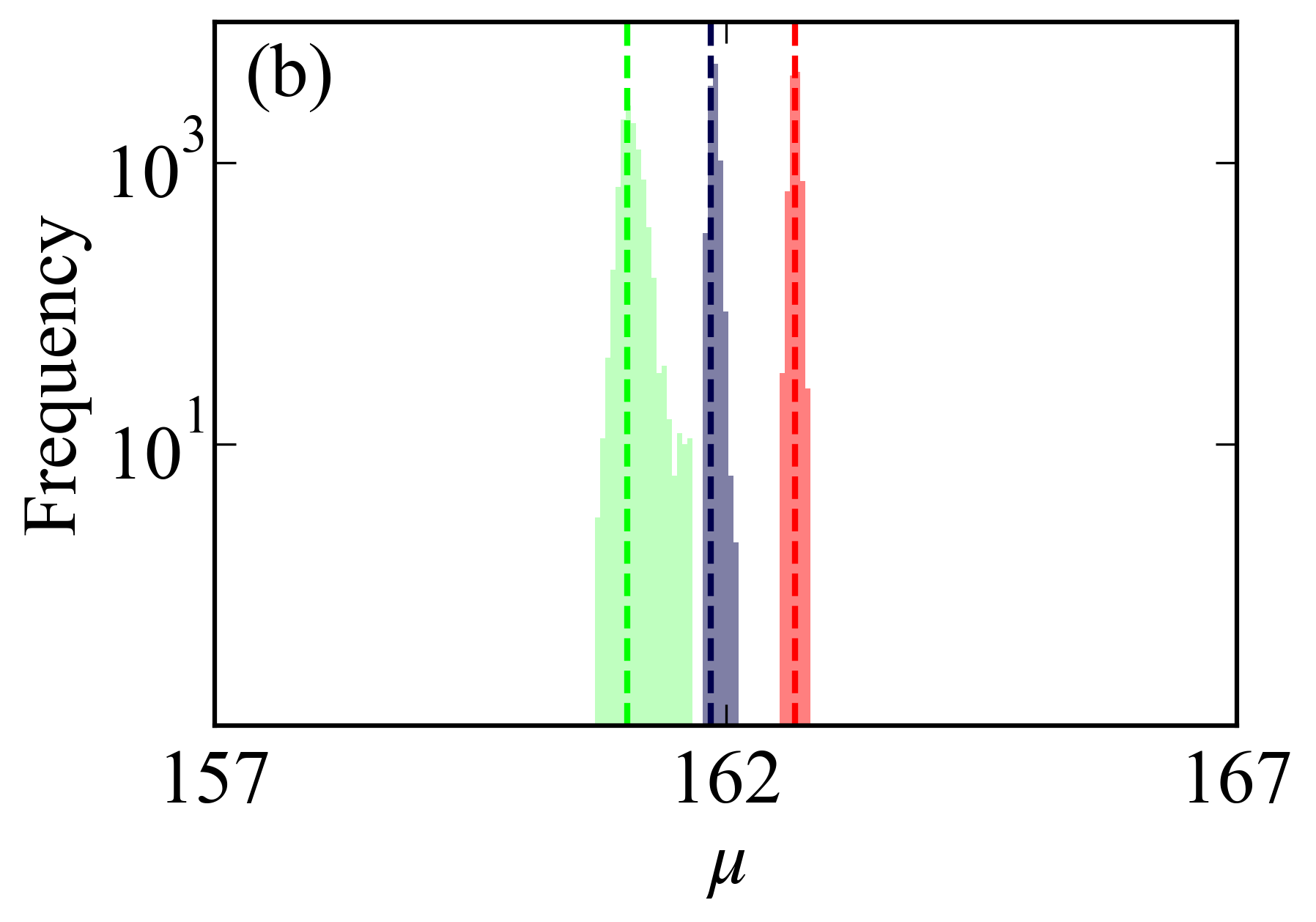}
    \includegraphics*[width = 5.0cm]{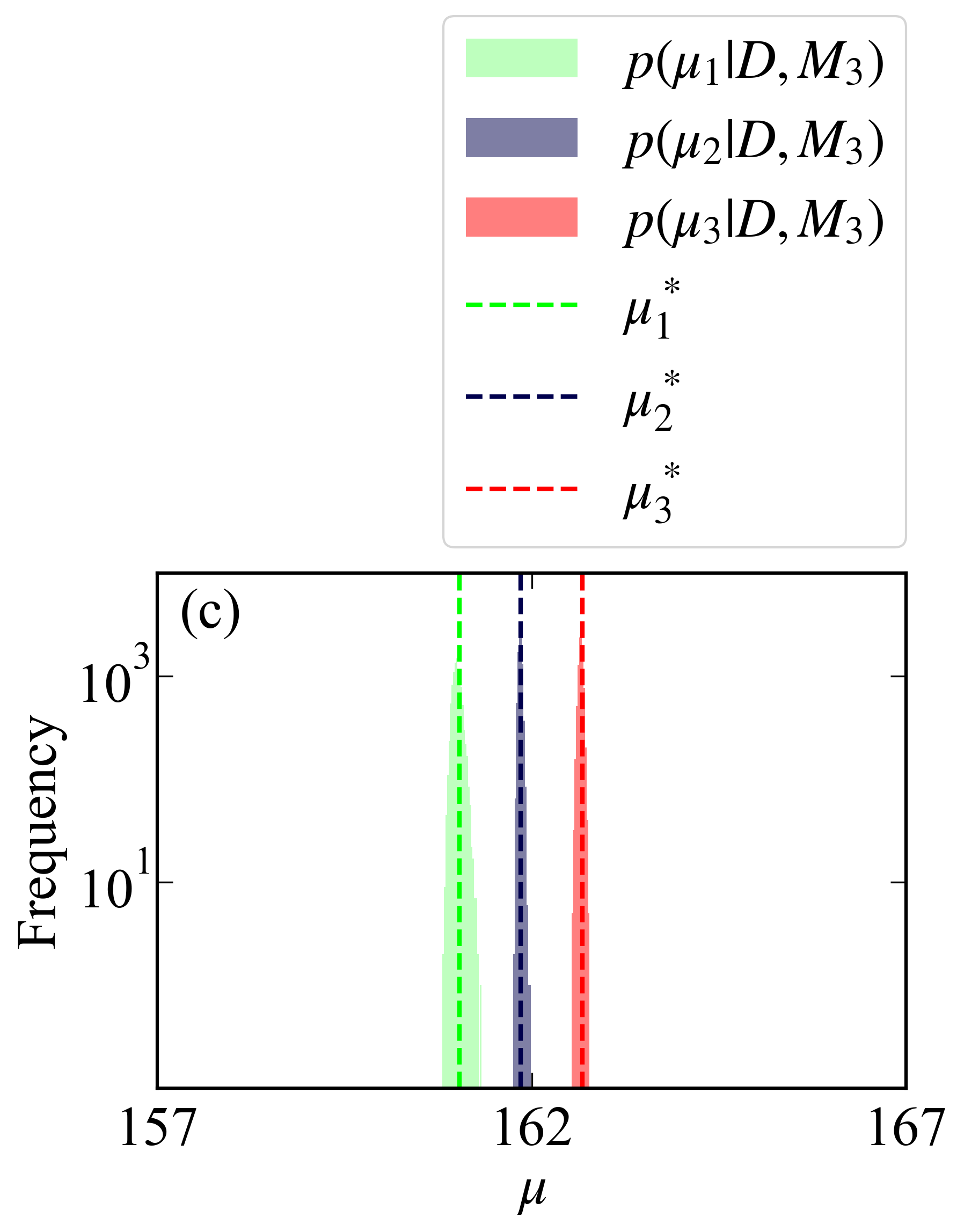}
    \caption{Parameter estimation by the posterior distribution $(p(\mu_1,\mu_2,\mu_3|D,M_3))$ of the peak positions $(\mu_1,\mu_2,\mu_3)$. The dashed line shows the true parameter values. Panel (a) shows the case of a static experiment with a total measurement time of $T_{sum} = 2000$. Panel (b) shows the case of a sequential experiment with a total measurement time of $T_{sum} = 2000$. Panel (c) shows the case of a static experiment with a total measurement time of $T_{sum} = 6000$.}
    \label{parameter_deconvolution}
  \end{figure}

  In addition, we estimate the peak position parameters $(\mu_1,\mu_2,\mu_3)$ assuming that the true model $K=3$ (The parameter indices are set so that $\mu_1 < \mu_2 < \mu_3$).
  The results of the parameter estimation are shown in Fig. \ref{parameter_deconvolution}.
  This figure shows that the width of the posterior distribution is shorter than that for a static experiment with $T_{sum}=2000$, and is similar to that for a static experiment with  $T_{sum}=6000$.
  This result indicates that the time required for the experiment has been reduced to one-third of the original. \par
\begin{figure}[h]
    \centering
    \includegraphics*[width = 15.0cm]{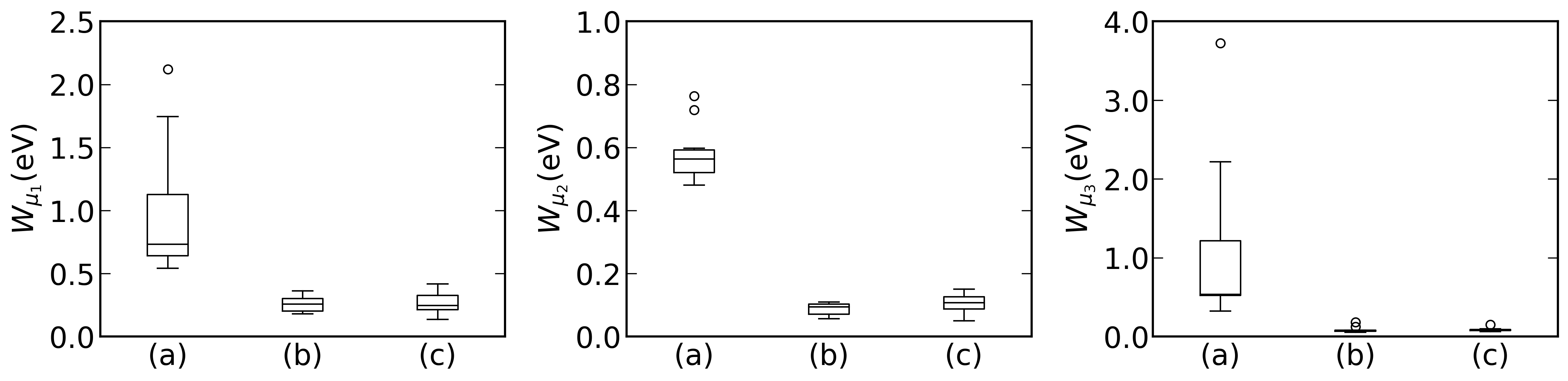}
    \caption{Boxplots representing the parameter estimation accuracy of the peak positions. The left panel, middle panel, and right panel show the boxplots of $W_{\mu_1},W_{\mu_2}$, and $W_{\mu_3}$ respectively. Label (a) highlights the case of a static experiment with a total measurement time $T_{sum} = 2000$. Label (b) highlights the case of a sequential experiment with a total measurement time of $T_{sum} = 2000$. Label (c) highlights the case of a static experiment with the total measurement time of $T_{sum} = 6000$.}
    \label{Width_deconvolution}
  \end{figure}

  Furthermore, we repeat the above trial 10 times independently to confirm the statistical properties. We define the indices to evaluate the width of the peak position parameter estimation as follows:
  \begin{align}
    W_{\mu_1} = \max_{\alpha \in [0.025,0.975]}|\mu _1^* - \mu_{1,\alpha}|, W_{\mu_2} = \max_{\alpha \in [0.025,0.975]}|\mu_2^* - \mu_{2,\alpha}|,W_{\mu_3} = \max_{\alpha \in [0.025,0.975]}|\mu_3^* - \mu_{3,\alpha}|,
  \end{align}
  where 
  \begin{align}
    \mu_{1,\alpha} &= \min_{\mu} \left\{ \left(\int_{\mu_1 < \mu}p(\mu_1|D,K)\textup{d}\mu_1\right) > \alpha \right\}, \\
    \mu_{2,\alpha} &= \min_{\mu} \left\{ \left(\int_{\mu_2 < \mu}p(\mu_2|D,K)\textup{d}\mu_2\right) > \alpha \right\}, \\
    \mu_{3,\alpha} &= \min_{\mu} \left\{ \left(\int_{\mu_3 < \mu}p(\mu_3|D,K)\textup{d}\mu_3\right) > \alpha \right\}. \\
   \end{align}
   These indices represent the deviations between the 95\% confidence intervals of the parameter estimation and the true parameters $\{\mu_1^*,\mu_2^*,\mu_3^*\}$.
   The boxplots of $W_{\mu_1},W_{\mu_2}$, and $W_{\mu_3}$ for the 10 trials are shown in Fig. \ref{Width_deconvolution}.
  This figure shows that the estimations afforded by our method are more accurate than those by the static experiment with $T_{sum}=2000$ and are as accurate as those afforded by the static experiment with $T_{sum}=6000$.\par
\begin{figure}[h]
    \centering
    \includegraphics*[width = 5.0cm]{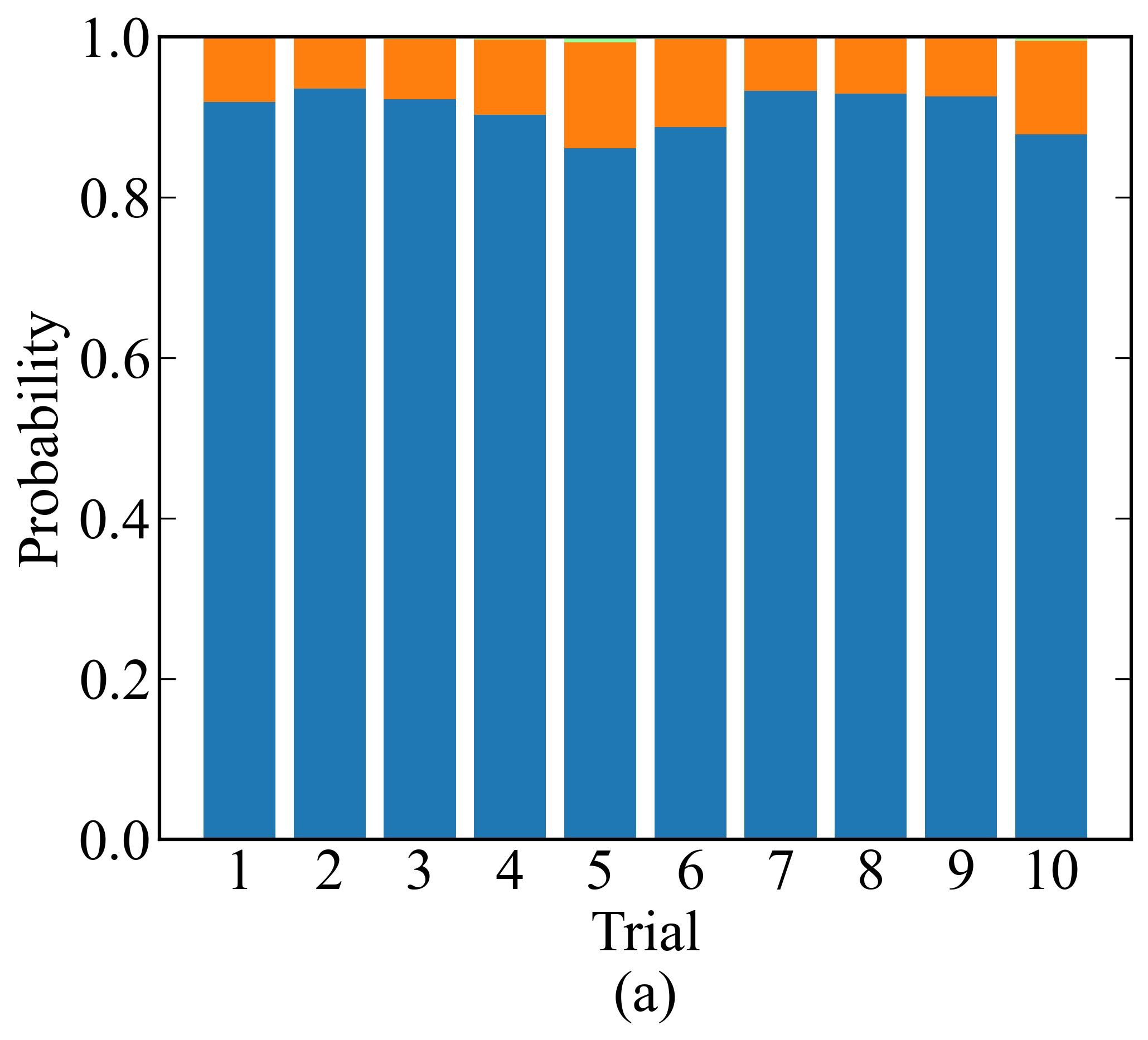}
    \includegraphics*[width = 5.0cm]{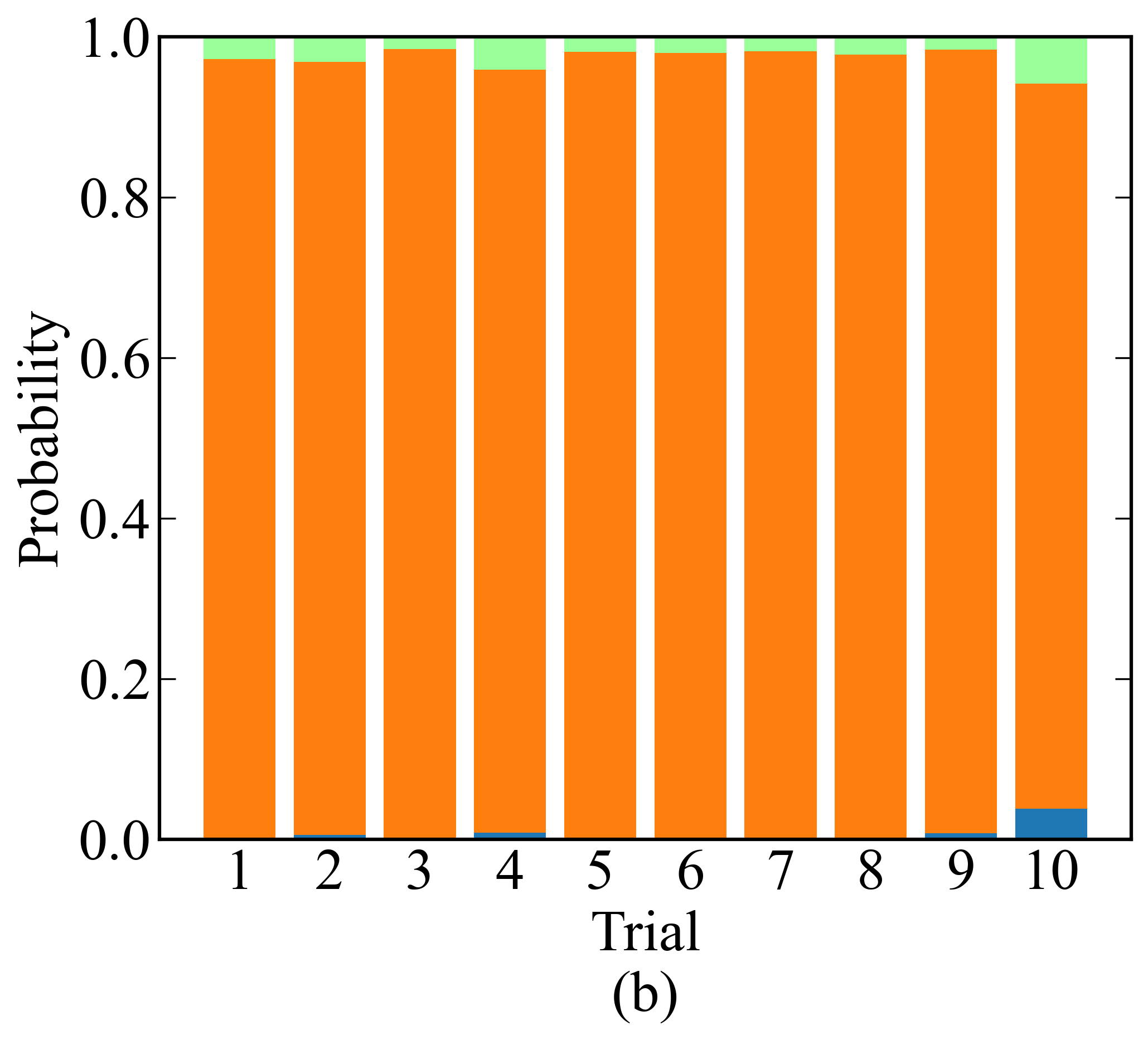}
    \includegraphics*[width = 5.0cm]{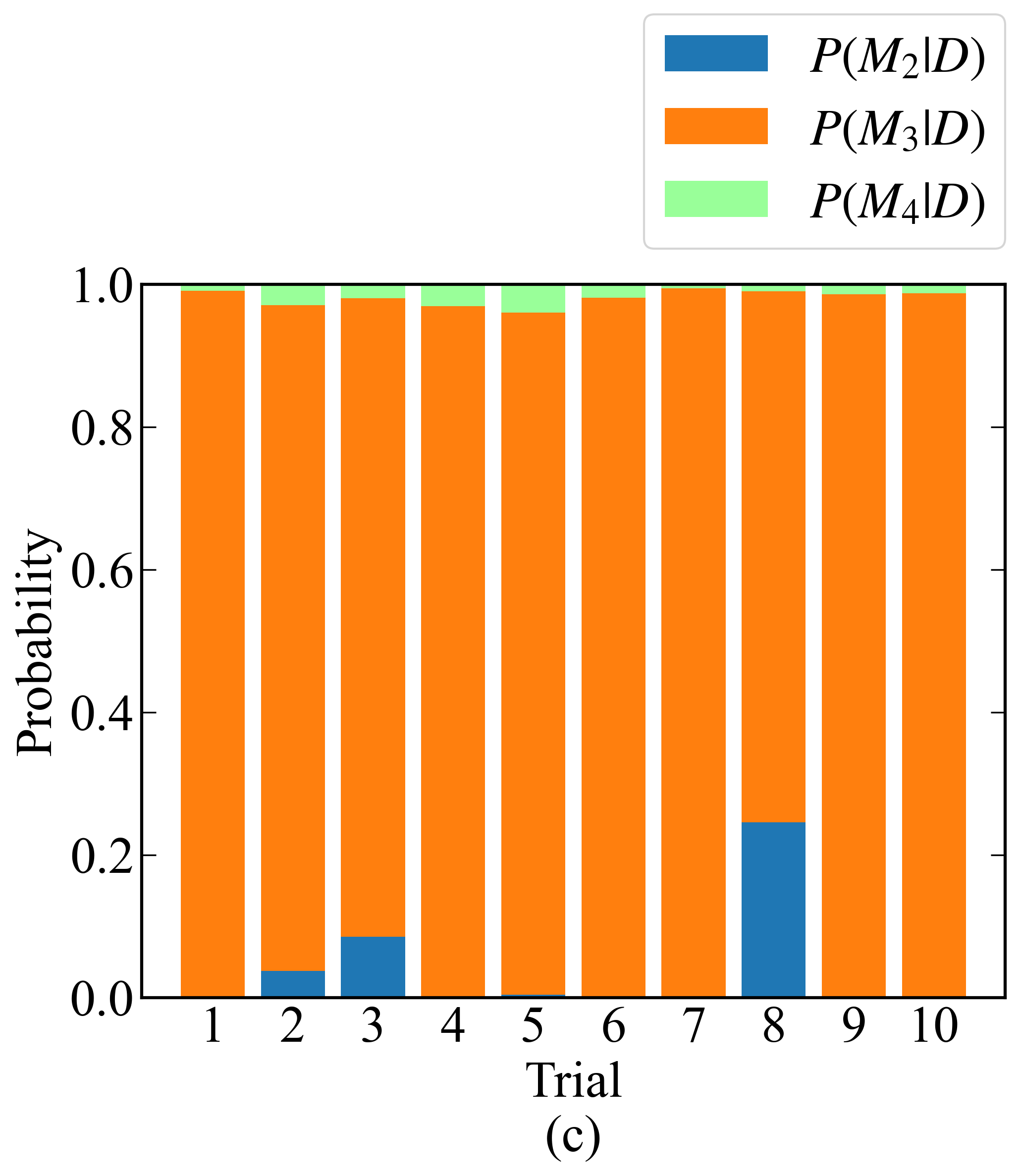}
    \caption{Bar graphs of $P(M_2|D),P(M_3|D),P(M_4|D)$ for the 10 independent trials. Panel (a) shows the case of a static experiment with a total measurement time of $T_{sum} = 2000$. Panel (b) shows the case of a sequential experiment with a total measurement time of $T_{sum} = 2000$. Panel (c) shows the case of a static experiment with a total measurement time of $T_{sum} = 6000$.}
    \label{Probability_deconvolution}
  \end{figure}

  Moreover, we calculated the posterior distribution $(P(M_2|D),P(M_3|D),P(M_4|D))$ for the 10 independent trials, with \ $\mathcal{M} = \{M_2,M_3,M_4\}$. The bar graphs of $P(M_2|D),P(M_3|D),P(M_4|D)$ are shown in Fig. \ref{Probability_deconvolution}.
  This figure shows that the model selection accuracy of our method is higher than that of the static experiment with $T_{sum}=2000$ and is similar to that of the static experiment with $T_{sum}=6000$.
  These results also indicate that the time required for the experiment has been reduced to one-third of the original.
  \section{Validation of our method for Bayesian Hamiltonian selection}
  \label{section4}
  In this section, we consider the Bayesian Hamiltonian selection in XPS, a problem setting, to select the most plausible Hamiltonian from the candidates and estimate its parameter \cite{Mototake2019}.
  In Sect. \ref{section4.1}, we describe the problem setting of the Bayesian Hamiltonian selection in XPS. In Sect. \ref{section4.2}, we describe the detailed algorithm of the sequential experimental design in the Bayesian Hamiltonian selection.
  In Sect. \ref{section4.3}, we discuss the results obtained with artificial data and evaluate its effectiveness.
  \subsection{Problem setting of the Bayesian Hamiltonian selection in XPS}
  \label{section4.1}
  We consider selecting a better generative model, which is defined by the effective Hamiltonian, using the simplified 4f-electron-derived 3d core-level XPS spectra data of rare-earth insulating compounds.
  Let $\epsilon_L,\epsilon_f^0$ and $\epsilon_c$ be the energies of the conducting electrons of 4f rare-earth metals (5d, 6s electrons), the 4f
  electron, and the core electron, respectively. 
  We set the index $\nu$ ($\nu = 1,...,N_f,N_f = 14$) as the quantum number of the
  spin and f orbital. Moreover, let $V,U_{ff}$ and $-U_{fc}$ be the energies of the
  hybridization interaction between the 4f electrons and the conduction
electrons, the Coulomb interaction between the 4f electrons,
and the core-hole Coulomb potential for the 4f electrons,
respectively. We define $\Delta = \epsilon_f^0 - \epsilon_L$.
Let $M_2$ be a model using a two-state Hamiltonian $(H_2)$ and $M_3$ be a model using a three-state Hamiltonian $(H_3)$, and let $\mathcal{M} = \{M_2,M_3\}$ be the set of candidate models. \par
The two-state Hamiltonian $(H_2)$ is the effective Hamiltonian for the XPS spectrum of $\textup{La}_2\textup{O}_3$ and was proposed by Kotani and Toyozawa \cite{Kotani1974}. The Hamiltonian is given by 
\begin{align}
    H_2 = \epsilon_L \sum_{\nu = 1}^{N_f} a_{L\nu}^{\dag}a_{L\nu} + \epsilon_f^0\sum_{\nu=1}^{N_f}a_{f\nu}^{\dag}a_{f\nu} + \epsilon_c a_c^{\dag} a_c + \frac{V}{\sqrt{N_f}}\sum_{\nu = 1}^{N_f}(a_{L\nu}^{\dag}a_{f\nu} + a_{L\nu}a_{f\nu}^{\dag}) - U_{fc}\sum_{\nu = 1}^{N_f}a_{f\nu}^{\dag}a_{f\nu}(1-a_c^{\dag}a_c),
\end{align}
where $\ket{G}$ is the eigenstate of the minimum energy $(E_G)$ in the initial state and $\ket{F_j}(j=0,1)$ is the eigenstate of the two energy levels $(E_j(j=0,1))$ in the final state.
To compare the two models, we introduce the energy shift parameter $(b)$.   
Here, we set the parameter $\theta_{M_2} = \{\Delta, V, \Gamma, U_{fc},b\}$ and the modeling function as
\begin{align}
    f_{M_2}(x;\theta_{M_2}) = \sum_{j = 0}^1|\bra{F_j}a_c\ket{G}|^2 \frac{\Gamma/\pi}{(x - (E_j - E_g) - b)^2 + \Gamma^2}.
\end{align} 
\par
$H_3$ is the effective Hamiltonian for the XPS spectrum of $\textup{Ce}\textup{O}_2$ and was proposed by Kotani et al. \cite{Kotani1985}. The Hamiltonian is given by
\begin{align}
    H_3 = \epsilon_L \sum_{\nu = 1}^{N_f} a_{L\nu}^{\dag}a_{L\nu} + \epsilon_f^0\sum_{\nu=1}^{N_f}a_{f\nu}^{\dag}a_{f\nu} + \epsilon_c a_c^{\dag} a_c + \frac{V}{\sqrt{N_f}}\sum_{\nu = 1}^{N_f}(a_{L\nu}^{\dag}a_{f\nu} + a_{L\nu}a_{f\nu}^{\dag}) + U_{ff}\sum_{\nu > \nu'}a_{f\nu}^{\dag}a_{f\nu}a_{f\nu'}^{\dag}a_{f\nu'} - U_{fc}\sum_{\nu = 1}^{N_f}a_{f\nu}^{\dag}a_{f\nu}(1-a_c^{\dag}a_c),
\end{align}
where $\ket{G}$ is the eigenstate of the minimum energy $(E_G)$ in the initial state, and $\ket{F_j}(j=0,1,2)$ is the eigenstate of the three energy levels $(E_j(j=0,1,2))$ in the final state.
As in the case of $H_2$, we introduce the energy shift parameter $(b)$.   
Here, we set the parameter $\theta_{M_3} = \{\Delta, V, \Gamma, U_{fc},U_{ff}, b\}$ and the modeling function as follows:
\begin{align}
    f_{M_3}(x;\theta_{M_3}) = \sum_{j = 0}^2|\bra{F_j}a_c\ket{G}|^2 \frac{\Gamma/\pi}{(x - (E_j - E_g) - b)^2 + \Gamma^2}.
\end{align} 
\par
As in Sect. \ref{section3.1}, the probability distribution of the number of observed photons $(p(y|f_{M}(x;\theta_M)))$ is considered to be $\textup{Poisson}(y;f_{M}(x;\theta_M)\times T)$, with measurement time $T$.
The posterior probability distributions $(p(M_K|D),p(\theta_K|D,M_K))$ for data $D = \{(x_1,y_1),...,(x_N,y_N) \}$
can be calculated by the exchange Monte Carlo method \cite{Mototake2019}.
In this problem setting, our goal is to select better models from $M_2,M_3$ and to estimate the parameters of $\Delta, U_{fc}$ with high accuracy from experiments with a short total measurement time.
\subsection{Detailed algorithms in the Bayesian Hamiltonian selection}
\label{section4.2}
Unlike the case in Sect. \ref{section3.2}, the model set $(\mathcal{M} = \{M_2,M_3\})$ is fixed. 
The specific algorithm is shown in Algorithm \ref{Algorithm_Hamiltonian}.
\begin{figure}[h]
    \begin{algorithm}[H]
        \caption{Adaptive Experiment for Bayesian Hamiltonian Selection}
        \label{Algorithm_Hamiltonian}
        \begin{algorithmic}[1]
        \REQUIRE Number of measurement points per one experiment $n$, Number of experiments $k$, Measurement points set $\mathcal{X} = \{x_i\}_{i=1}^N$
        \ENSURE Data $D = \{(x_i,y_i)\}_{i = 1}^{N + n \times k} $
        \STATE Measure $y_1,...,y_N$ with $x_1,...,x_N$.
        \STATE Data $D = \{(x_i,y_i)\}_{i\in \{1,...,N\} } $
        \STATE $\mathcal{M} = \{M_2,M_3\}$
        \FOR{$i \in \{1,...,n\}$}
            \STATE $\widehat{M} = \text{argmax}_{M} P(M|D)$, $M^{\circ} = \text{argmin}_{M} P(M|D)$.
            \STATE Calculate estimation accuracies $G(x;\widehat{M}), G(x;M^{\circ})$.
            \STATE Select $\frac{n}{2}$ points $\{x_{next,1},...,x_{next,\frac{n}{2}}\}$ from $\mathcal{X}$ in descending order of $\{G(x_i;\widehat{M})\}_{x_i \in \mathcal{X}}$.
            \STATE Select $\frac{n}{2}$ points $\{x_{next,\frac{n}{2} + 1},...,x_{next,n}\}$ from $\mathcal{X}$ in descending order of $\{G(x_i;M^{\circ})\}_{x_i \in \mathcal{X}}$.
            \STATE Measure $\{y_{next,1},...,y_{next,n}\}$ in selected points $\{x_{next,1},...,x_{next,n}\}$
            \STATE $D = D \cup \{(x_{next,1},y_{next,1}),...,(x_{next,n},y_{next,n}) \}$
        \ENDFOR
        \end{algorithmic}
    \end{algorithm}
  \end{figure}
  \subsection{Results of the Bayesian Hamiltonian selection}
  \label{section4.3}
\begin{figure}[h]
    \centering
    \includegraphics[width = 5.0cm]{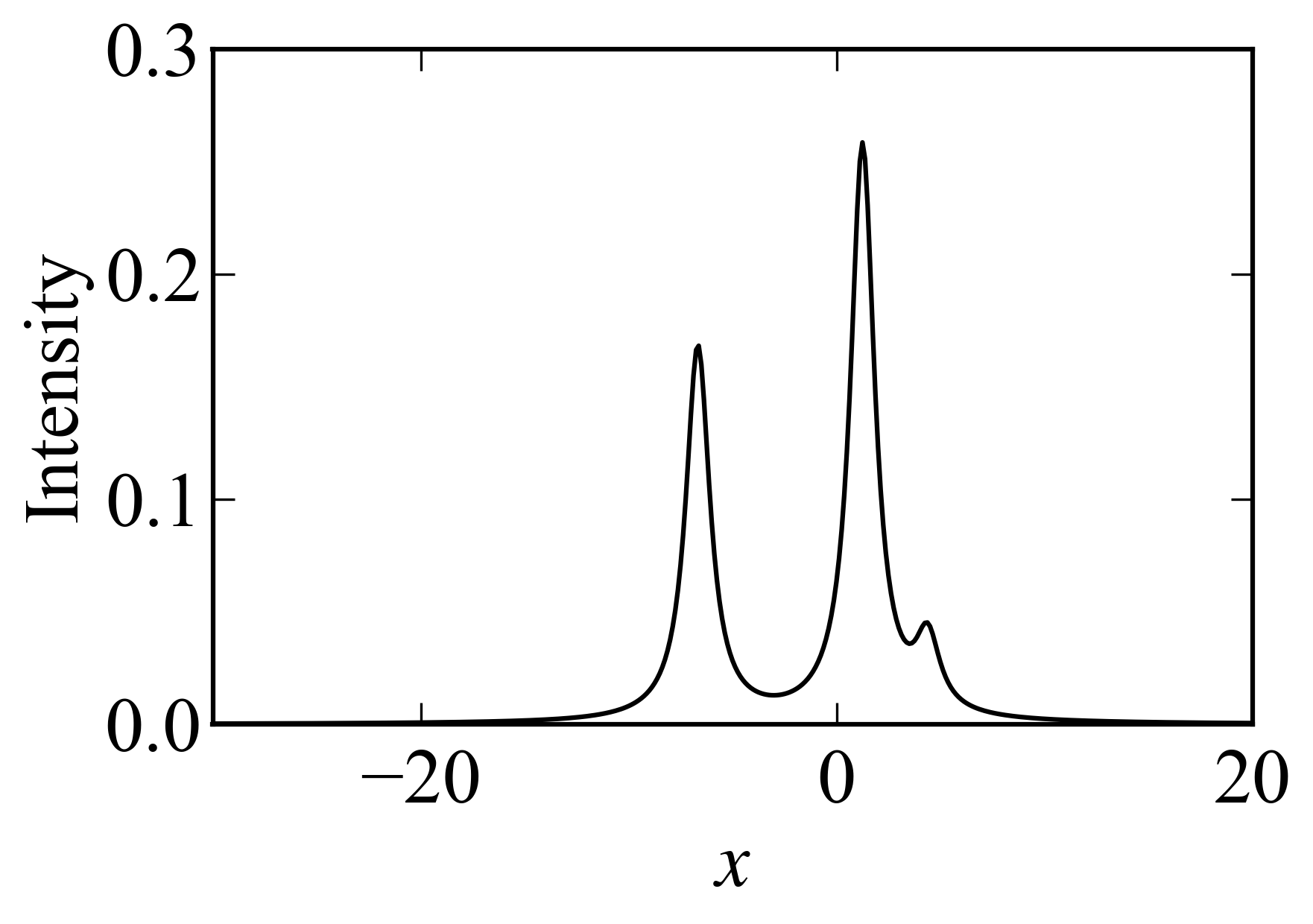}
    \caption{Plot of the modeling function $(f_{M_3}(x;\theta_{M_3}^*))$. The peak around $x=5$ is small, indicating that the model selection is difficult.}
    \label{True_Hamiltonian}
  \end{figure}

  Let the true model be the model $M_3$ with $H_3$ and the true values of its parameters be as follows:
  \begin{align}
    \Delta^* = 7.66, V^* = 0.76, U_{ff} = 10.5, U_{fc} = 12.7, \Gamma = 0.7 .
  \end{align}
  This true parameter is derived from Mototake et al. \cite{Mototake2019}.
  Although this parameter is different from the real parameter of $\ce{CeO_2}$, we set this parameter to complicate the model selection.
  The modeling function with the true parameter $(\theta_{M_3}^* = \{\Delta^*, V^*, \Gamma^*, U_{fc}^*,U_{ff}^*, b^*\})$ is shown in Fig. \ref{True_Hamiltonian}.
  The peak around $x=5$ is small, indicating that the model selection from model $M_2$ that generates two peaks and model $M_3$ that generates three peaks is difficult.
  In this situation, we set the prior distribution of the parameters as follows:
  \begin{align}
    \varphi(\Delta) &= U(0,20), \\
    \varphi(V) &= U(0,4), \\
    \varphi(U_{ff}) &= U(0,20), \\
    \varphi(U_{fc}) &= U(0,20), \\
    \varphi(\Gamma) &= U(0.01,1), \\
    \varphi(b) &= U(-5.0,5.0).
 \end{align} \par
 Let the prior distribution of the model set be $\varphi(M_2) = \varphi(M_3) = \frac{1}{2}$, the time for one measurement in the sequential experiment $T$ be $T=6$, the vertical resolution of the experiment be $0.125 (eV)$, 
 the number of data $N$ be $N=400$, and the candidate set of measurement points $\mathcal{X}$ be $\mathcal{X} = \{-30 + 0.125(i-1)\}_{i=1}^{400}$.
 First, we measure all points on $\mathcal{X}$ with a measurement time of $T=6$. Thereafter, we repeat the experiment $k = 160$ times by sequentially selecting $n = 10$ points to be measured next so that the total measurement time is $T_{sum} = N \times T + n \times k \times T = 12,000$.
 To evaluate the effectiveness of our method, we compare the result with that of a static experimental design in which the total measurement time is $T_{sum} = 12,000$, with $T = 30$ at all measurement points and that of an experiment in which the total measurement is $T_{sum} = 36,000$, with $T = 90$ at all measurement points.\par
\begin{figure}[h]
    \centering
        \includegraphics*[width = 5.0cm]{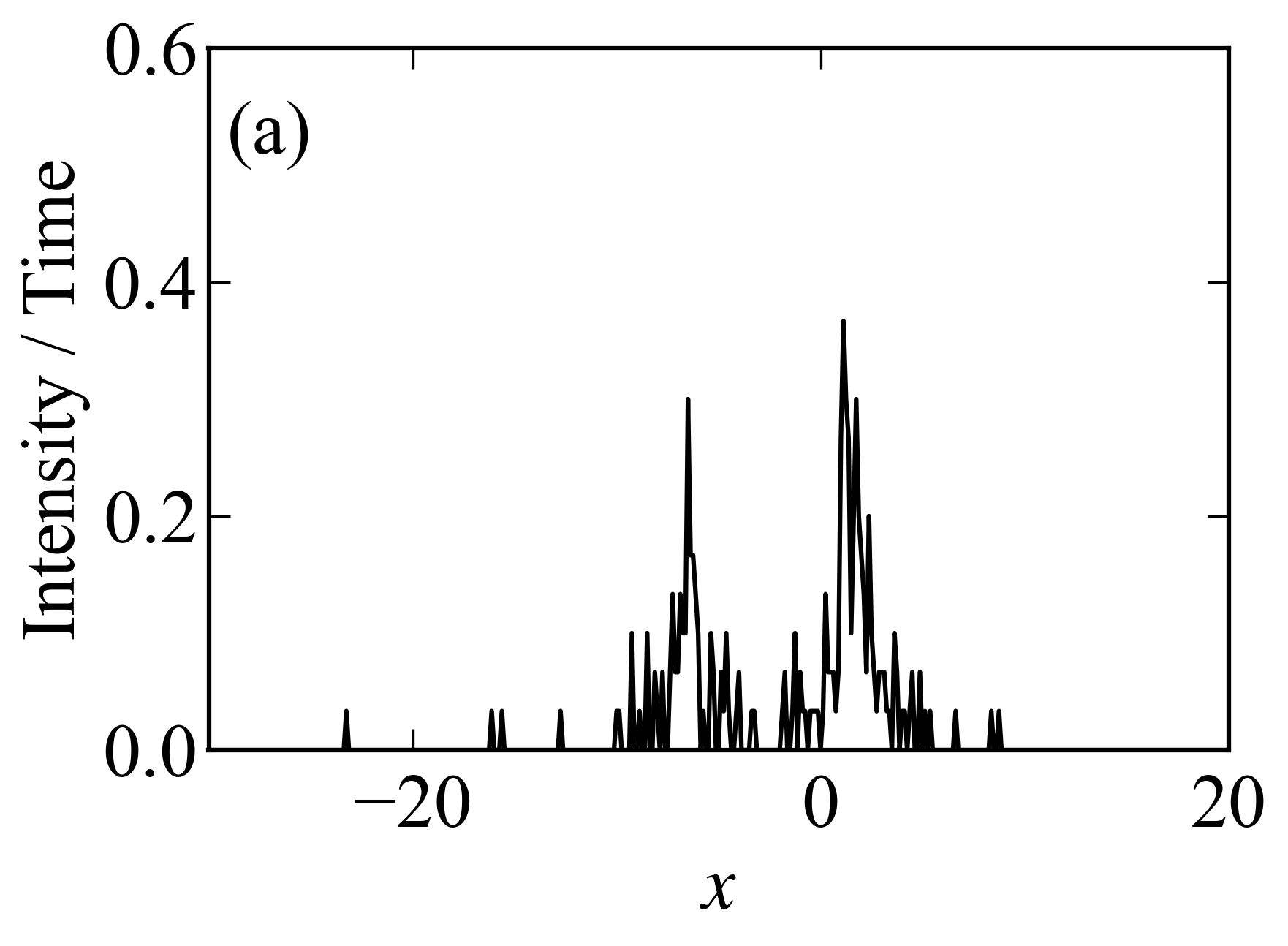}
        \includegraphics*[width = 5.0cm]{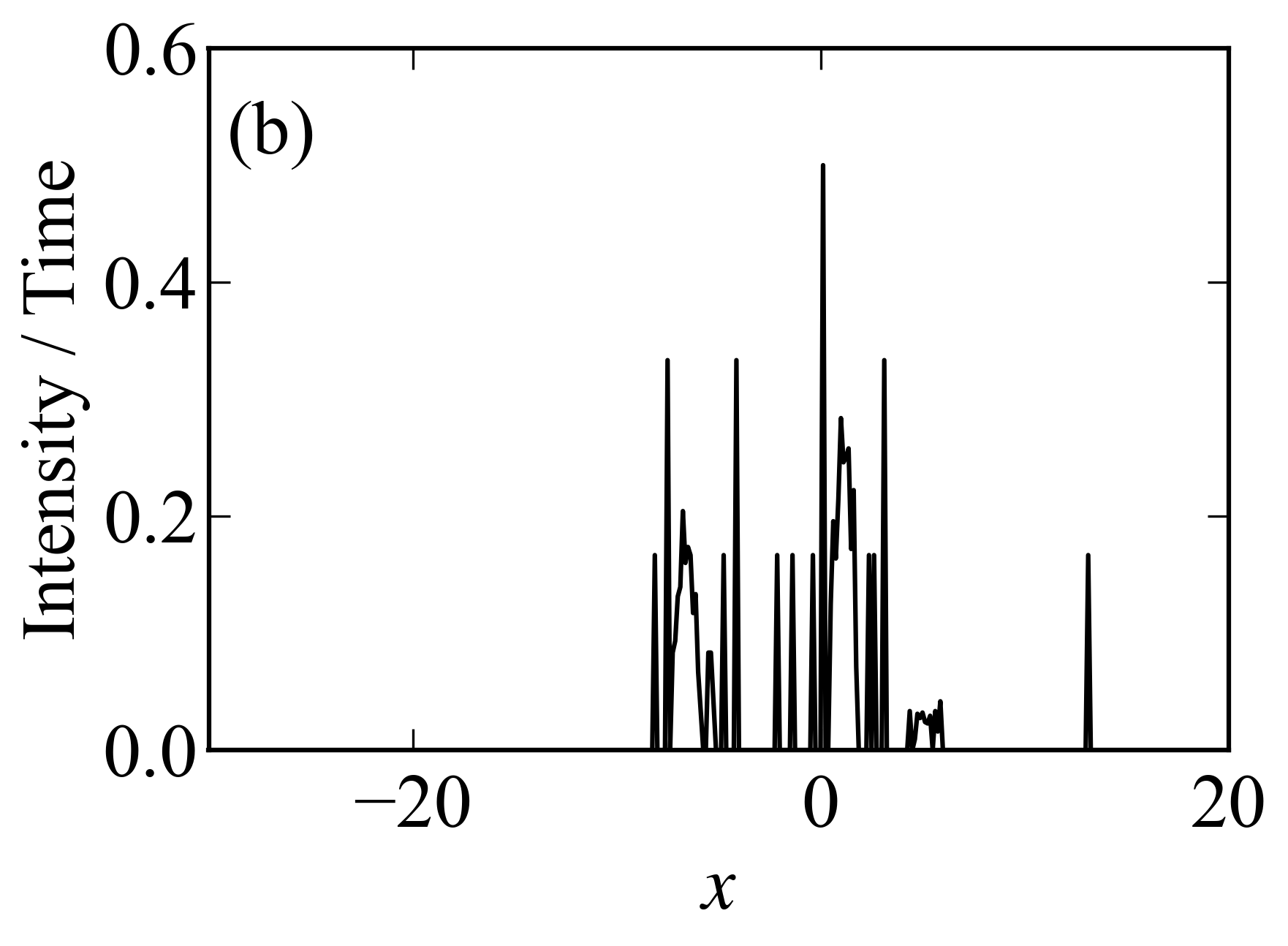}
        \includegraphics*[width = 5.0cm]{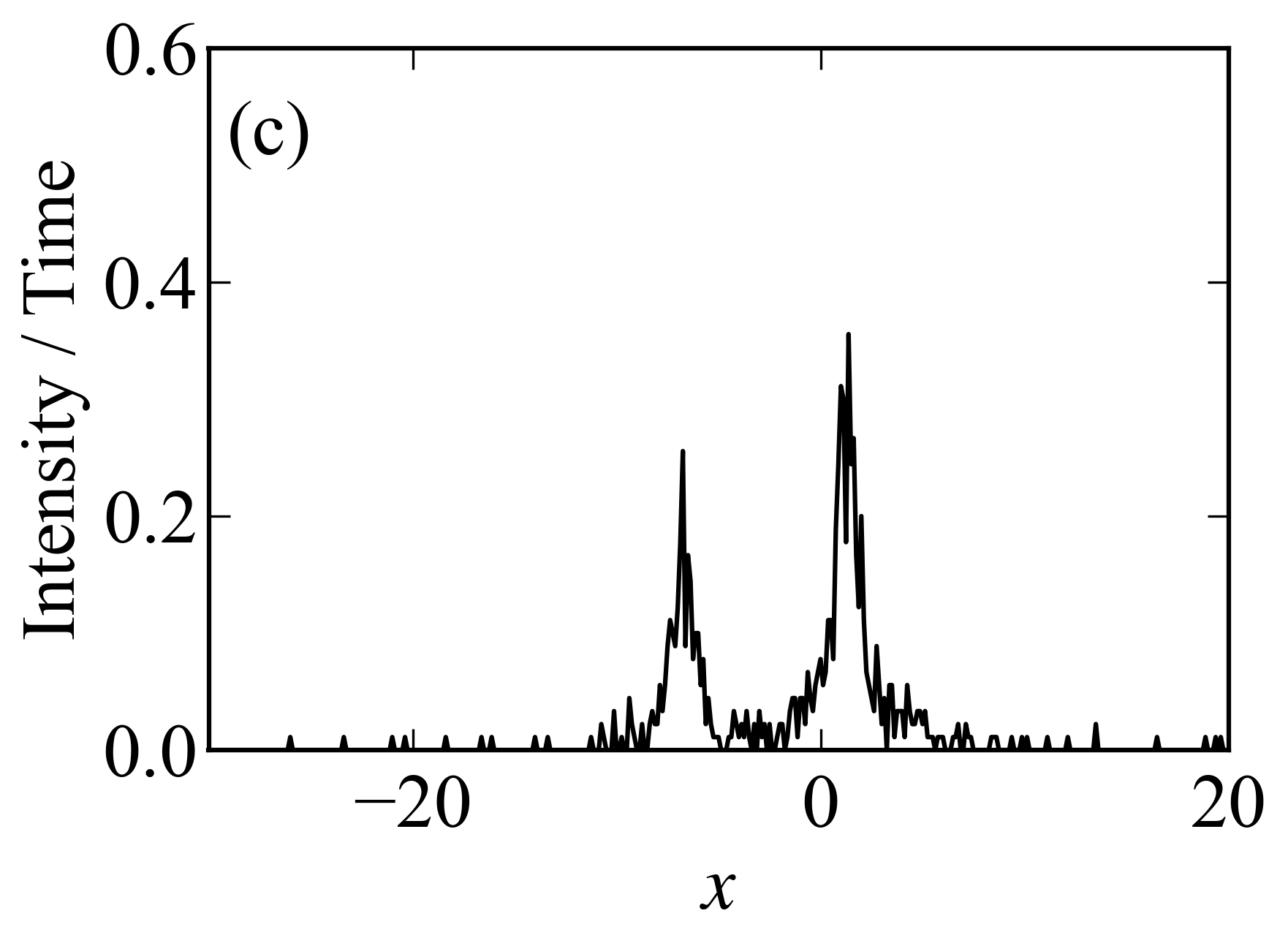} \\
        \includegraphics*[width = 5.0cm]{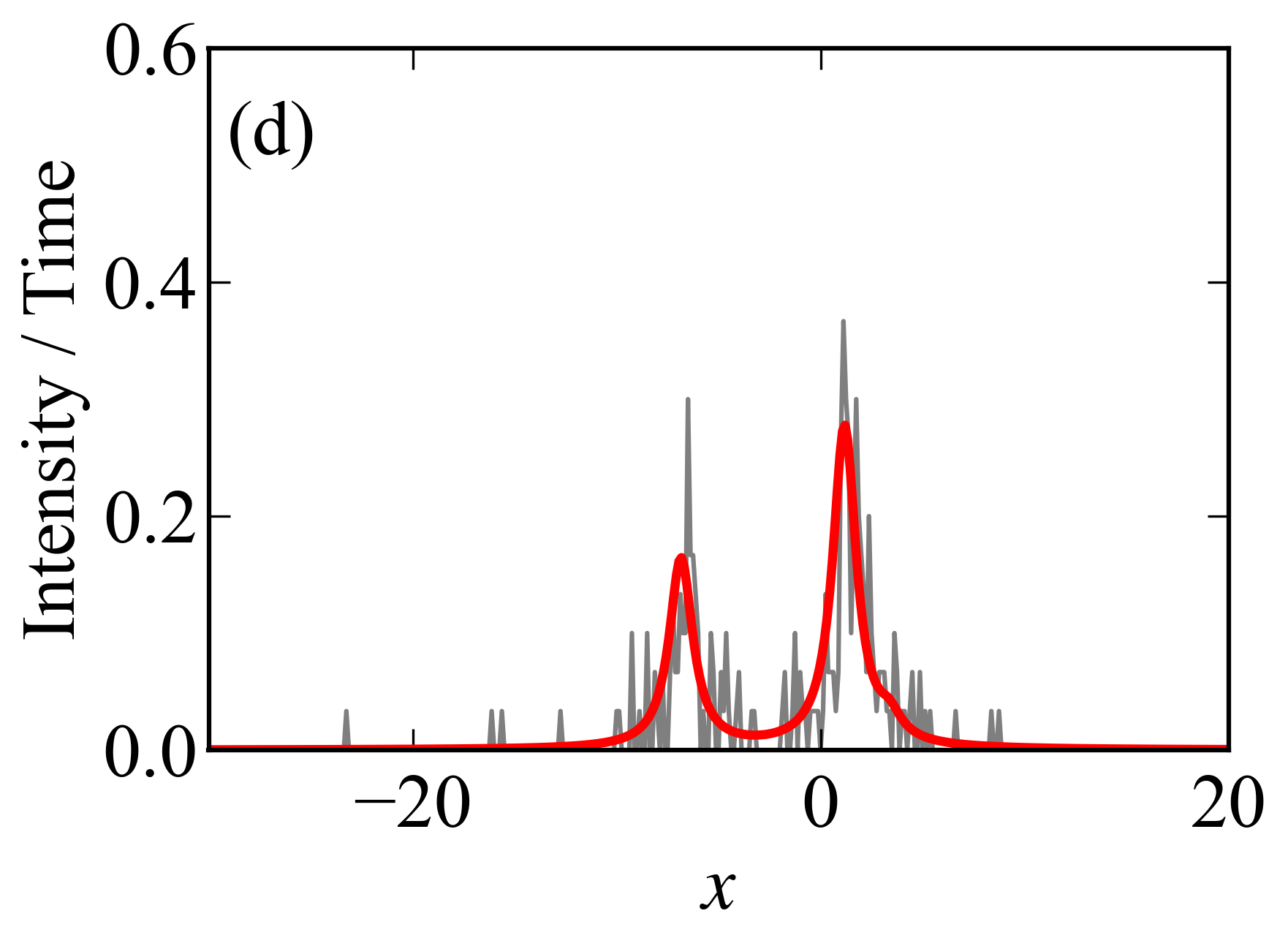}
        \includegraphics*[width = 5.0cm]{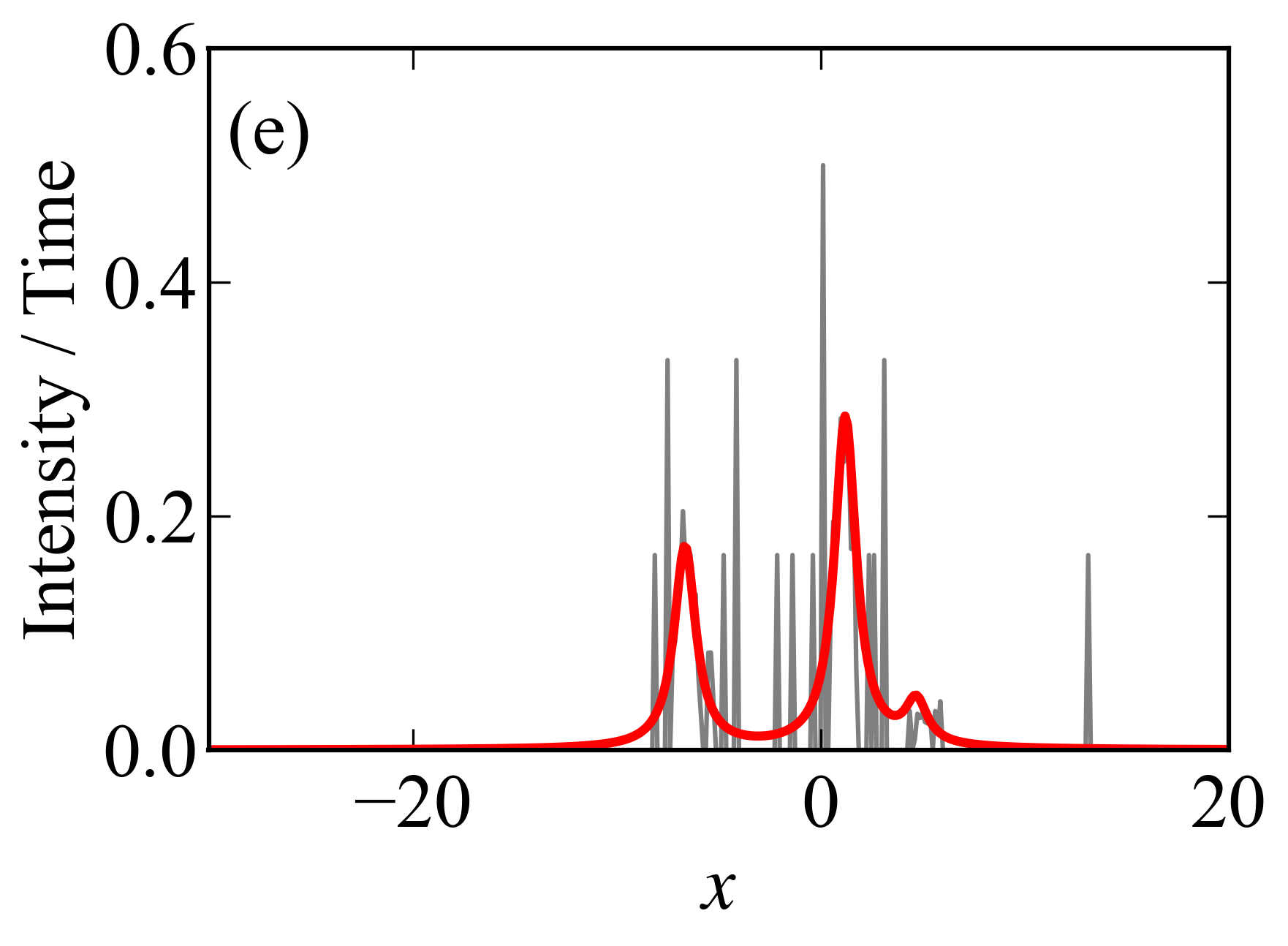}
        \includegraphics*[width = 5.0cm]{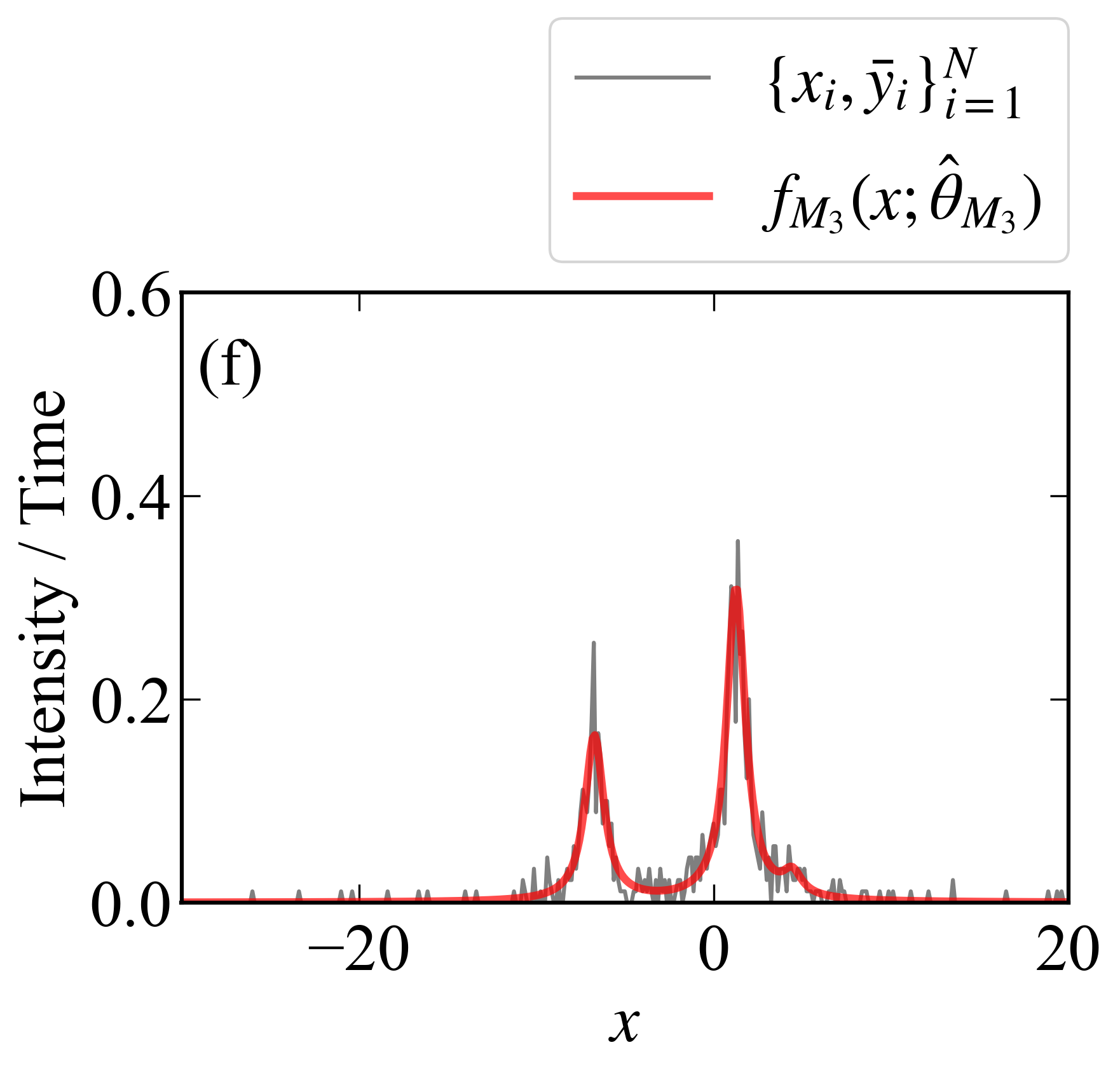} \\
        \includegraphics*[width = 5.0cm]{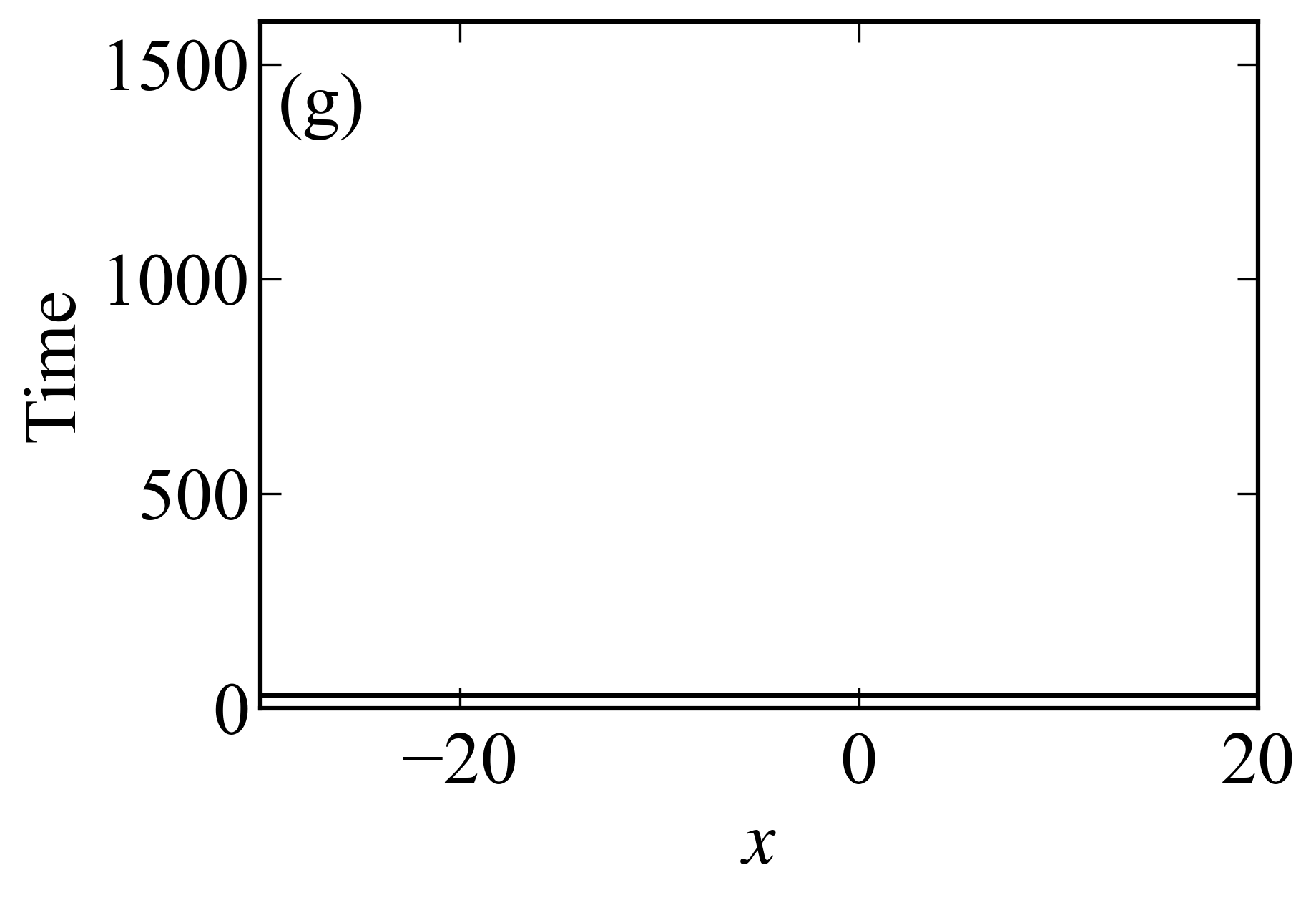}
        \includegraphics*[width = 5.0cm]{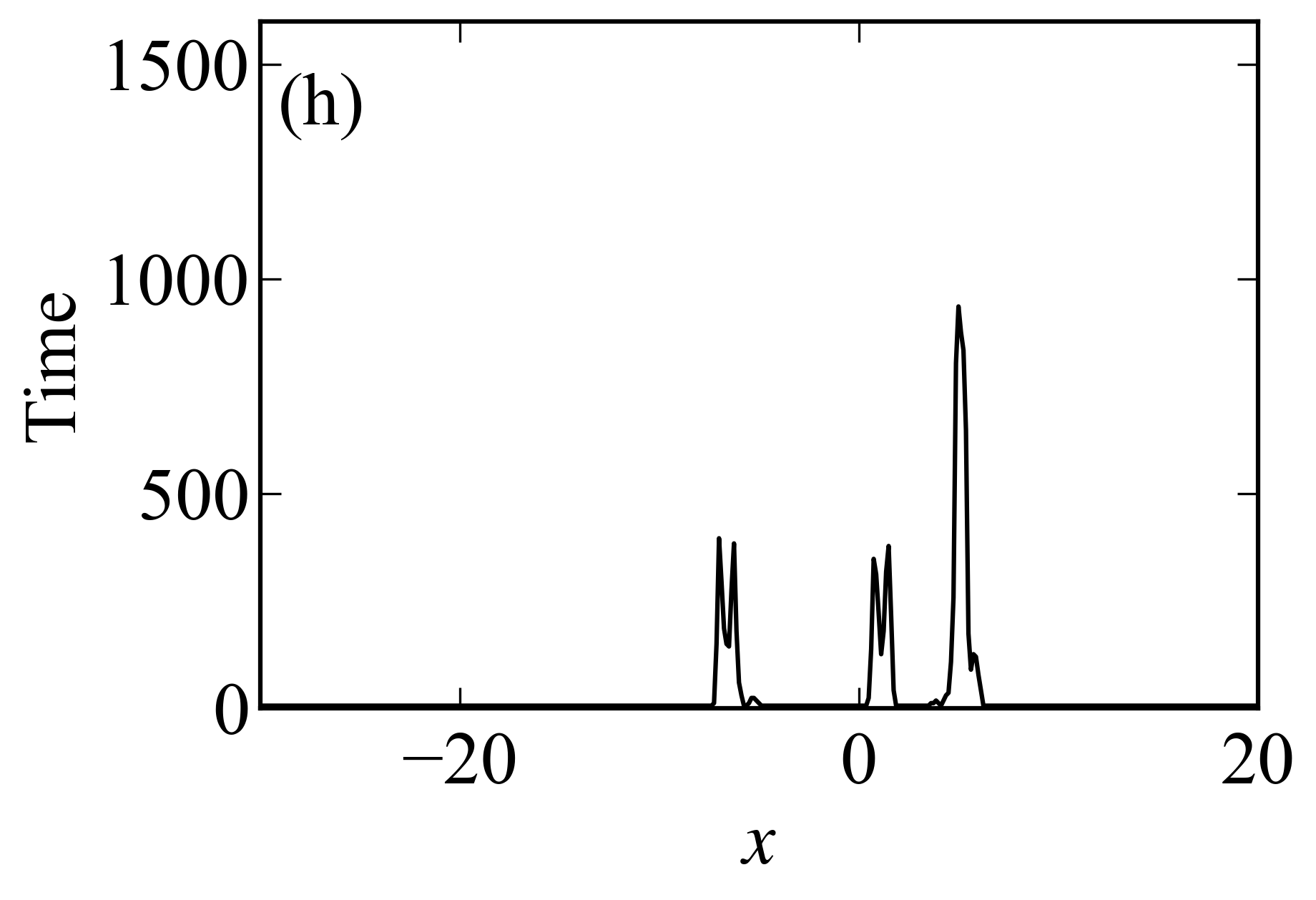}
        \includegraphics*[width = 5.0cm]{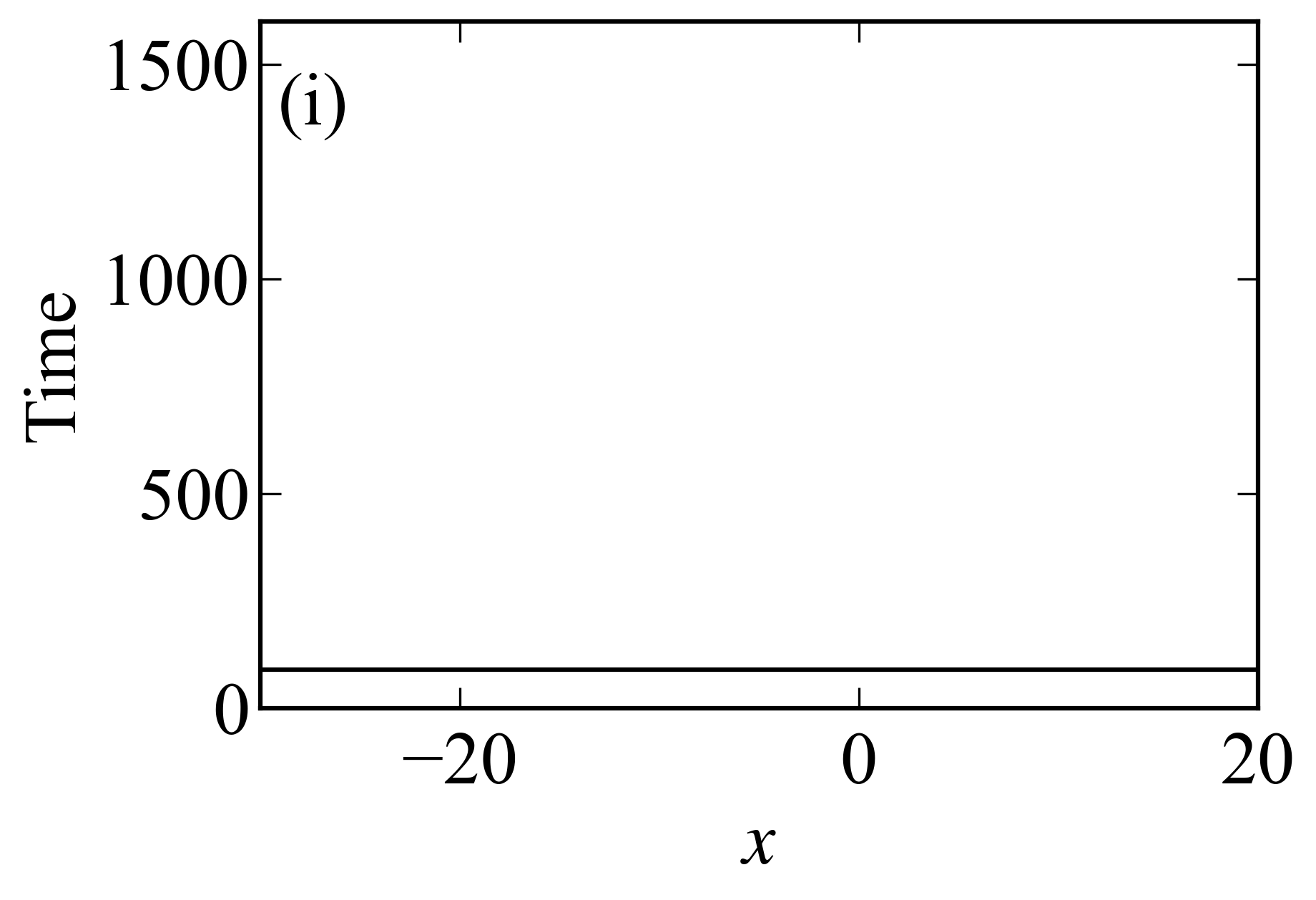} \\
        \caption{Data and fitting obtained by sequential experiments on the Bayesian Hamiltonian selection. The upper figure shows the number of photons observed per unit time $(\bar{y}_i)$. The middle figure shows the fitting by the map estimator $(\grave{\theta}_{M_3} = \underset{\theta}{\text{argmax}}p(\theta| D, M_3))$. The lower figure shows that the total measurement time per measurement point. 
        Panels (a),(d),(g) show the cases of a static experiment with the total measurement time of $T_{sum} = 12,000$. Panels (b),(e),(h) show the cases of a sequential experiment with a total measurement time $T_{sum} = 12,000$. Panels (c),(f),(i) show the cases of a static experiment with a total measurement time $T_{sum} = 36,000$.}
    \label{fitting_Hamiltonian}
\end{figure}
\begin{figure}[h]
    \centering
    \includegraphics*[width = 5.0cm]{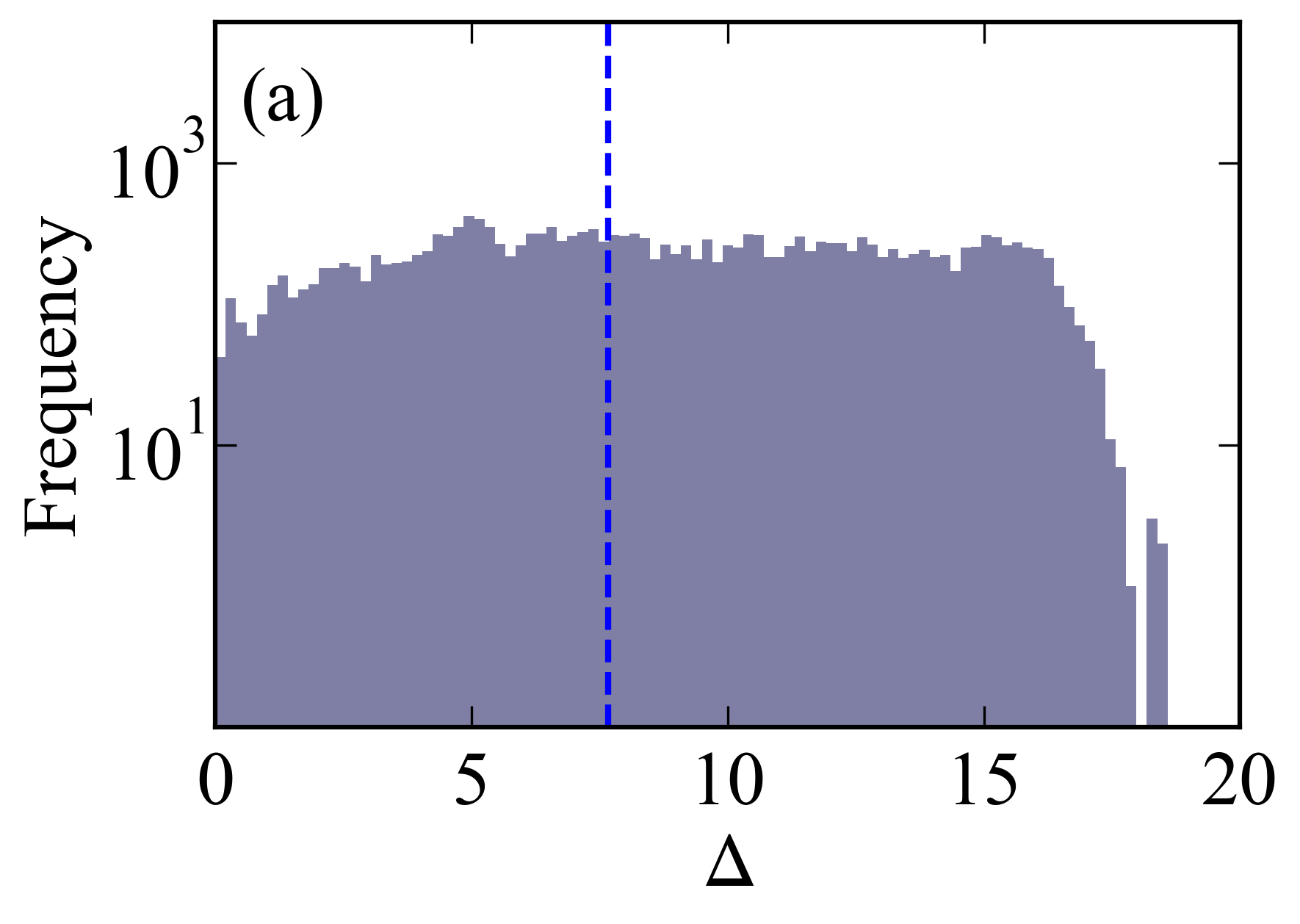}
    \includegraphics*[width = 5.0cm]{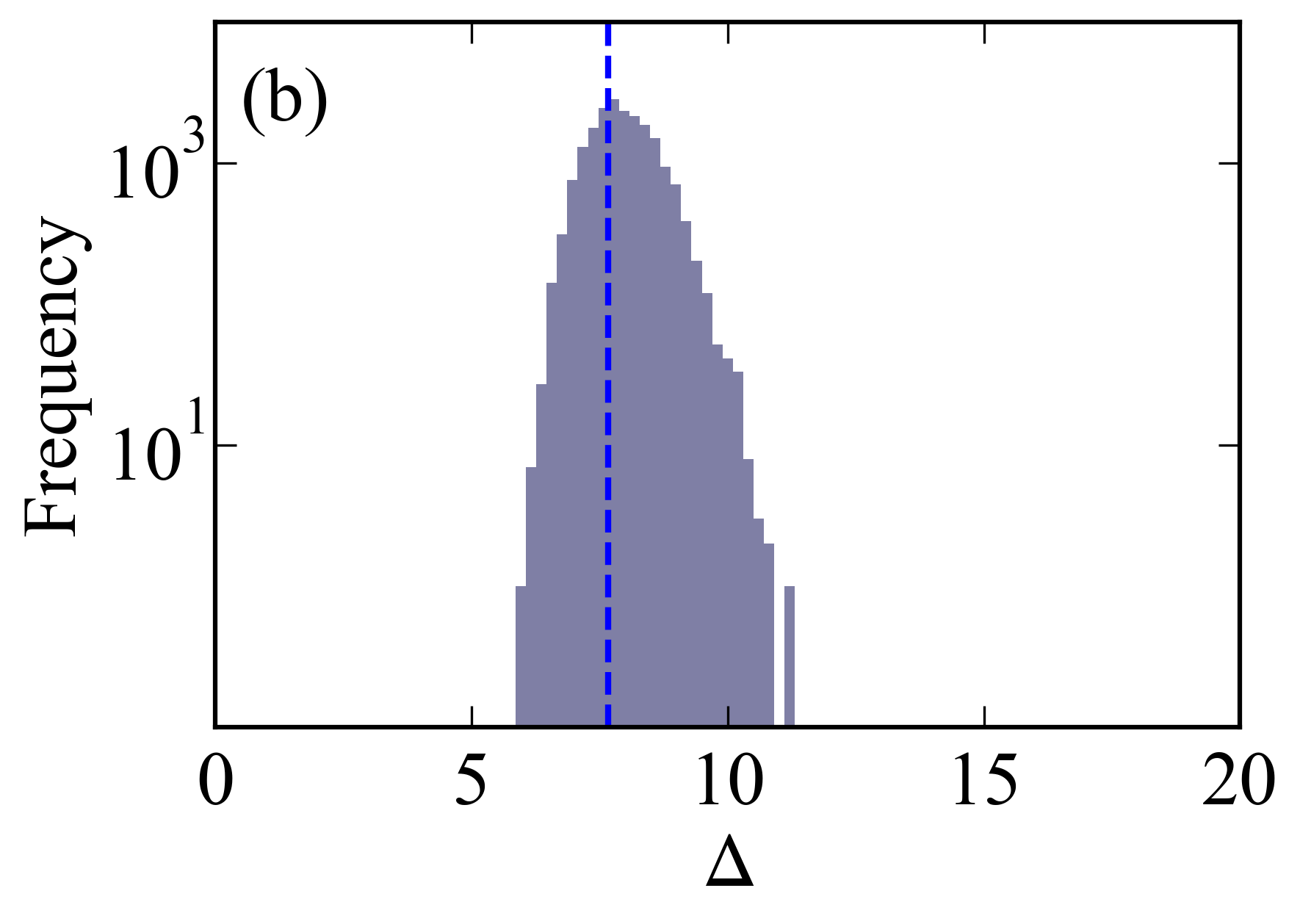}
    \includegraphics*[width = 5.0cm]{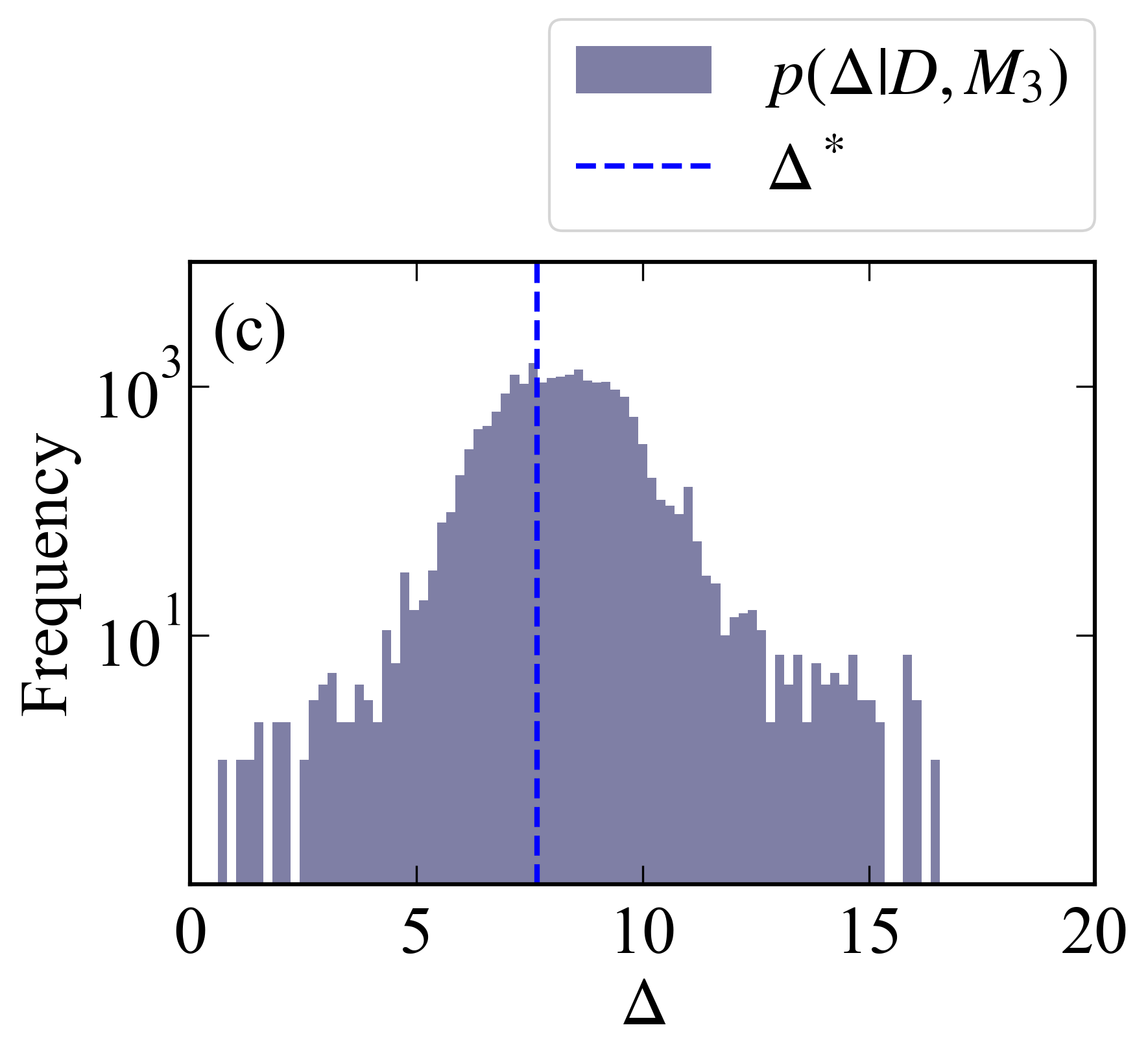} \\
    \includegraphics*[width = 5.0cm]{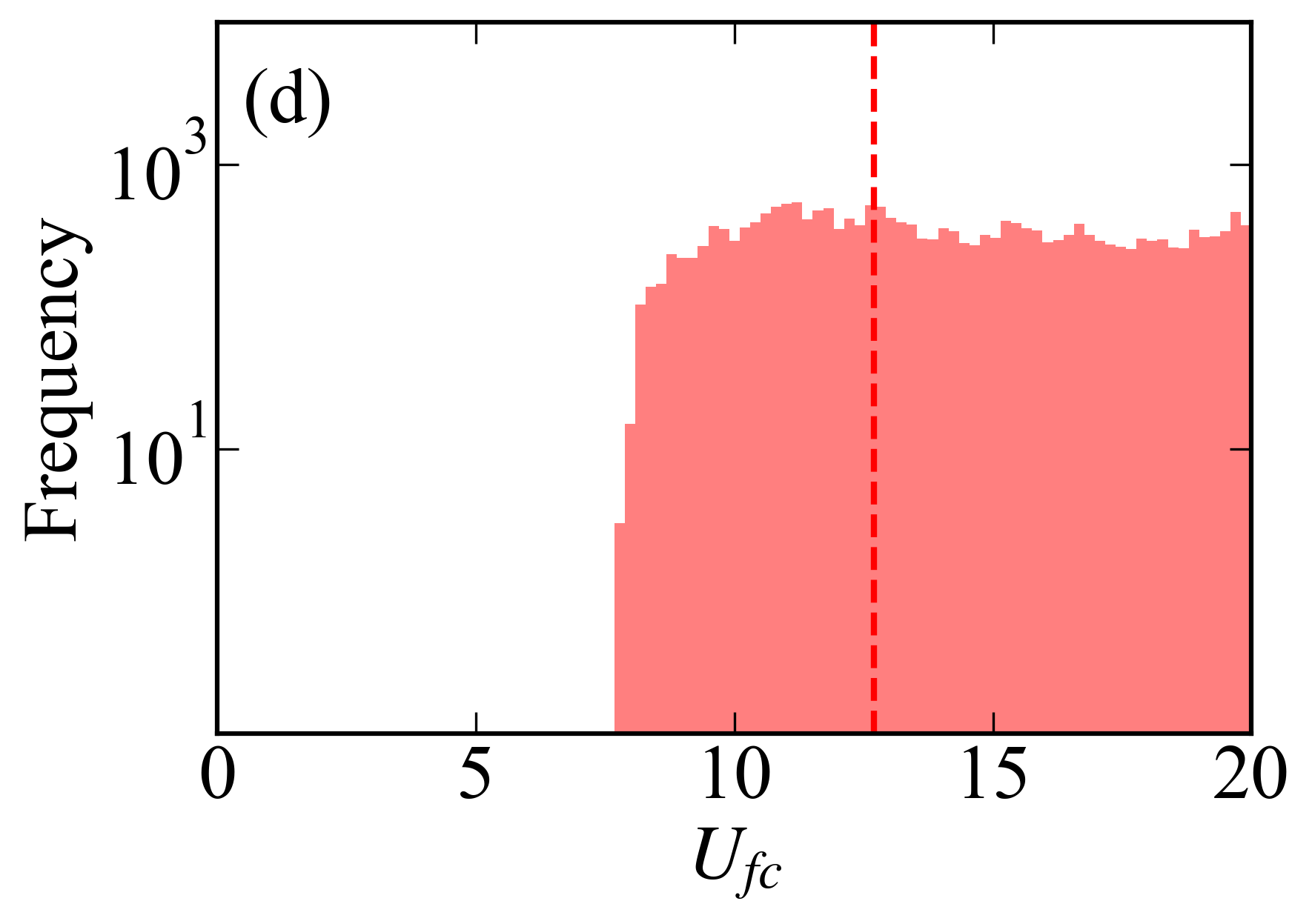}
    \includegraphics*[width = 5.0cm]{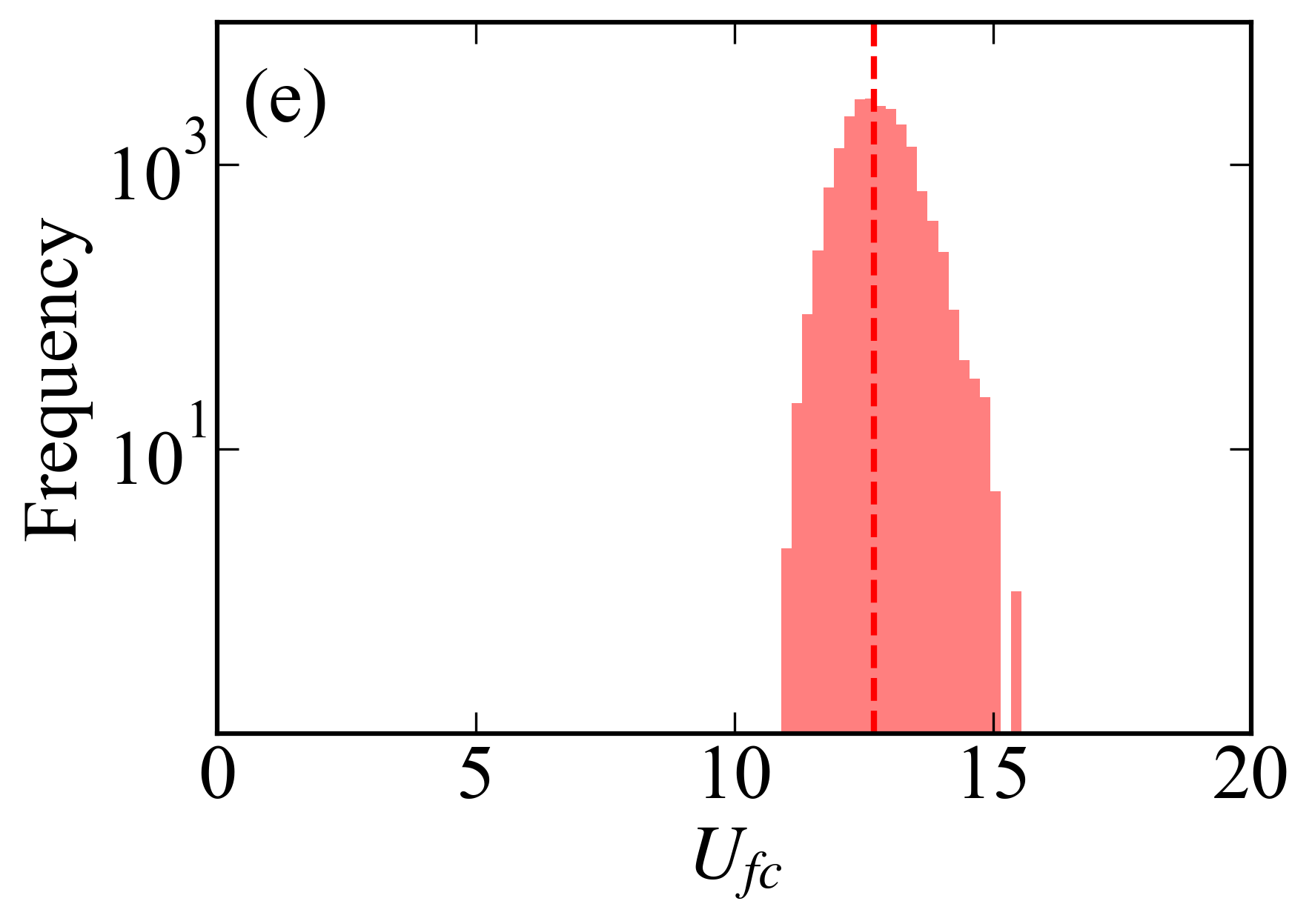}
    \includegraphics*[width = 5.0cm]{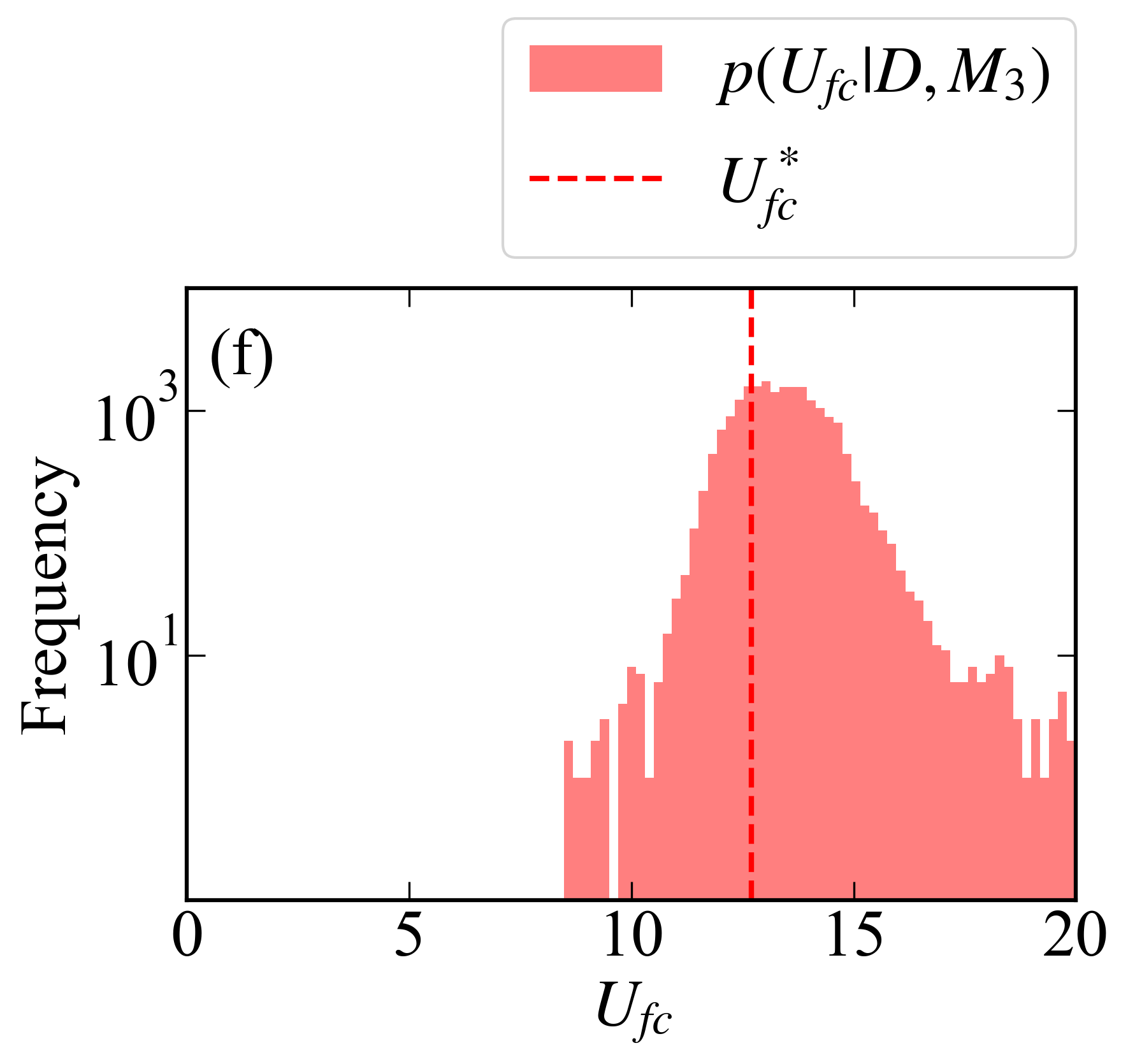}
    \caption{Parameter estimation by the posterior distribution. The upper figures show the posterior distributions of $\Delta$, and the lower figures show the posterior distribution of $U_{fc}$. The dashed line shows the true parameter values. Panel (a) shows the case of a static experiment with a total measurement time of $T_{sum} = 12,000$. Panel (b) shows the case of a sequential experiment with a total measurement time of $T_{sum} = 12,000$. Panel (c) shows the case of a static experiment with a total measurement time of $T_{sum} = 36,000$.
    }
    \label{parameter_Hamiltonian_delta}
  \end{figure}
\begin{figure}[h]
    \centering
    \includegraphics*[width = 12.0cm]{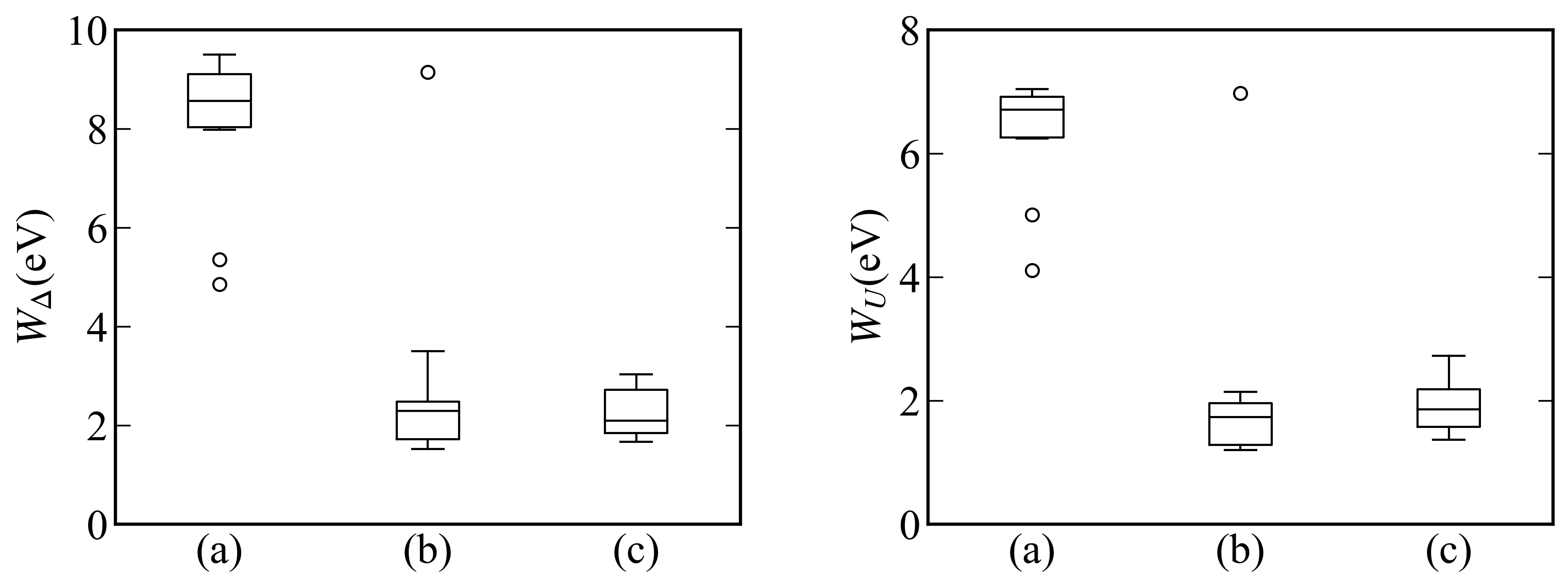}
    \caption{Boxplots representing the parameter estimation accuracy of $\Delta, U_{fc}$. Left and right panels show the boxplots of $W_{\Delta}$ and $W_{U}$, respectively. Label (a) highlights the case of a static experiment with a total measurement time of $T_{sum} = 12,000$. Label (b) highlights the case of a sequential experiment with a total measurement time of $T_{sum} = 12,000$. Label (c) highlights the cases of a static experiment with a total measurement time of $T_{sum} = 36,000$.}
    \label{Width_Hamiltonian}
  \end{figure}

 The data and the fitting by the map estimator $(\grave{\theta}_{M_3} = \underset{\theta}{\text{argmax}}p(\theta| D, M_3))$ are shown in Fig. \ref{fitting_Hamiltonian}. It can be observed that the area near the peaks, particularly near the small peak around $x=5$, is measured intensively. \\
In addition, we estimate the parameter of $\Delta, U_{fc}$, assuming that the true model is $M_3$. The results of the parameter estimation are shown in Fig. \ref{parameter_Hamiltonian_delta}. 
  Furthermore, we repeat the above trial 10 times independently to confirm the statistical property, similar to the case in Sect. \ref{section3}. We define indices to evaluate the width of the parameter estimation, as follows:
  \begin{align}
    W_{\Delta} = \max_{\alpha \in [0.025,0.975]}|\Delta^* - \Delta_{\alpha}|, W_{U} = \max_{\alpha \in [0.025,0.975]}|U_{fc}^* - U_{\alpha}|,
  \end{align}
  where 
  \begin{align}
    \Delta_{\alpha} &= \min_{\Delta'} \left\{ \left(\int_{\Delta < \Delta'}p(\Delta|D,K)\textup{d}\Gamma\right) > \alpha \right\}, \\
    U_{\alpha} &= \min_{U'} \left\{ \left(\int_{U_{fc} < U'}p(U_{fc}|D,K)\textup{d}U_{fc}\right) > \alpha \right\}. \\
  \end{align}
  These indices represent the deviations between the 95\% confidence intervals of the parameter estimation and the true parameters $(\Delta^*, U_{fc}^*)$.
  The boxplots of $W_{\Delta}, W_{U_{fc}}$ for the 10 trials are shown in Fig. \ref{Width_Hamiltonian}. The figure shows that the estimations by our method are more accurate than those by the static experiment with $T_{sum} = 12,000$ and are as accurate as those by the static experiment with $T_{sum} = 36,000$.\par
\begin{figure}[h]
    \centering
    \includegraphics*[width = 5.0cm]{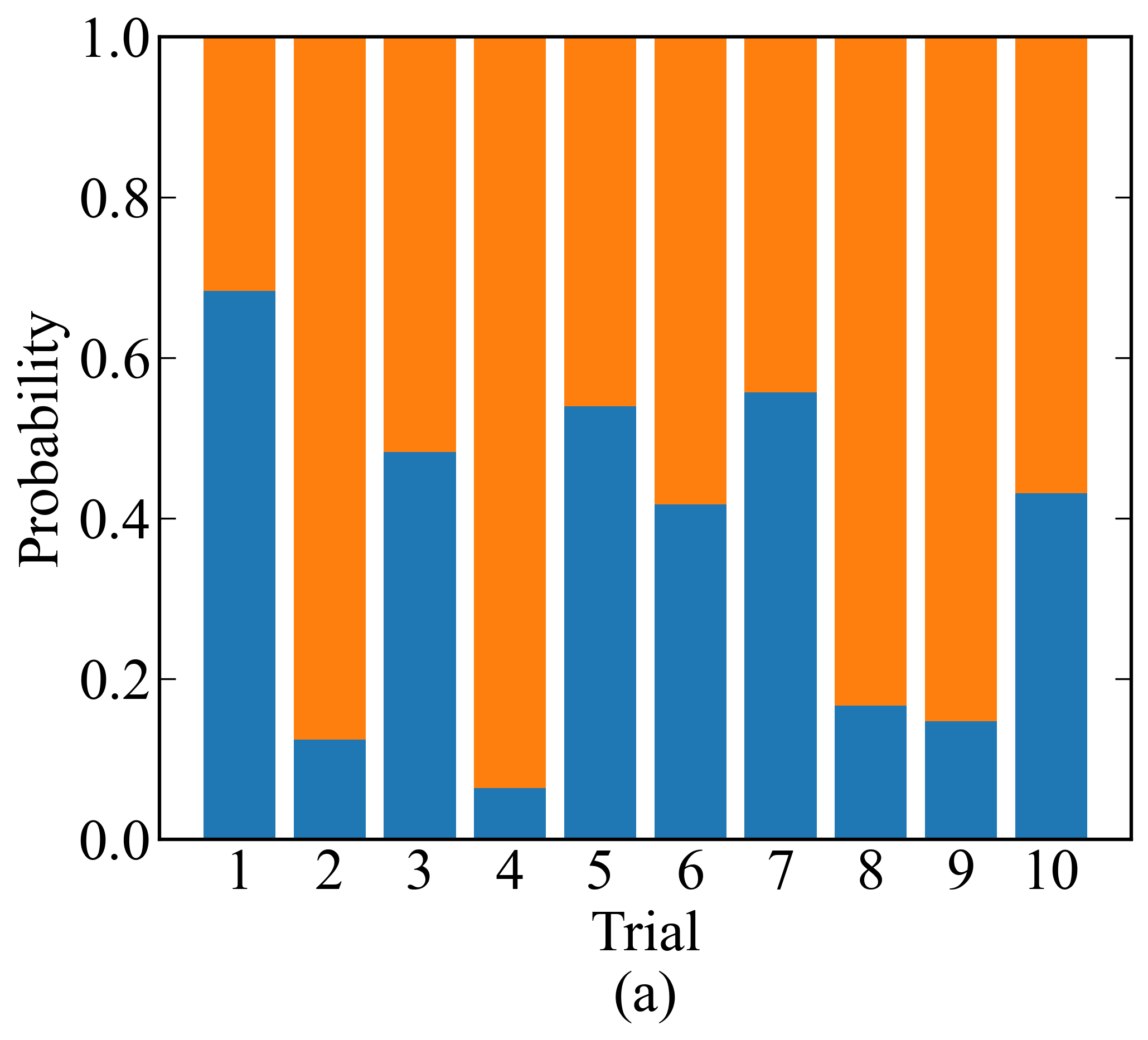}
    \includegraphics*[width = 5.0cm]{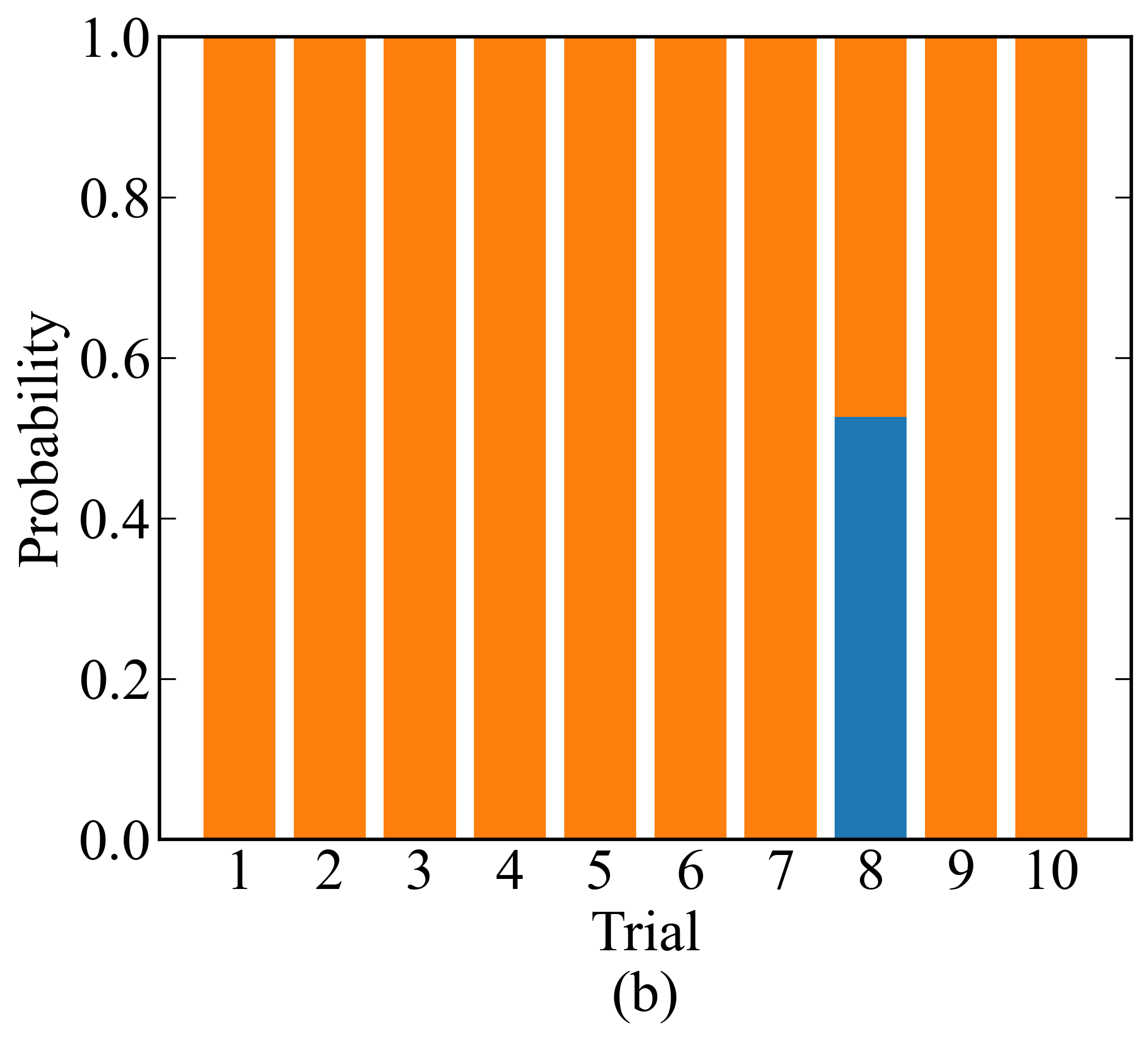}
    \includegraphics*[width = 5.0cm]{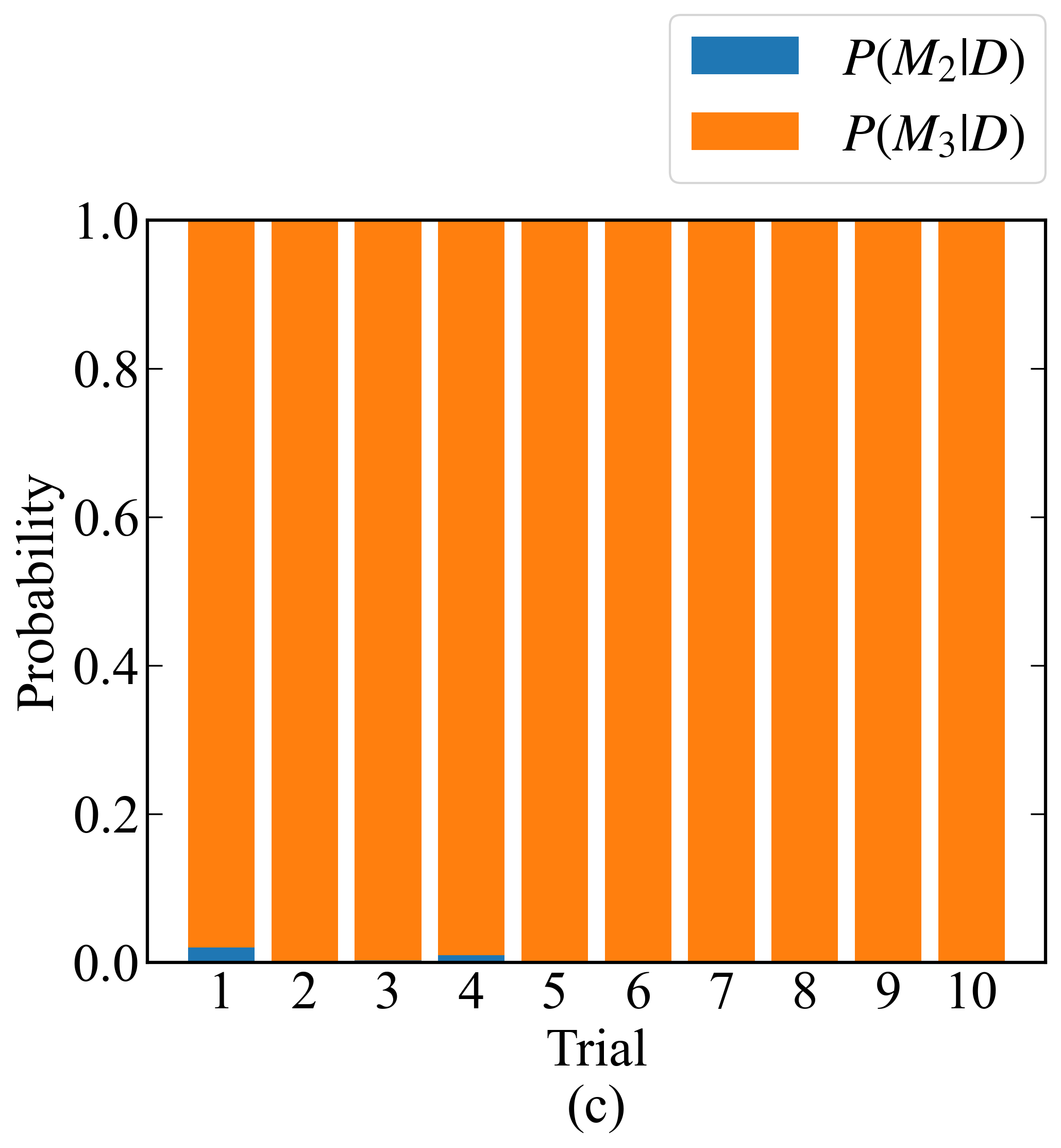}
    \caption{Bar graphs of $P(M_2|D), P(M_3|D)$ for the 10 independent trials. Panel (a) shows the case of a static experiment with a total
    measurement time of $T_{sum} = 12,000$. Panel (b) shows the case of a sequential experiment with a total measurement time of $T_{sum} = 12,000$. Panel (c) shows
    the case of a static experiment with a total measurement time $T_{sum} = 36,000$.}
    \label{Probability_Hamiltonian}
  \end{figure}

  Moreover, we calculate the posterior distribution $(P(M_2|D), P(M_3|D))$ for the 10 independent trials with $\mathcal{M} = \{M_2, M_3\}$.
The bar graphs of $P(M_2|D), P(M_3|D)$ are shown in Fig. \ref{Probability_Hamiltonian}. This figure shows that the model selection accuracy by our method
is higher than that by the static experiment with $T_{sum} = 12,000$ and is similar to that by the static experiment with
$T_{sum} = 36,000$.  These results indicate that the time required for the experiment has been reduced to one-third of the original.
\section{Conclusion and Future Work}
\label{section5}
  Here, we propose a sequential experimental design for spectral measurement using active learning with parametric models as the predictors and applied it to Bayesian spectral deconvolution and Bayesian Hamiltonian selection.
  Using artificial XPS data, we demonstrated that our method achieves accurate model selection and parameter estimation in a shorter time compared with the conventional method that uses equal measurement time for all measurement points.
  As discussed in Appendix A, the Gaussian process regression does not work well in the settings employed in this study, indicating the superiority of active learning using a parametric model over that using a nonparametric model.\par
  In this study, we evaluated the effectiveness of the proposed method in XPS. However, the method can be applied to any spectral measurement for which data analysis using Bayesian inference is effective.
  Data analysis using Bayesian inference has been proposed for various experiments, such as Moessbauer spectroscopy \cite{Moriguchi2022}, X-Ray absorption near edge structure \cite{Kashiwamura2022}, and NMR spectroscopy \cite{Ueda2023}, and it is expected that our method will accelerate all these experiments in the future.\par
  To apply our method to various experiments, increasing the computation speed is necessary. This is because we perform Bayesian inference is performed after each measurement using the exchange Monte Carlo method, which can take significant computation time depending on the modeling function.
  To increase the speed, we can consider using the Monte Carlo method, which specializes in parallel computation \cite{Hukushima2003,Barash2017}, and utilizing the previous inference results.\par
  Applying our method to actual experiments is also a task for future studies. In actual experiments, systematic errors (deviations between the assumed model and the measured data) may be significant.
  In that case, it is necessary to confirm the robustness of our method against systematic errors and possibly improve the algorithm's robustness against systematic errors.\par
  The active learning approach with parametric models proposed in this paper is also applicable to the selection of data points from those already obtained.
  In the fitting process, the use of many data points increases the number of calculations of the modeling function in the fitting algorithm, resulting in a large computation time.
  For example, in the XPS data analysis, simulators, such as SESSA \cite{Smekal2005} and DFT calculations \cite{Paul2013}, are occasionally employed; however they are time-consuming.
  Further, in NMR spectroscopy, various models based on quantum chemistry have been proposed \cite{Helgaker2008}; however, some models require a large computation time.
  In these cases, reducing the data volume used for fitting is required. Active learning using parametric models can be adopted to sequentially increase the number of data points from a state with zero data points to achieve accurate analysis with a small number of data points.
  \section{Acknowledgments}
  This work was supported by JST, CREST (Grant Number JPMJCR1761 and JPMJCR1861), Japan.
  \appendix
  \section{Active learning with Gaussian Process}
  \label{AppendixA}
  In this section, we describe a method to realize the sequential experimental design in Sect. \ref{section2} with Gaussian process regression and compare its effectiveness with that of our method.
  In Appendix A.\ref{AppendixA.1}, we describe the method for selecting the next measurement point using Gaussian process regression. In Appendix A.\ref{AppendixA.2}, we describe the results applied to the Bayesian spectral deconvolution, and in Appendix A.\ref{AppendixA.3}, we describe the results applied to the Bayesian Hamiltonian selection.
  \subsection{Selection of the next measurement points by the Gaussian process regression}
  \label{AppendixA.1}
  Let us assume that for an input $x \in \mathcal{X}$, the response, $y$, can be written as $y = f(x) + \varepsilon \ (\varepsilon \sim \mathcal{N}(0,\xi^2))$ using the modeling function $(f)$, which follows the Gaussian process.
  For the input, $x_1,...,x_s \in \mathcal{X}$, $f(x_1),...,f(x_s)$ follows a Gaussian distribution with mean $\mu\bm{1}$, where $\mu = \frac{\sum_{i=1}^s{y_i}}{s}$, and the covariance matrix $K = \{k(x_{n},x_{n'})\}_{n,n'}$, where the kernel function $k(x,x')$ is a Gaussian kernel given by 
  \begin{align}
    k(x,x') = \theta_1\exp\left(- \left(\frac{x-x'}{\theta_2}\right)^2 \right).
  \end{align}
  Given data $D = \{ (x_1,y_1),...,(x_N,y_N)\}$, the mean $(\grave{\mu}(x))$ and the covariance $(\grave{\sigma}^2(x))$ of the prior distribution of $f(x)$ is given as follows \cite{Ueno2021}:
  \begin{align}
    \grave{\mu}(x) = \mu_0 + \bm{k}_N(x)^\top(\bm{K} + \xi^2\bm{I}_n)^{-1}(\bm{y} - \mu_0\bm{1}), \\
    \grave{\sigma}^2(x) = k(x,x) - \bm{k}_N(x)^\top(\bm{K} + \xi^2\bm{I}_N)^{-1}\bm{k}_N(x),
  \end{align}
  where $\mu_0 = \frac{\sum_{i=1}^N y_i}{N}$ and $\bm{k}_N(x) = (k(x_1,x),...,k(x_N,x))$.
  Here, hyperparameters $\theta_1,\theta_2$, and $\xi$ are determined to maximize the likelihood $p(\bm{y}|\bm{x},\theta_1,\theta_2,\xi)$, where $\bm{x} = (x_1,...,x_N)$ and $\bm{y} = (y_1,...,y_N)$.
  We use $\grave{\sigma}^2(x)$ in the sequential experimental design in Sect. \ref{section2}, i.e., we select $n$ measurement points from $x\in\mathcal{X}$ with a large $\grave{\sigma}^2(x)$ as the next measurement points.
  In Appendices A.2 and A.3, we implement the Gaussian process using the GPy Python package \cite{Gpy}.
  \subsection{Bayesian spectral deconvolution with the Gaussian process regression}
  \label{AppendixA.2}
\begin{figure}[h]
    \centering
    \includegraphics*[width = 12.0cm]{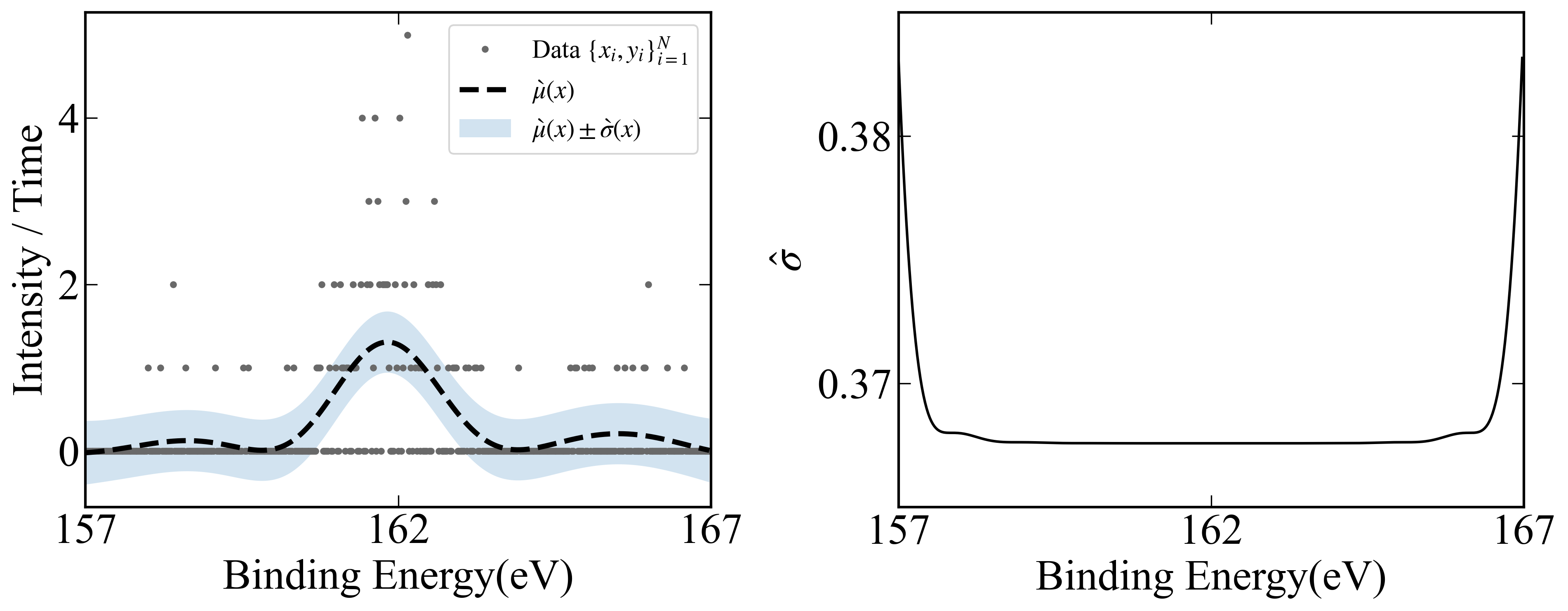}
    \caption{Example of the Gaussian process regression applied to the Bayesian spectral deconvolution. The left panel shows the estimation results by the Gaussian process regression, and the right panel shows the width of the confidence intervals. The width of the confidence intervals is large at both ends.}
    \label{Estimation_Gauss_Deconvolution}
  \end{figure}

  In the same problem setting as in Sect. \ref{section3}, we performed a sequential experimental design by active learning using Gaussian process regression.
  The posterior mean $(\grave{\mu})$ and the variance $(\grave{\sigma})$ for the data observed at all measurement points with a measurement time of $T=1$ are shown in Figure \ref{Estimation_Gauss_Deconvolution}.
  The maximum values are taken at both ends, which are not important for the analysis of the data. This is because the noise is so large that the Gaussian process regression cannot be properly performed.
  As in Sect. \ref{section3.3}, let the time for one measurement in the sequential experiment $(T)$ be $T = 1$, the vertical resolution of the experiment be $0.025\ \textup{(eV)}$, the number of data $(N)$ be $N = 400$, and 
  the candidate set of measurement points $\mathcal{X}$ be $\mathcal{X} = \{157 + 0.025(i-1)\ \textup{(eV)}\}_{i=1}^{400}$.
  Thus, we repeat the experiment $k=160$ times by sequentially selecting $n=10$ points to be measured next
so that the total measurement time is $T_{sum} = N\times T + n\times k \times T = 2000$.  \par
\begin{figure}
    \centering
    \includegraphics*[width = 12.0cm]{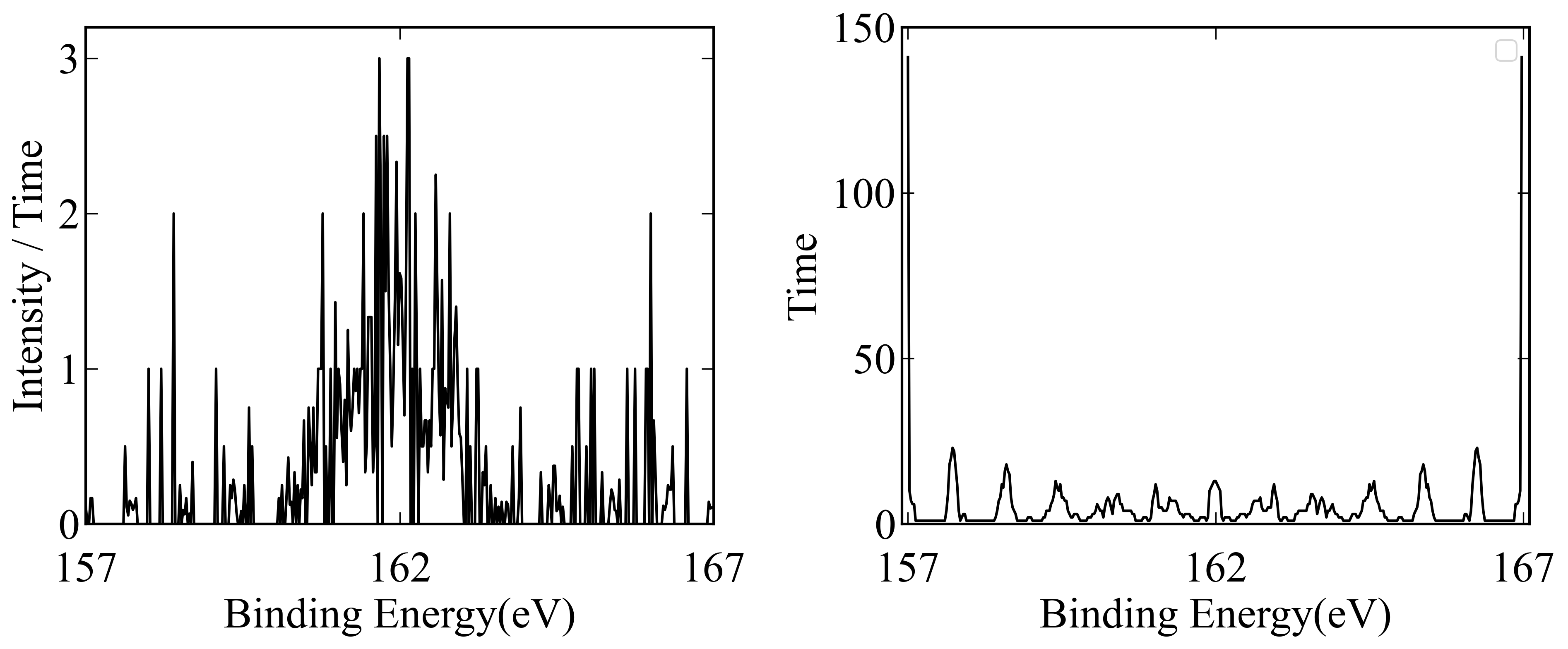}
    \caption{Data obtained in the sequential experiments using the Gaussian process regression. The left panel shows the number of observed photons per unit time, $\{x_i,\bar{y}_i\}_{i=1}^N$.  The right panel shows the measurement time for each measurement point.  The measurement time is larger at the edges that are not important for the Bayesian spectral deconvolution.}
    \label{Data_Gauss_Deconvolution}
\end{figure}

\begin{figure}[h]
    \centering
    \includegraphics*[width = 6.0cm]{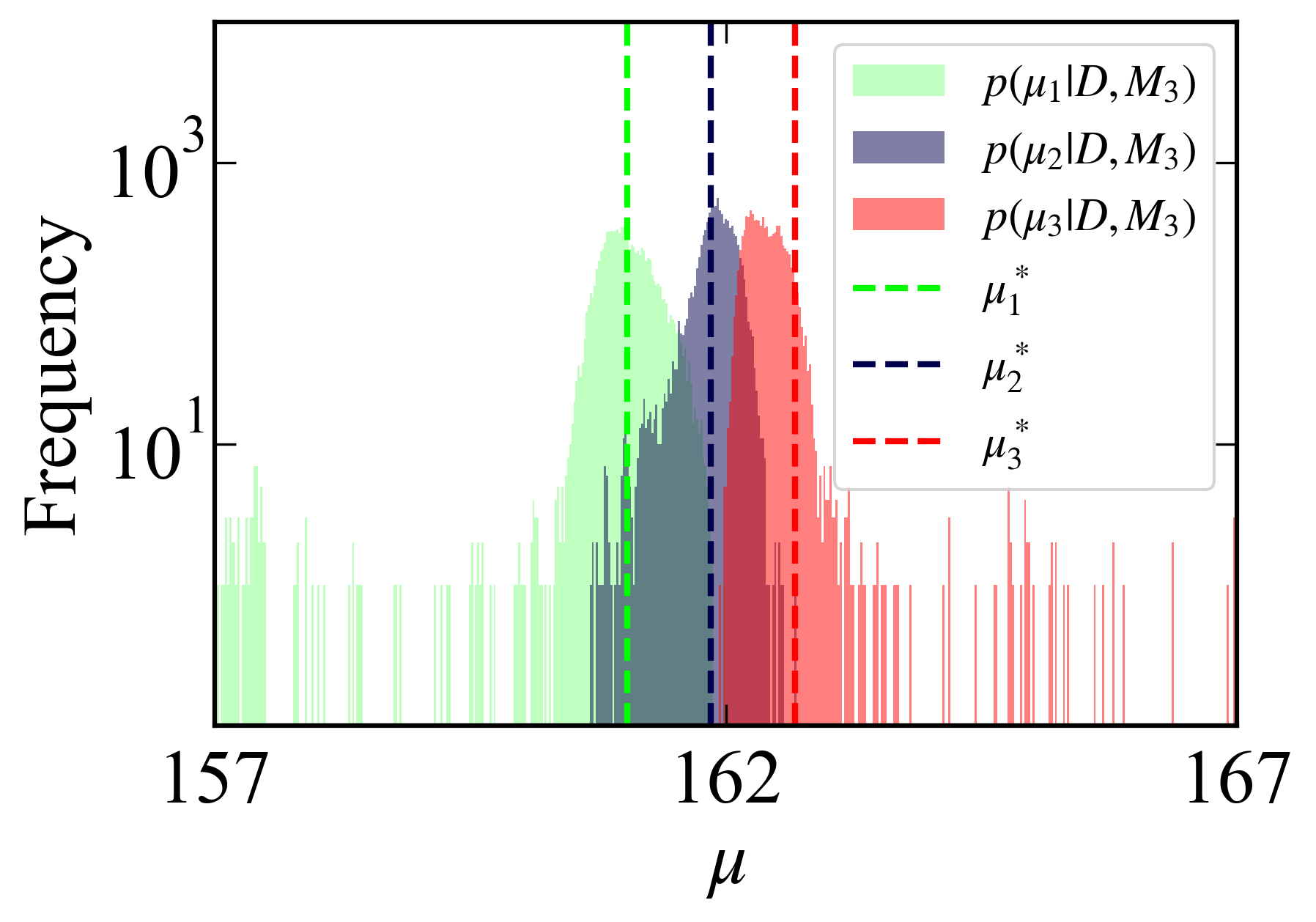}
    \caption{Parameter estimation by the posterior distribution $(p(\mu_1,\mu_2,\mu_3|D,M_3))$ of peak positions $\mu_1,\mu_2$, and $\mu_3$. The dashed line shows the true parameter values.}
    \label{Data_Gauss_parameter}
\end{figure}

 
The obtained data are shown in Fig. \ref{Data_Gauss_Deconvolution}. The measurements are focused on the range without peaks, which is not important in the spectral deconvolution.
Furthermore, the parameter estimation of $\mu_1,\mu_2$, and $\mu_3$ by the posterior distribution for this data is as shown in Fig. \ref{Data_Gauss_parameter}.
The estimation accuracy of our method is better than that shown in Fig. \ref{parameter_deconvolution}.
  \subsection{Bayesian Hamiltonian selection with Gaussian process regression}
  \label{AppendixA.3}
\begin{figure}[h]
    \centering
    \includegraphics*[width = 12.0cm]{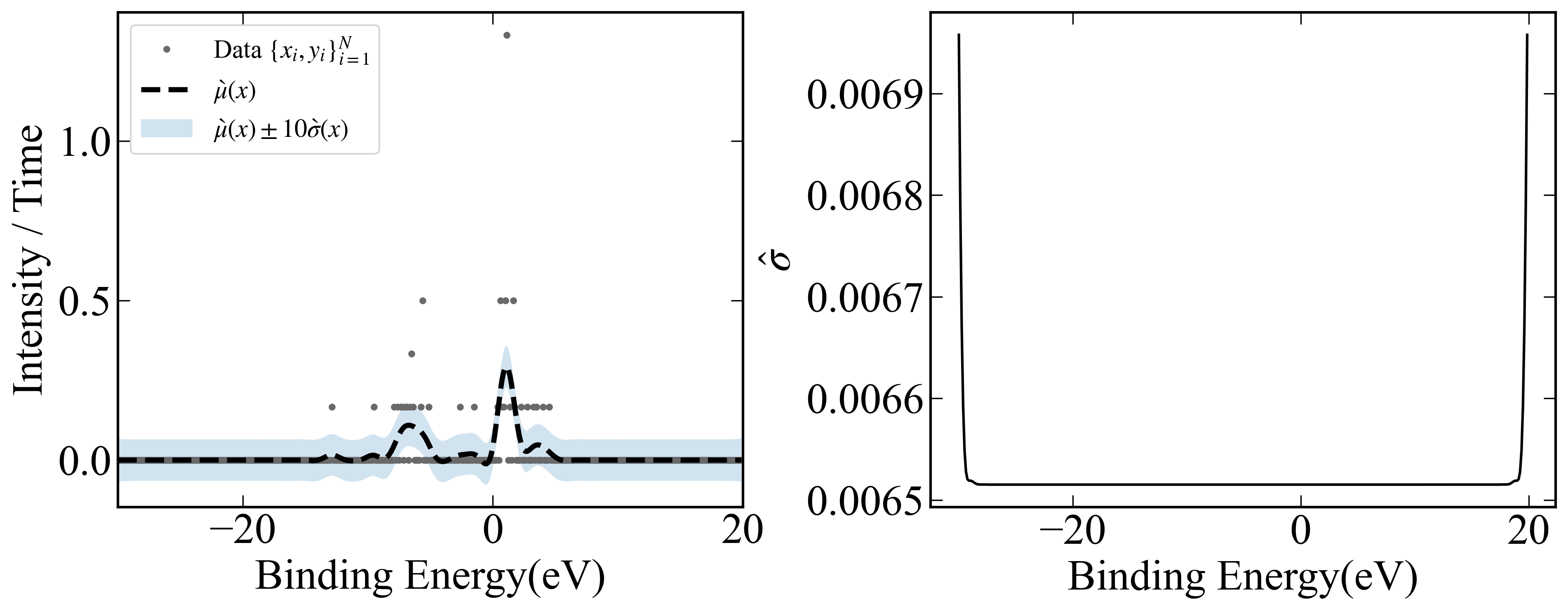}
    \caption{Example of the Gaussian process regression applied to the Bayesian Hamiltonian selection. The left panel shows the estimation results by the Gaussian process regression, and the right panel shows the width of the confidence intervals. The width of the confidence intervals is large at both ends.}
    \label{Estimation_Gauss_Hamiltonian}
  \end{figure}

  In the same problem setting as in Sect. \ref{section4}, we performed a sequential experimental design by active learning using the Gaussian process regression.
  The posterior mean $(\grave{\mu})$ and the variance $(\grave{\sigma})$ for the data observed at all measurement points with a measurement time of $T=6$ are shown in Fig. \ref{Estimation_Gauss_Hamiltonian}.
  The maximum values are taken at both ends, which are not important for the data analysis. This is because the noise is so large that the Gaussian process regression cannot be properly performed.
  As in Sect. \ref{section4.3}, let the prior distribution of the model set be $\varphi(M_2) = \varphi(M_3) = \frac{1}{2}$, the time for one measurement in the sequential experiment $(T)$ be $T=6$, the vertical resolution of the experiment be $0.125 (eV)$, 
  the number of data $(N)$ be $N=400$, and the candidate set of the measurement points $(\mathcal{X})$ be $\mathcal{X} = \{-30 + 0.125(i-1)\}_{i=1}^{400}$.
  Then, we repeat the experiment $k=160$ times by sequentially selecting $n=10$ points to be measured next 
so that the total measurement time is $T_{sum} = N\times T + n\times k \times T = 12,000$.  \par
\begin{figure}
    \centering
    \includegraphics*[width = 12.0cm]{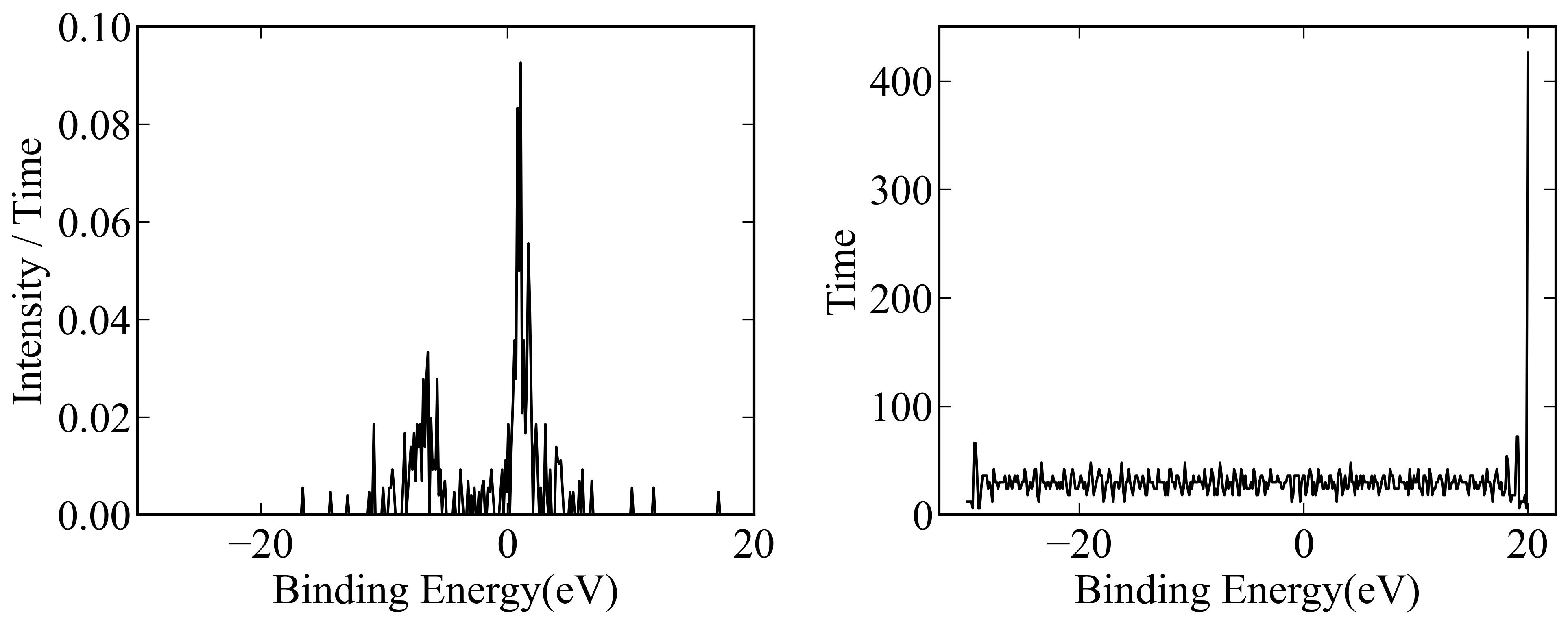}
    \caption{Data obtained in the sequential experiments using the Gaussian process regression. The left panel shows the number of observed photons per unit time $\{x_i,\bar{y}_i\}_{i=1}^N$.  The right panel shows the measurement time for each measurement point.  The measurement time is large at the right edge, which is not important for the Bayesian Hamiltonian selection.}
    \label{Data_Gauss_Hamiltonian}
\end{figure}

\begin{figure}[h]
    \centering
    \includegraphics*[width = 6.0cm]{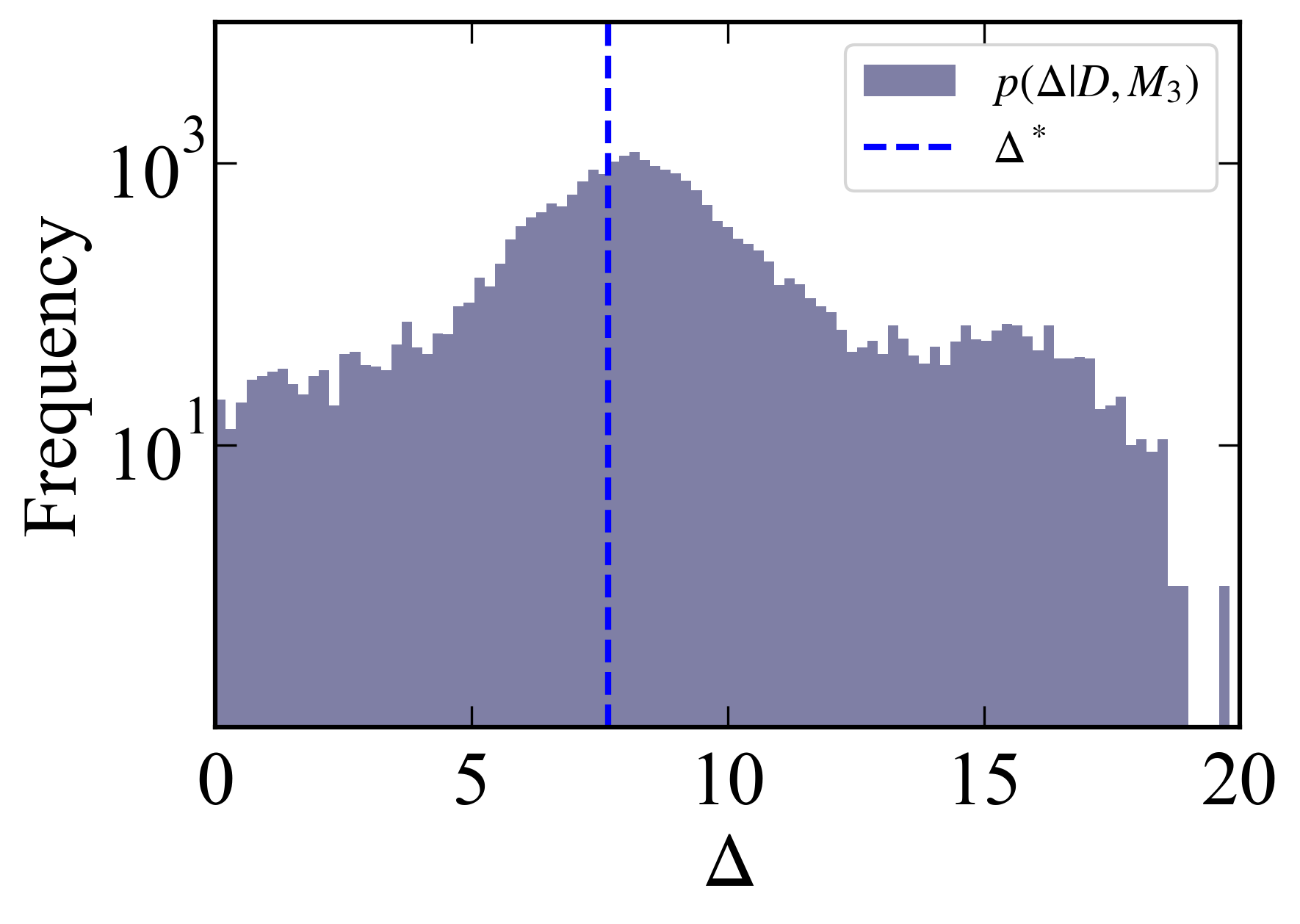}
    \includegraphics*[width = 6.0cm]{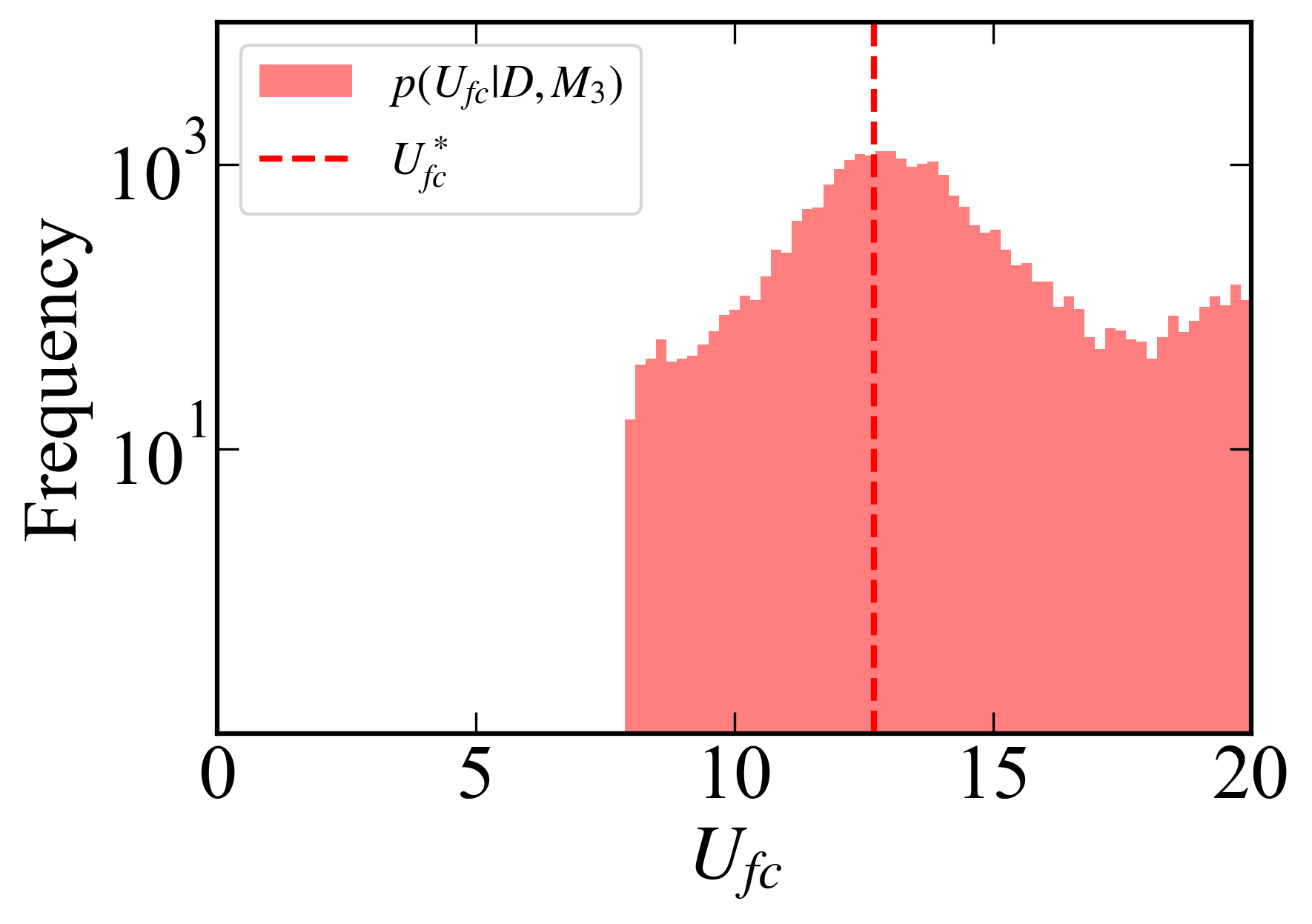}
    \caption{Parameter estimation by the posterior distribution $(p(\Delta|D,M_3), p(U_{fc}|D,M_3))$. The dashed line shows the true parameter values. The left and right panels show the prior distribution of $\Delta$ and $U_{fc}$, respectively.}
    \label{Data_Gauss_parameter_Hamiltonian}
\end{figure}

 
The obtained data are shown in Fig. \ref{Data_Gauss_Hamiltonian}. The measurements are focused on the range without peaks, which is not important in the Hamiltonian selection.
Furthermore, the parameter estimation of $\Delta,U_{fc}$ by the posterior distribution for this data is as shown in Fig. \ref{Data_Gauss_parameter_Hamiltonian}. 
The estimation accuracy of the proposed method is better than that shown in Fig. \ref{parameter_Hamiltonian_delta}.
  \section{Estimation results of the other parameters}
  \label{AppendixB}
  In this section, we show the result of parameter estimation in Sections \ref{section3.3} and \ref{section4.3} for the other parameters.
  \subsection{Bayesian spectrum deconvolution}
\begin{figure}[h]
    \centering
    \includegraphics*[width = 5.0cm]{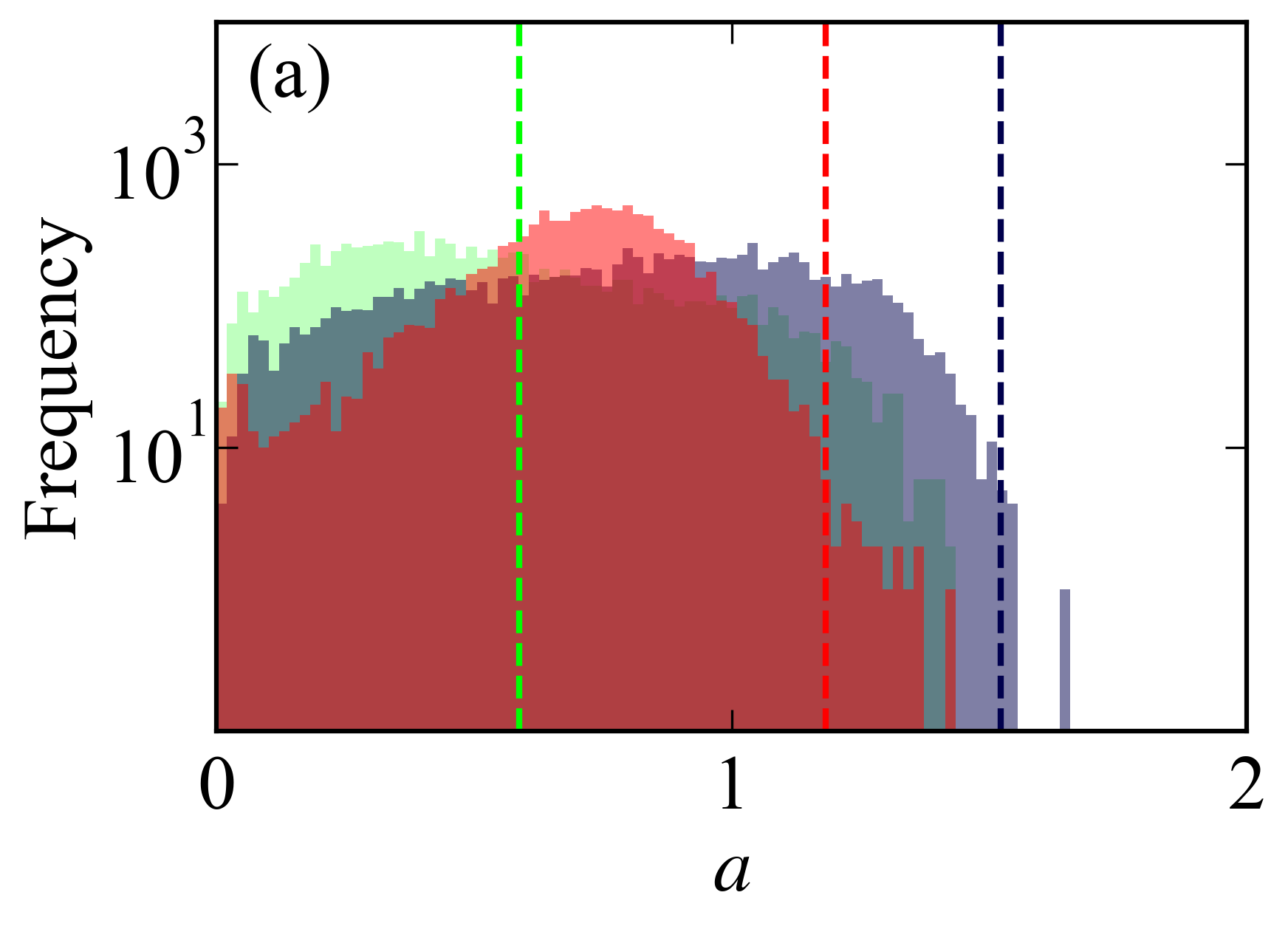}
    \includegraphics*[width = 5.0cm]{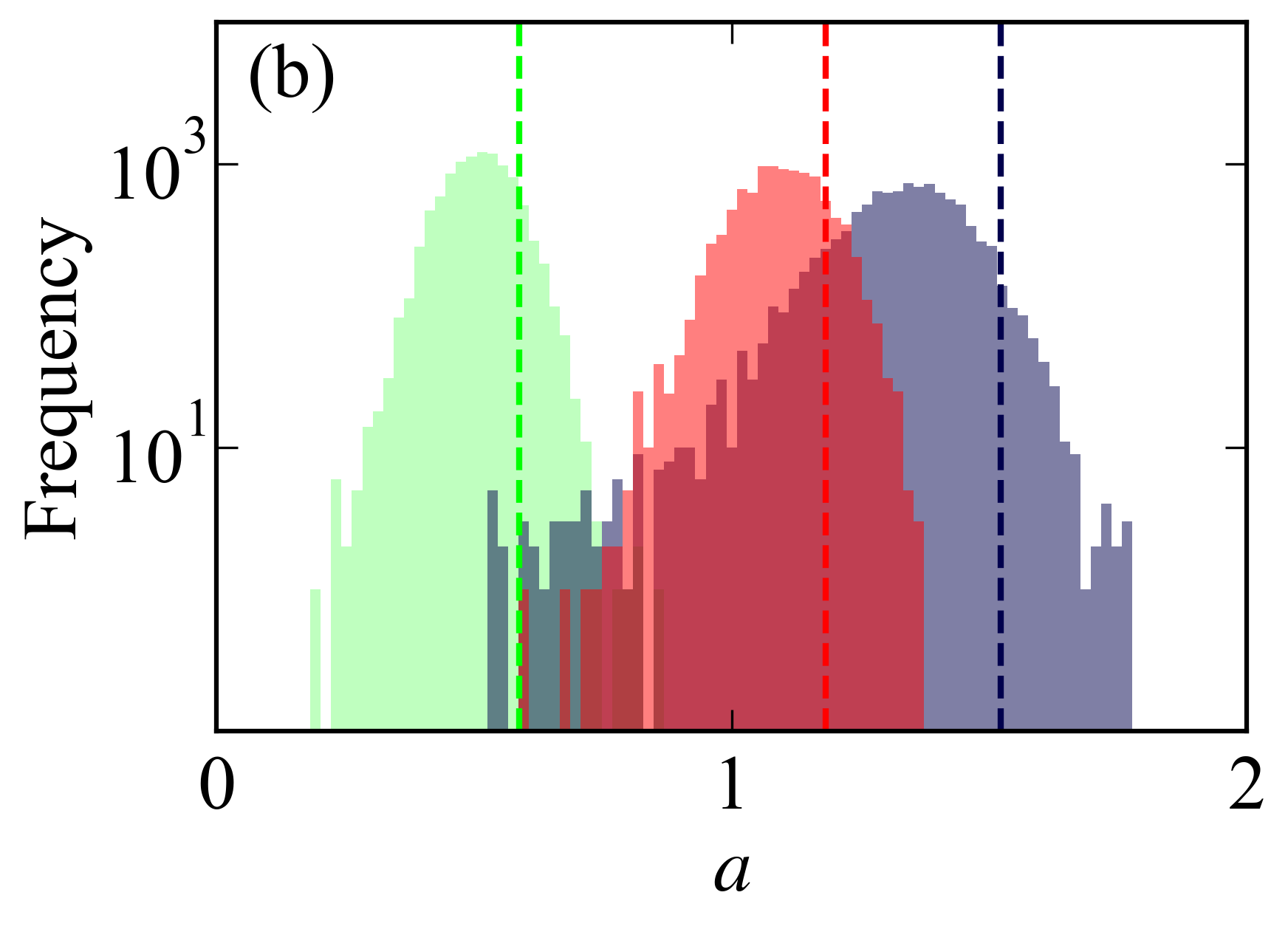}
    \includegraphics*[width = 5.0cm]{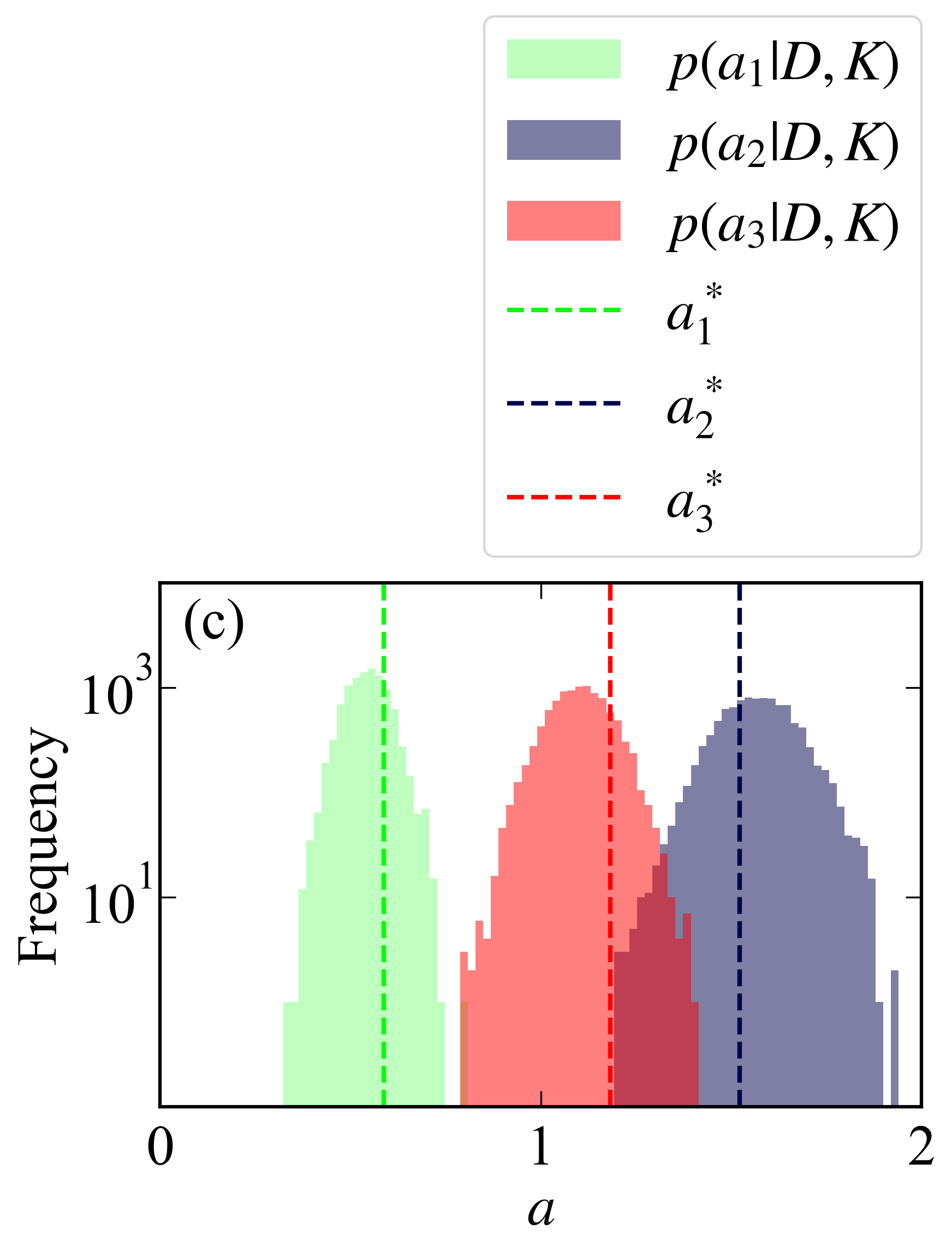}
    \caption{Parameter estimation by the posterior distribution $(p(a_1,a_2,a_3|D,M_3))$ of peak intensities $a_1,a_2$, and $a_3$. The dashed line shows the true parameter values. Panel (a) shows the case of a static experiment with a total measurement time of $T_{sum} = 2000$. Panel (b) shows the case of a sequential experiment with a total measurement time of $T_{sum} = 2000$. Panel (c) shows the case of a static experiment with a total measurement time of $T_{sum} = 6000$.}
    \label{parameter_deconvolution_a}
  \end{figure}

\begin{figure}[h]
    \centering
    \includegraphics*[width = 5.0cm]{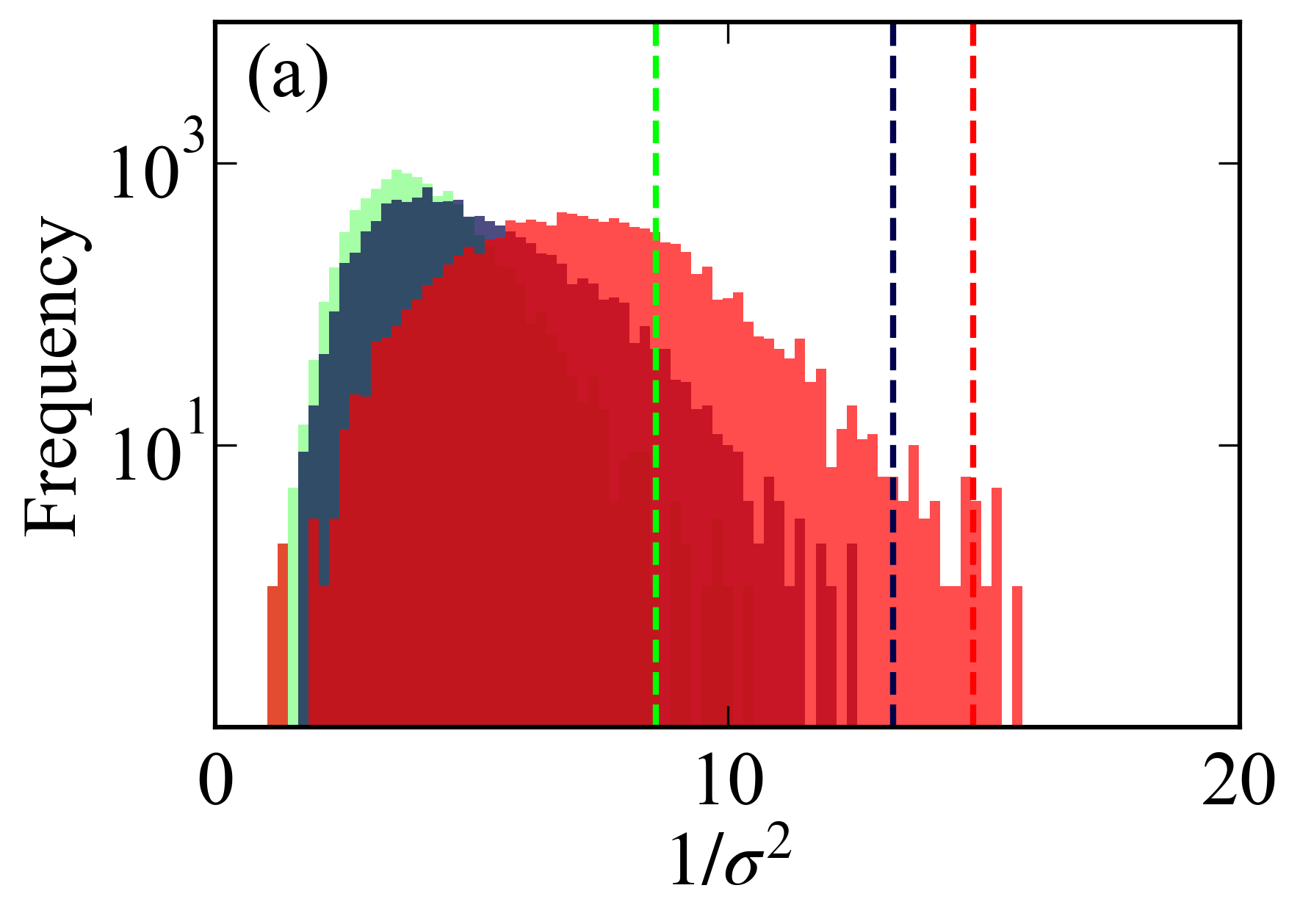}
    \includegraphics*[width = 5.0cm]{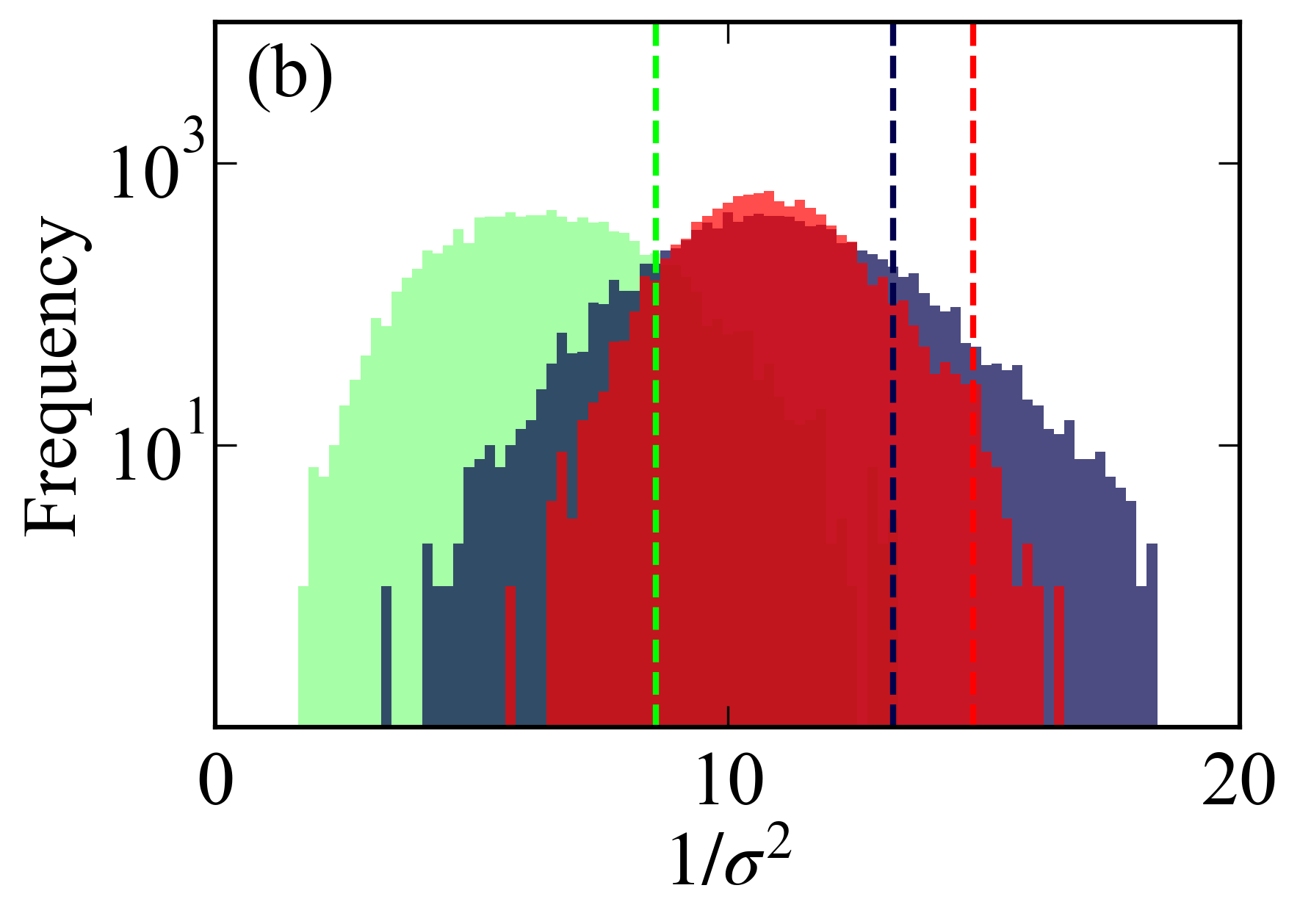}
    \includegraphics*[width = 5.0cm]{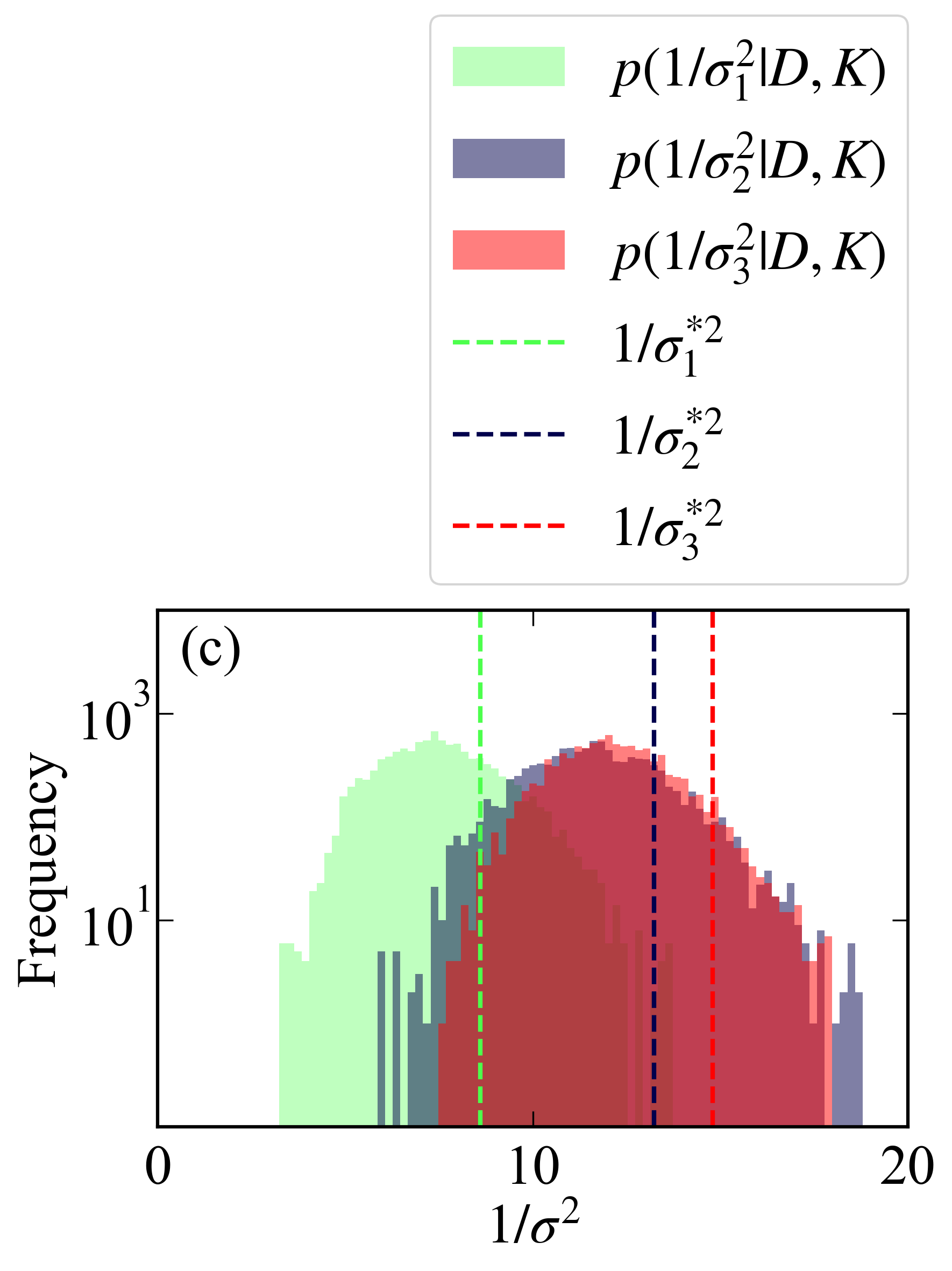}
    \caption{Parameter estimation by the posterior distribution $(p(\sigma_1,\sigma_2,\sigma_3|D,M_3))$ of peak variances $\sigma_1,\sigma_2$, and $\sigma_3$. The dashed line shows the true parameter values. Panel (a) shows the case of a static experiment with a total measurement time of $T_{sum} = 2000$. Panel (b) shows the case of a sequential experiment with a total measurement time of $T_{sum} = 2000$. Panel (c) shows the case of a static experiment with a total measurement time of $T_{sum} = 6000$.}
    \label{parameter_deconvolution_sigma}
  \end{figure}

\begin{figure}[h]
    \centering
    \includegraphics*[width = 5.0cm]{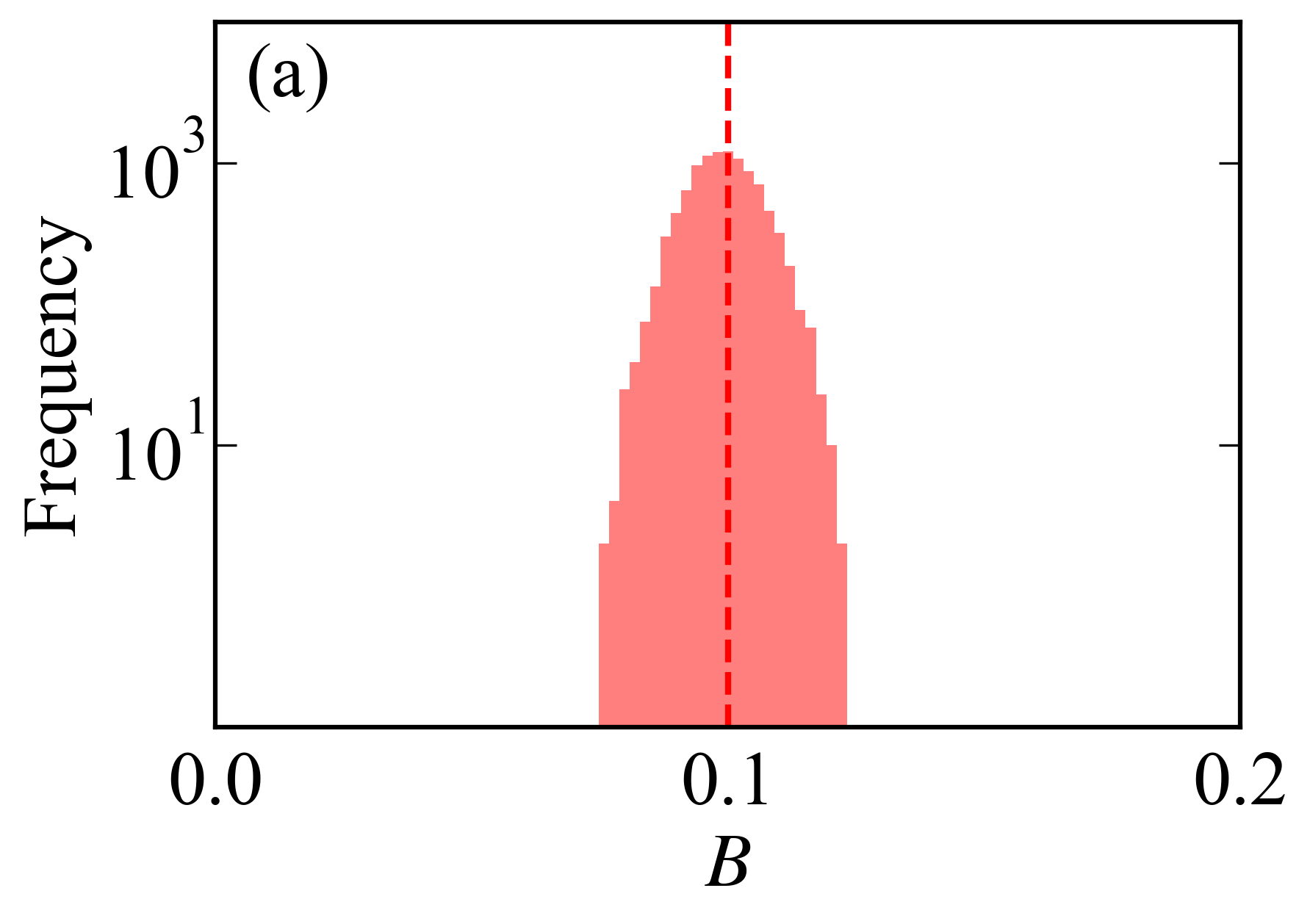}
    \includegraphics*[width = 5.0cm]{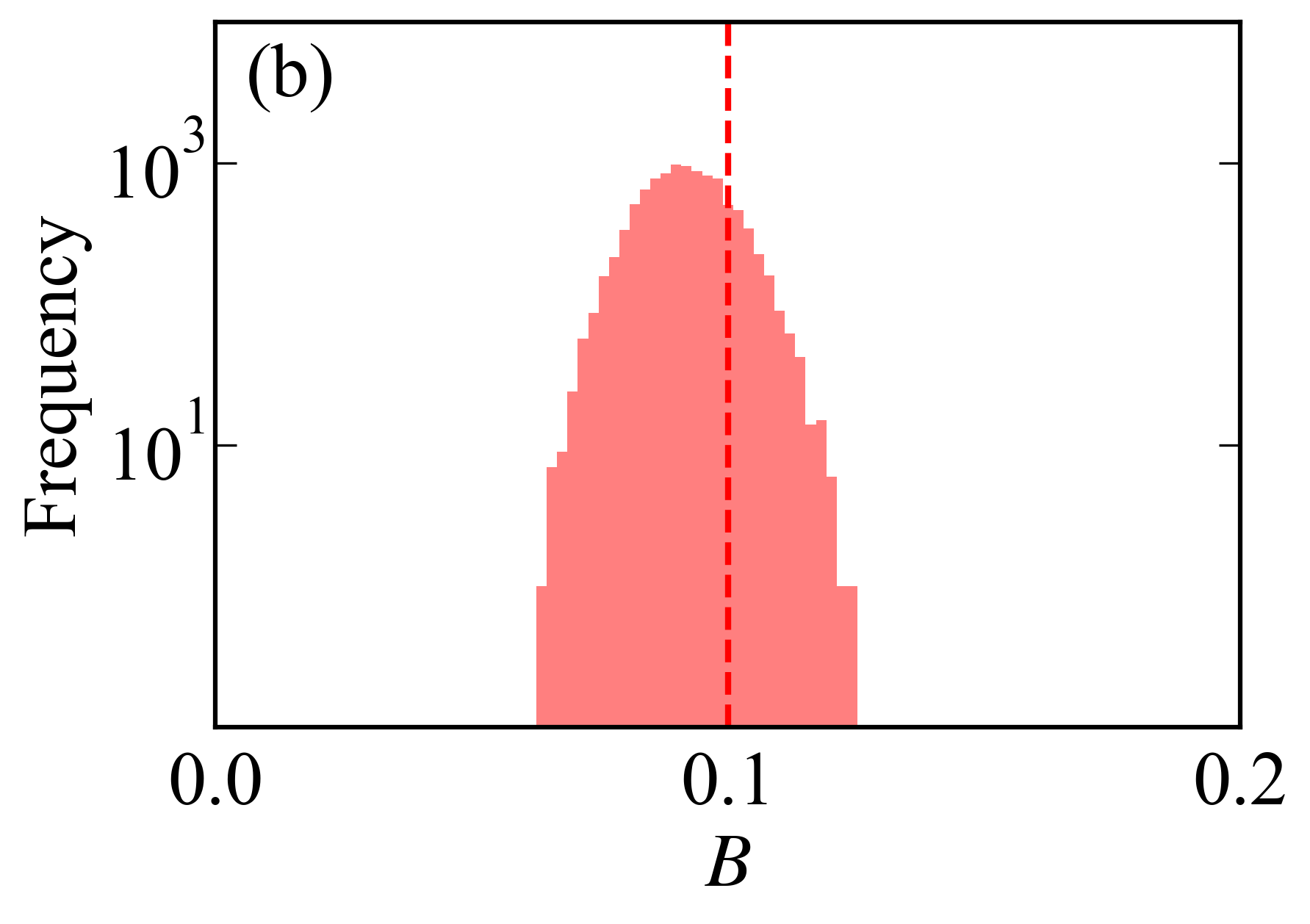}
    \includegraphics*[width = 5.0cm]{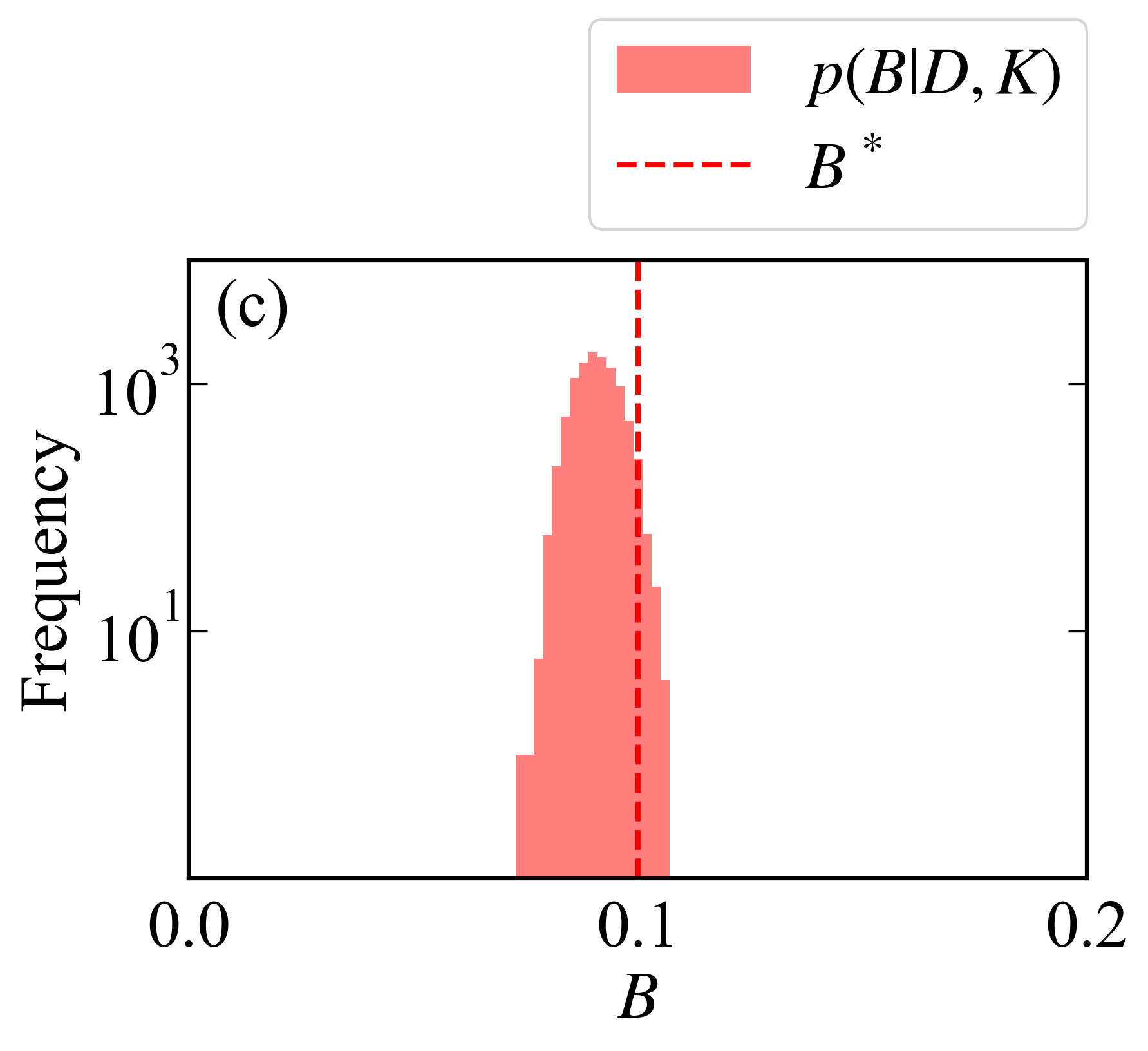}
    \caption{Parameter estimation by the posterior distribution $(p(B|D,M_3))$ of background intensity $B$. The dashed line shows the true parameter values. Panel (a) shows the case of a static experiment with a total measurement time of $T_{sum} = 2000$. Panel (b) shows the case of a sequential experiment with the total measurement time of $T_{sum} = 2000$. Panel (c) shows the case of a static experiment with a total measurement time of $T_{sum} = 6000$.}
    \label{parameter_deconvolution_B}
  \end{figure}

  The prior distribution of parameters $\{a_1,a_2,a_3\}, \{\sigma_1,\sigma_2,\sigma_3\}$, and $B$ are shown in Figs. \ref{parameter_deconvolution_a}, \ref{parameter_deconvolution_sigma}, and \ref{parameter_deconvolution_B}, respectively.
  \par
\begin{figure}[h]
    \centering
    \includegraphics*[width = 15.0cm]{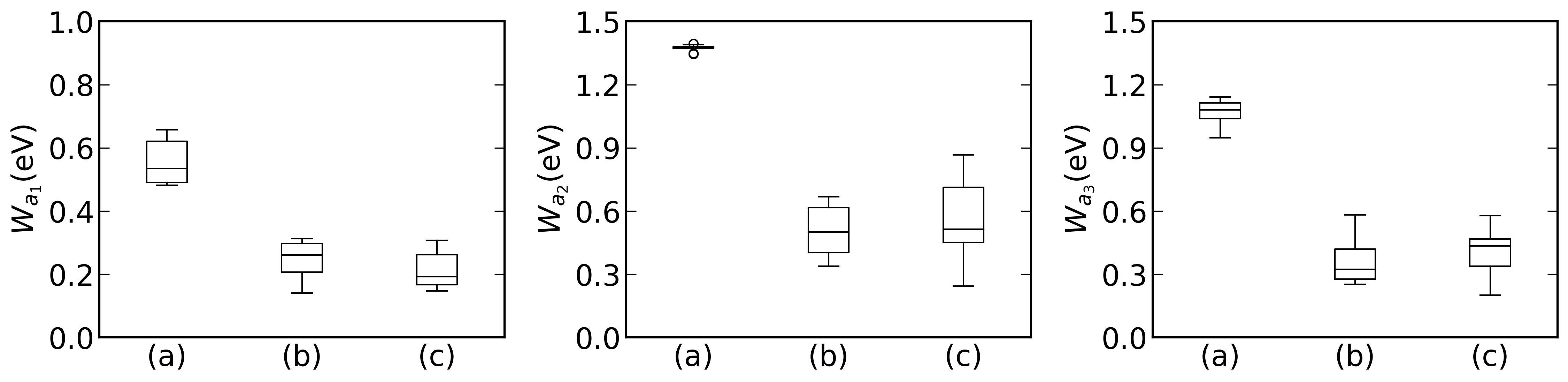}
    \caption{Boxplots representing the parameter estimation accuracy of the peak intensities. The left, middle, and right panel show the boxplots of $W_{a_1},W_{a_2}$, and $W_{a_3}$, respectively. Label (a) highlights the case of a static experiment with a total measurement time of $T_{sum} = 2000$. Label (b) highlights the case of a sequential experiment with a total measurement time of $T_{sum} = 2000$. Label (c) highlights the case of a static experiment with the total measurement time of $T_{sum} = 6000$.}
    \label{Width_deconvolution_a}
  \end{figure}
\begin{figure}[h]
    \centering
    \includegraphics*[width = 15.0cm]{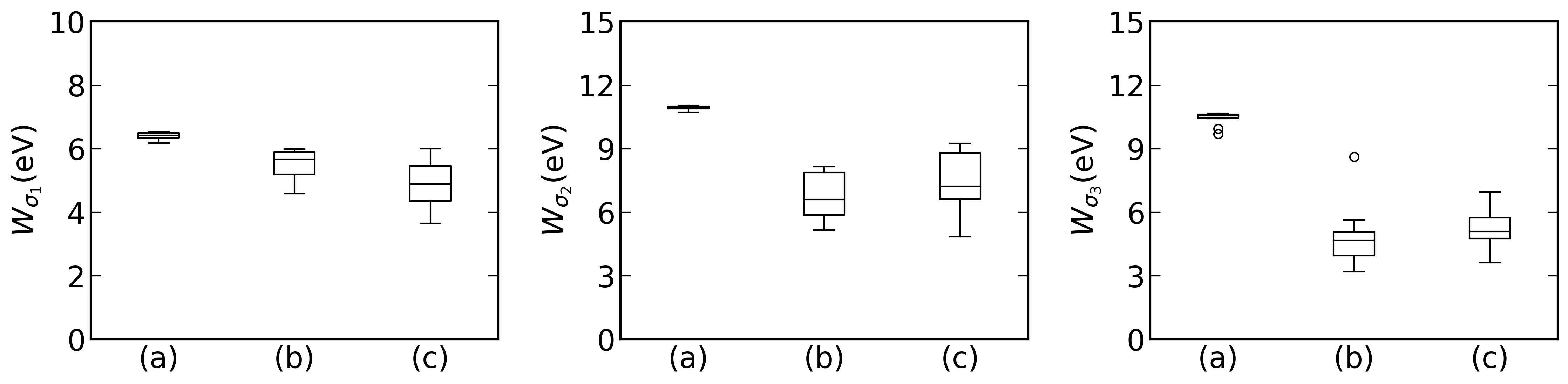}
    \caption{Boxplots representing the parameter estimation accuracy of the peak variances. The left, middle, and right panels show the boxplots of $W_{\sigma_1},W_{\sigma_2}$, and $W_{\sigma_3}$, respectively. Label (a) highlights the case of a static experiment with a total measurement time of $T_{sum} = 2000$. Label (b) highlights the case of a sequential experiment with a total measurement time of $T_{sum} = 2000$. Label (c) highlights the case of a static experiment with the total measurement time of $T_{sum} = 6000$.}
    \label{Width_deconvolution_sigma}
  \end{figure}
\begin{figure}[h]
    \centering
    \includegraphics*[width = 5.0cm]{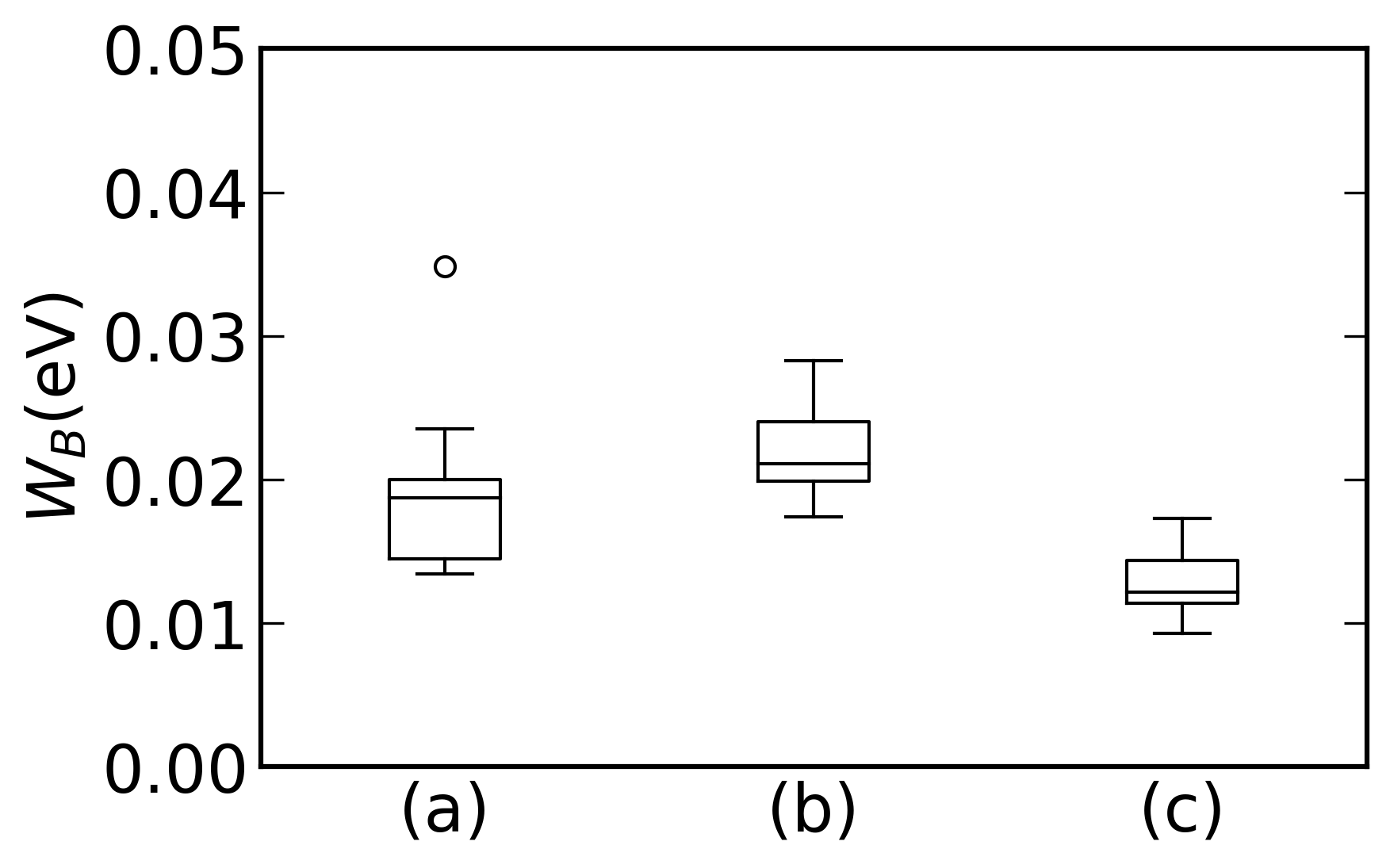}
    \caption{Boxplots representing the parameter estimation accuracy of the background. Label (a) highlights the case of a static experiment with a total measurement time of $T_{sum} = 2000$. Label (b) highlights the case of a sequential experiment with a total measurement time of $T_{sum} = 2000$. Label (c) highlights the case of a static experiment with a total measurement time $T_{sum} = 6000$.}
    \label{Width_deconvolution_B}
  \end{figure}

  To confirm the statistical properties, we define indices to evaluate the width of the parameter estimation, as follows:
  \begin{align}
    W_{a_1} = \max_{\alpha \in [0.025,0.975]}|a_1^* - a_{1,\alpha}|, W_{a_2} = \max_{\alpha \in [0.025,0.975]}|a_2^* - a_{2,\alpha}|,W_{a_3} = \max_{\alpha \in [0.025,0.975]}|a_3^* - a_{3,\alpha}|,
  \end{align}
  where 
  \begin{align}
    a_{1,\alpha} &= \min_{a} \left\{ \left(\int_{a_1 < a}p(a_1|D,K)\textup{d}a_1\right) > \alpha \right\}, \\
    a_{2,\alpha} &= \min_{a} \left\{ \left(\int_{a_2 < a}p(a_2|D,K)\textup{d}a_2\right) > \alpha \right\}, \\
    a_{3,\alpha} &= \min_{a} \left\{ \left(\int_{a_3 < a}p(a_3|D,K)\textup{d}a_3\right) > \alpha \right\}. \\
   \end{align}
   \begin{align}
    W_{\sigma_1} = \max_{\alpha \in [0.025,0.975]}|1/\sigma_1^{*2} - 1/\sigma_{1,\alpha}^2|, W_{\sigma_2} = \max_{\alpha \in [0.025,0.975]}|1/\sigma_2^{*2} - 1/\sigma_{2,\alpha}^2|,W_{\sigma_3} = \max_{\alpha \in [0.025,0.975]}|1/\sigma_3^{*2} - 1/\sigma_{3,\alpha}^2|,
  \end{align}
  where 
  \begin{align}
    \sigma_{1,\alpha} &= \min_{\sigma} \left\{ \left(\int_{\sigma_1 < \sigma}p(\sigma_1|D,K)\textup{d}\sigma_1\right) > \alpha \right\}, \\
    \sigma_{2,\alpha} &= \min_{\sigma} \left\{ \left(\int_{\sigma_2 < \sigma}p(\sigma_2|D,K)\textup{d}\sigma_2\right) > \alpha \right\}, \\
    \sigma_{3,\alpha} &= \min_{\sigma} \left\{ \left(\int_{\sigma_3 < \sigma}p(\sigma_3|D,K)\textup{d}\sigma_3\right) > \alpha \right\}. \\
   \end{align}
   \begin{align}
    W_{B} = \max_{\alpha \in [0.025,0.975]}|B^* - B_{\alpha}|
  \end{align}
  where 
  \begin{align}
    B_{\alpha} &= \min_{B'} \left\{ \left(\int_{B < B'}p(B|D,K)\textup{d}B\right) > \alpha \right\}. \\
   \end{align}
   The boxplots of $\{W_{a_1},W_{a_2},W_{a_3}\}$, $\{W_{\sigma_1},W_{\sigma_2}$, $W_{\sigma_3}\}$, and $W_{B}$ are shown in Figs. \ref{Width_deconvolution_a}, \ref{Width_deconvolution_sigma}, and \ref{Width_deconvolution_B} respectively.
   \par
  From these figures, it can be shown that our method improves the parameter estimation of the peak intensities and peak variances.
  Furthermore, our method deteriorates the parameter estimation of the background intensity, which is not very relevant to the model selection.
  \subsection{Bayesian Hamiltonian selection}
\begin{figure}[h]
    \centering
    \includegraphics*[width = 5.0cm]{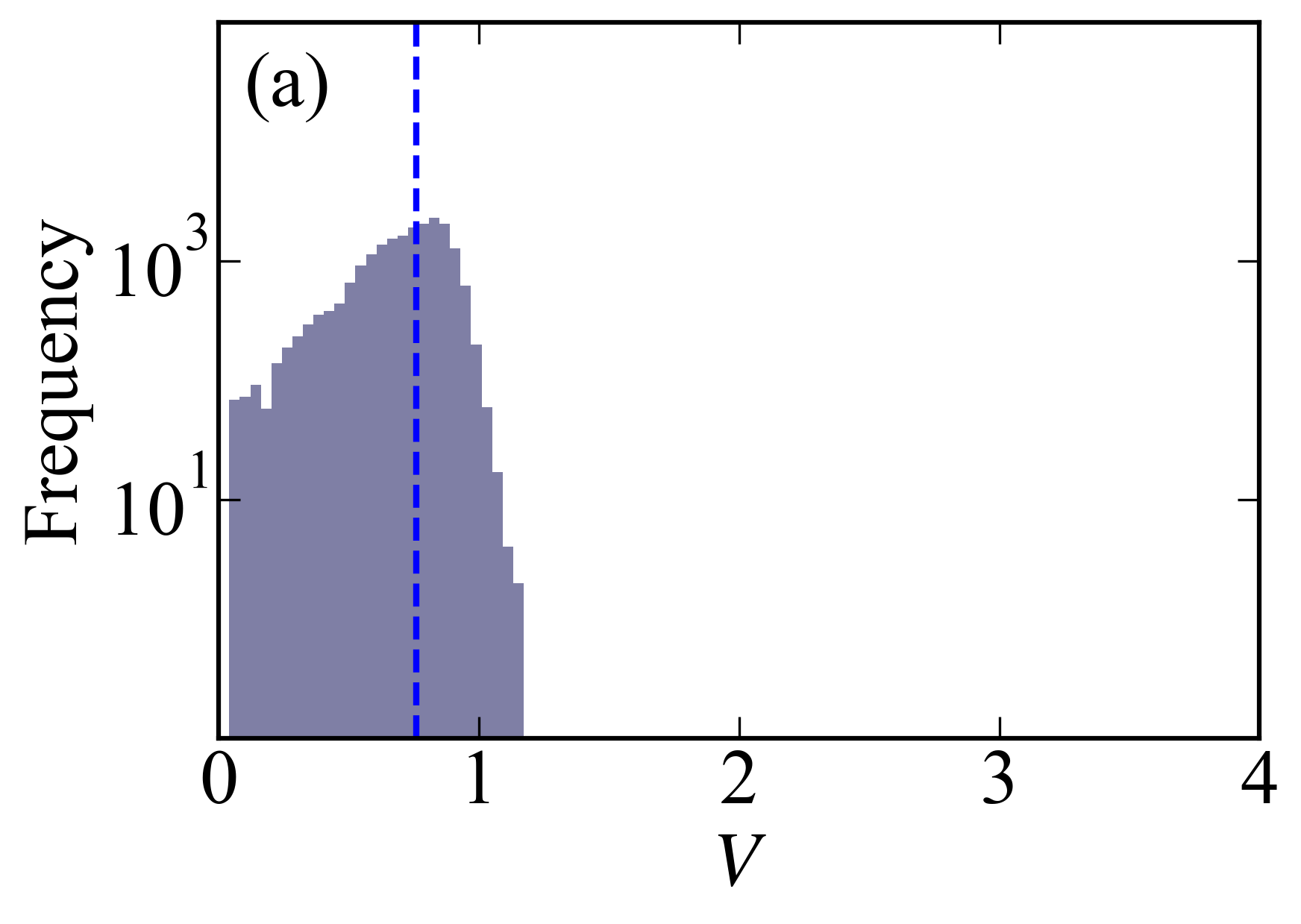}
    \includegraphics*[width = 5.0cm]{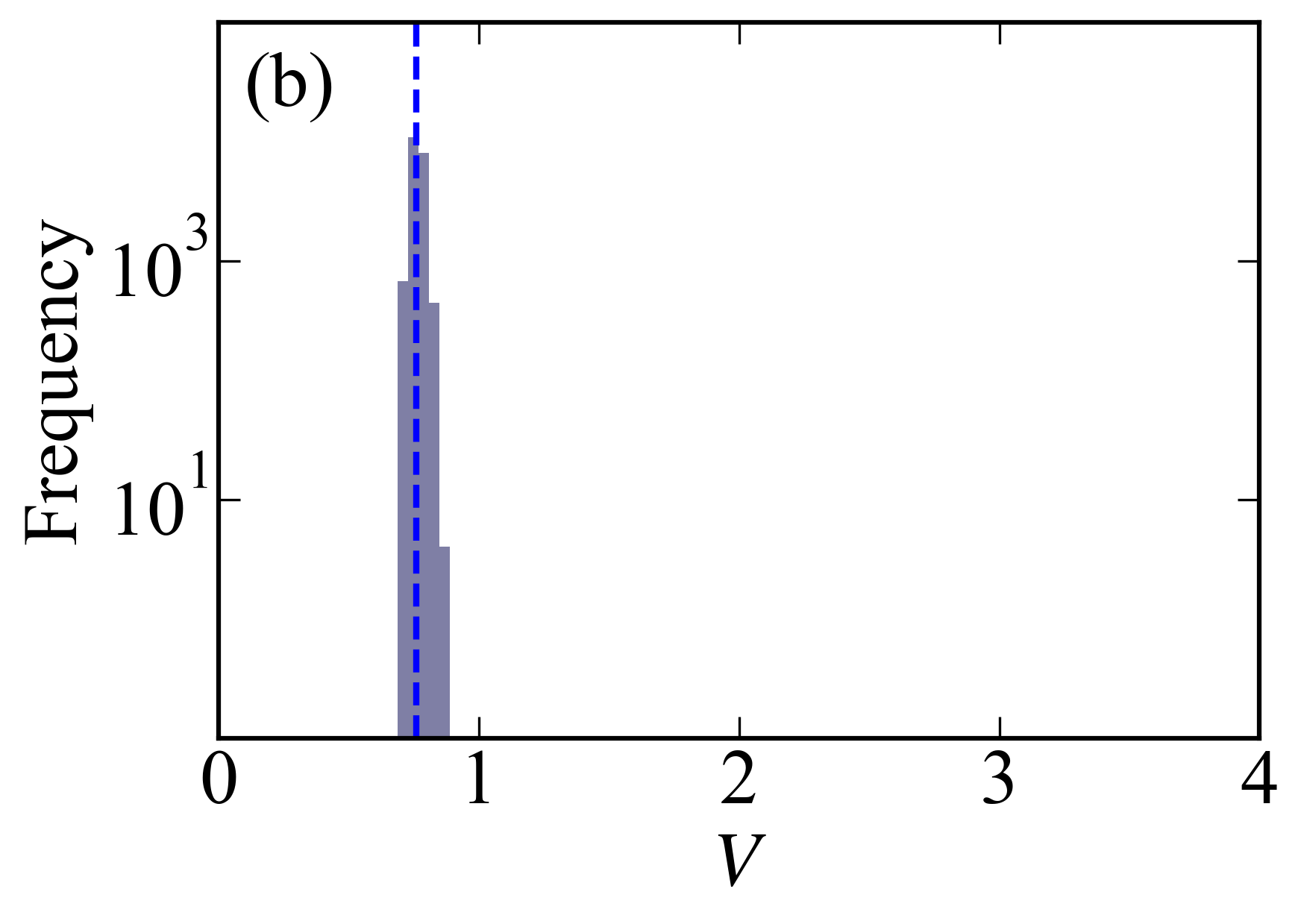}
    \includegraphics*[width = 5.0cm]{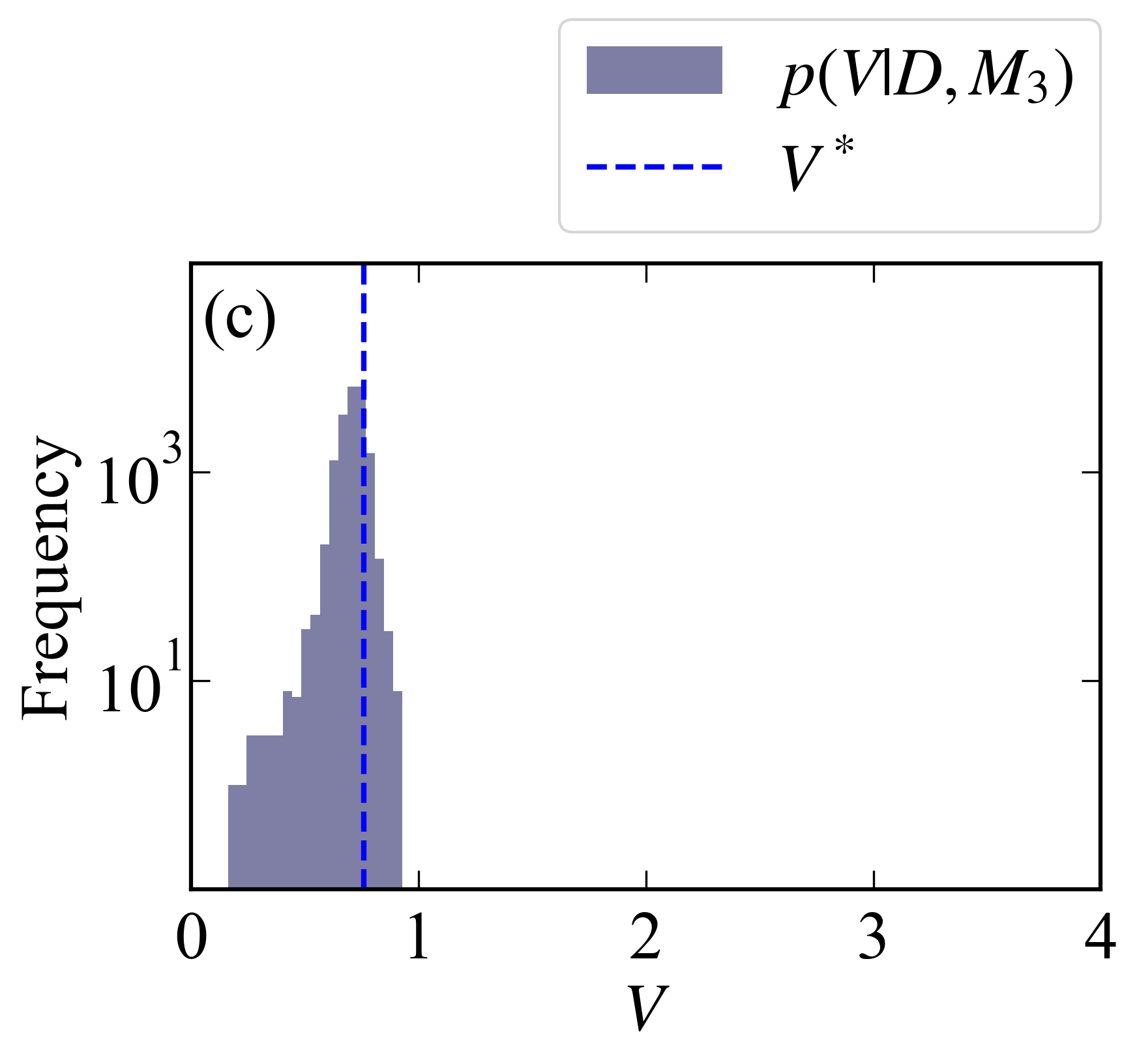}
    \caption{Parameter estimation by the posterior distribution $(p(V|D,M_3))$. The dashed line shows the true parameter values. Panel (a) shows the case of a static experiment with a total measurement time of $T_{sum} = 12,000$. Panel (b) shows the case of a sequential experiment with the total measurement time $T_{sum} = 12,000$. Panel (c) shows the case of a static experiment with a total measurement time $T_{sum} = 36,000$.}
    \label{parameter_Hamiltonian_V}
  \end{figure}

\begin{figure}[h]
    \centering
    \includegraphics*[width = 5.0cm]{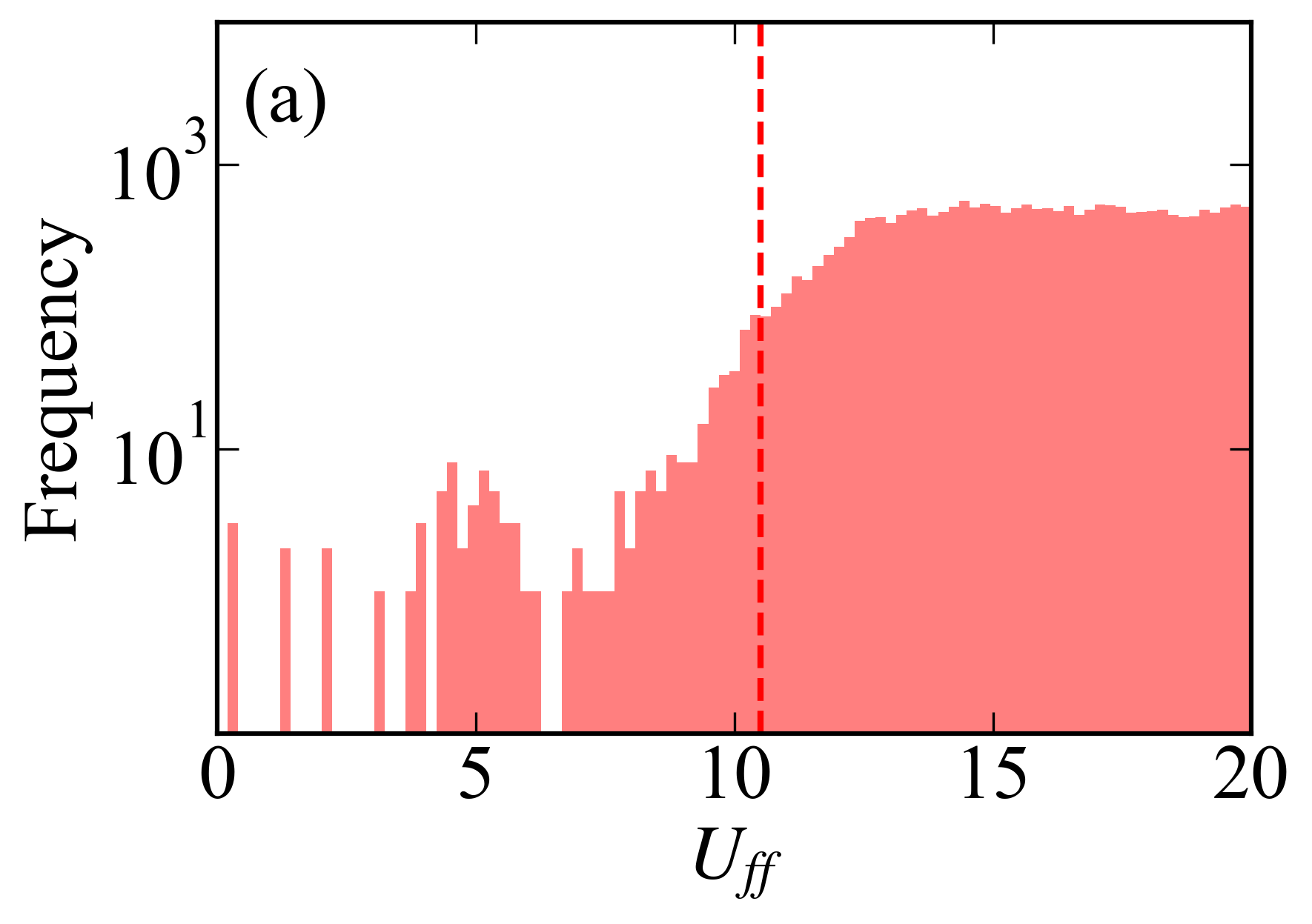}
    \includegraphics*[width = 5.0cm]{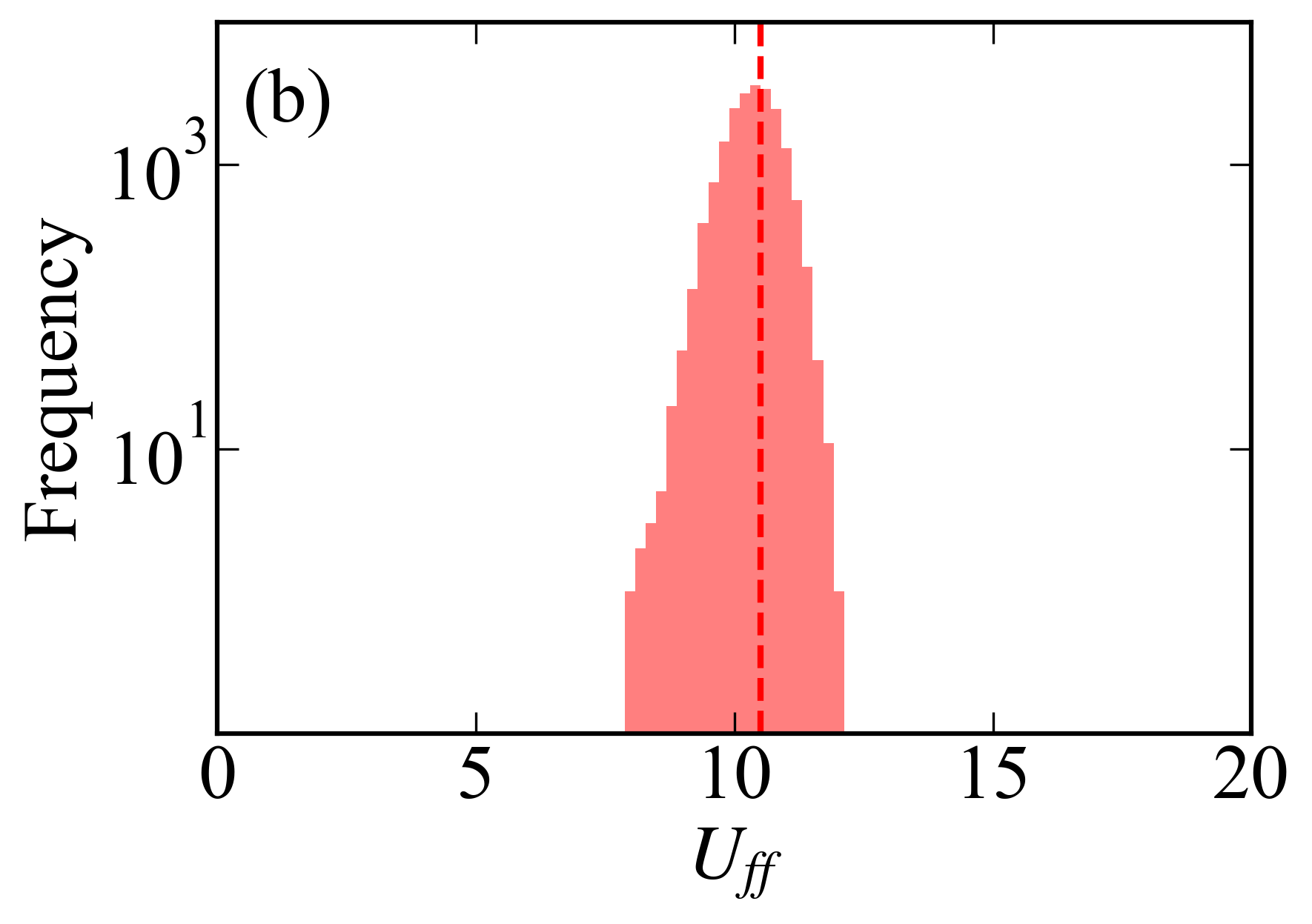}
    \includegraphics*[width = 5.0cm]{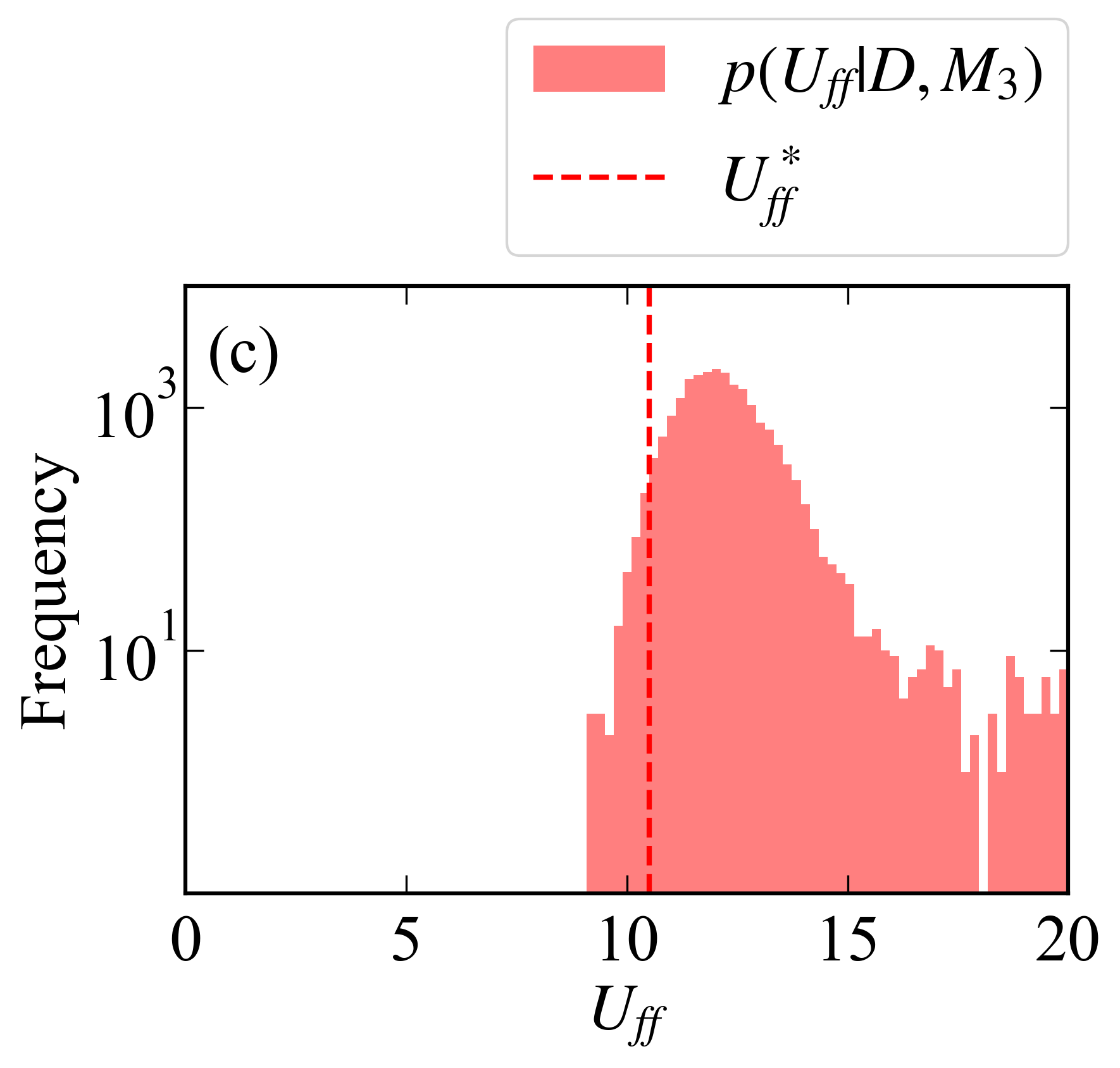}
    \caption{Parameter estimation by the posterior distribution $(p(U_{ff}|D,M_3))$. The dashed line shows the true parameter values. Panel (a) shows the case of a static experiment with a total measurement time of $T_{sum} = 12,000$. Panel (b) shows the case of a sequential experiment with a total measurement time of $T_{sum} = 12,000$. Panel (c) shows the case of a static experiment with a total measurement time of $T_{sum} = 36,000$.}
    \label{parameter_Hamiltonian_Uff}
  \end{figure}

\begin{figure}[h]
    \centering
    \includegraphics*[width = 5.0cm]{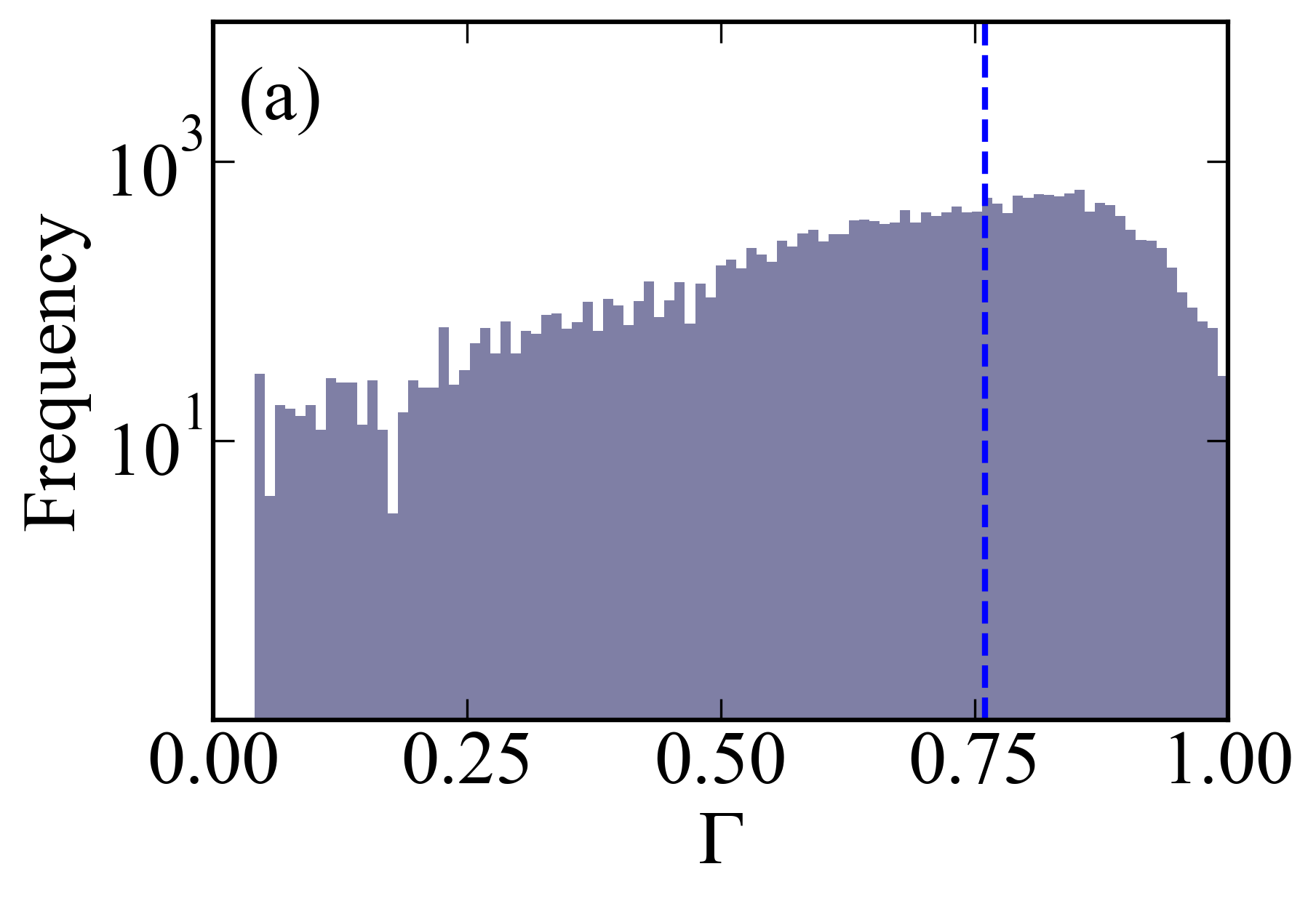}
    \includegraphics*[width = 5.0cm]{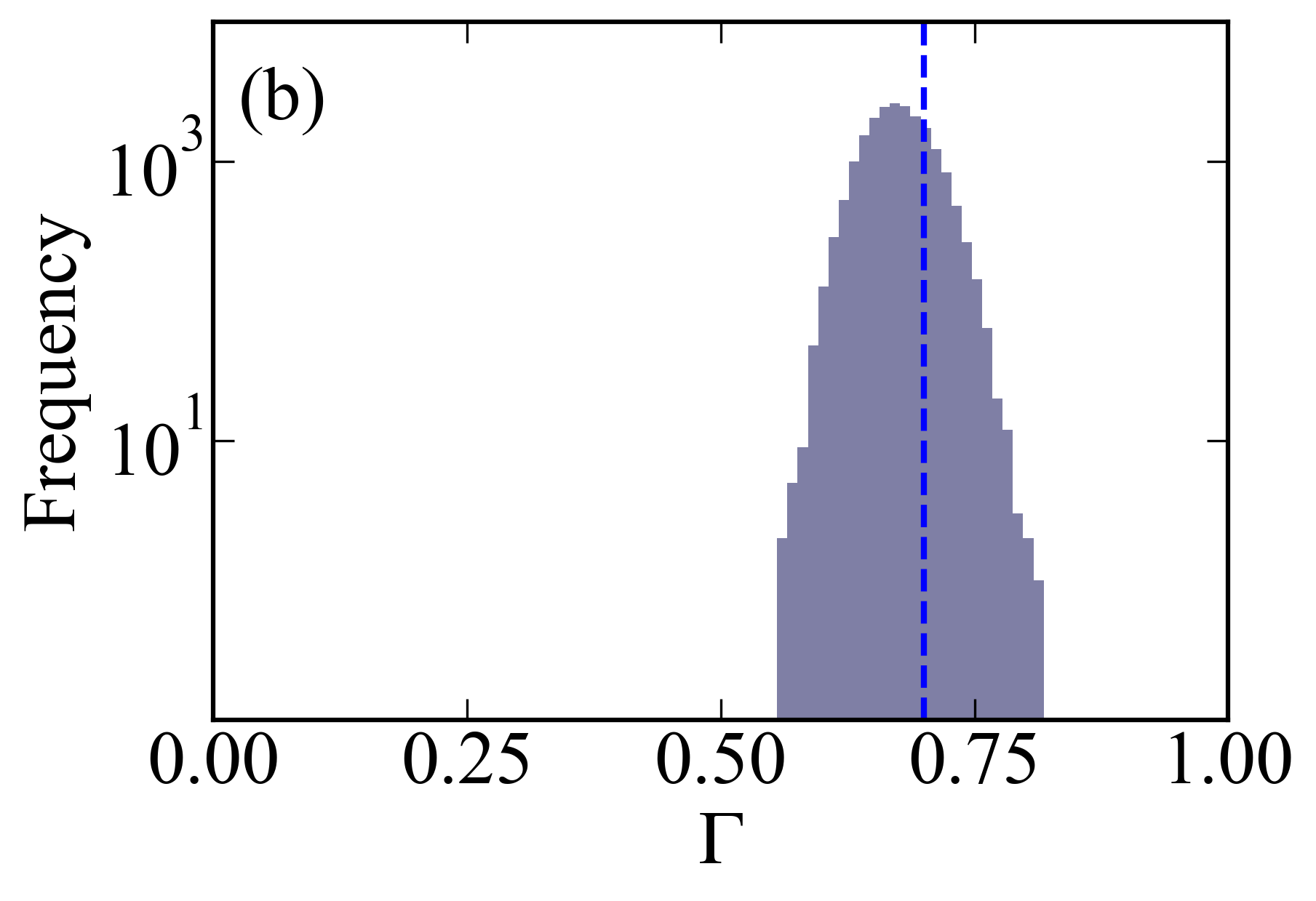}
    \includegraphics*[width = 5.0cm]{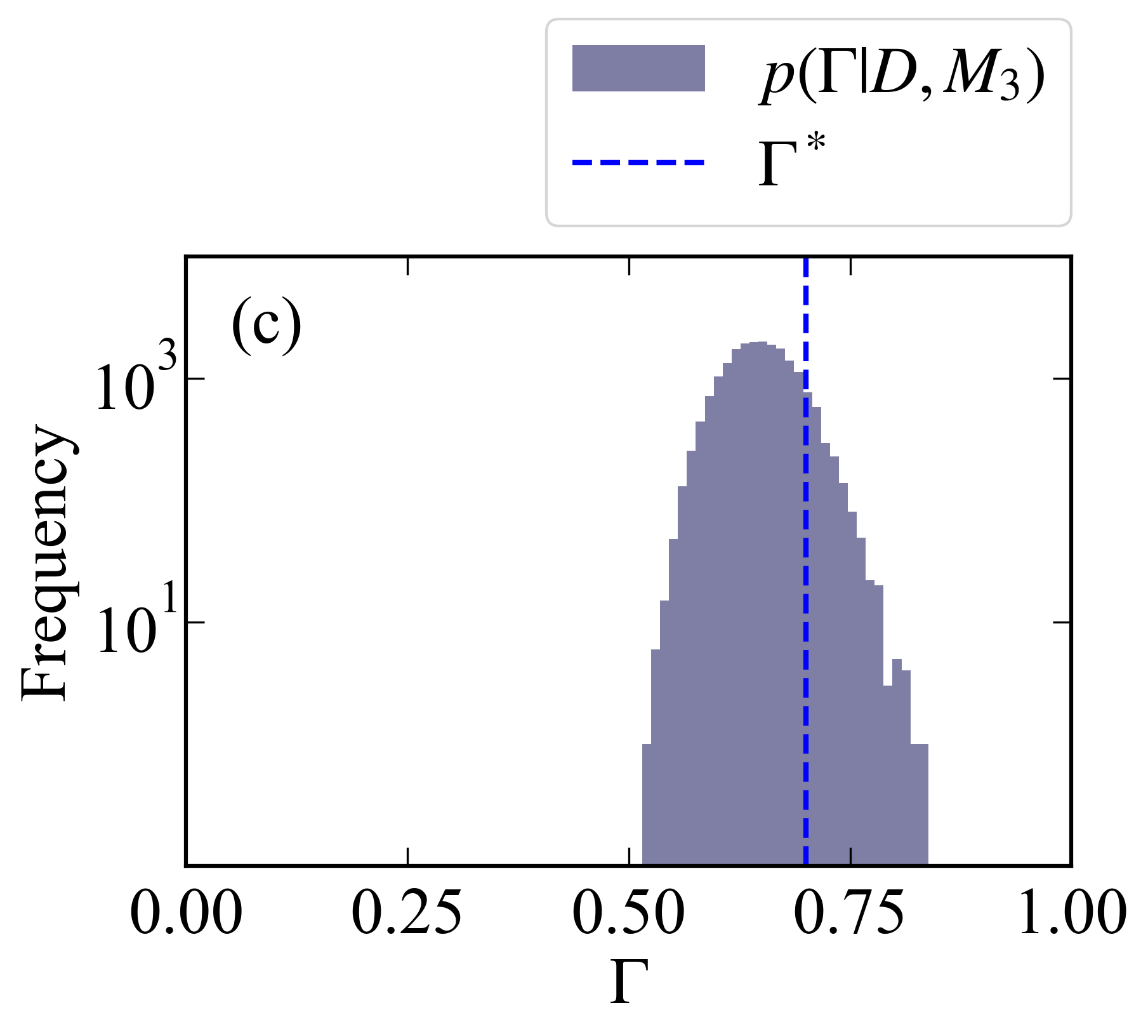}
    \caption{Parameter estimation by the posterior distribution $(p(\Gamma|D,M_3))$. The dashed line shows the true parameter values. Panel (a) shows the case of a static experiment with a total measurement time of $T_{sum} = 12,000$. Panel (b) shows the case of a sequential experiment with a total measurement time of $T_{sum} = 12000$. Panel (c) shows the case of a static experiment with a total measurement time $T_{sum} = 36,000$.}
    \label{parameter_Hamiltonian_Gamma}
  \end{figure}

\begin{figure}[h]
    \centering
    \includegraphics*[width = 5.0cm]{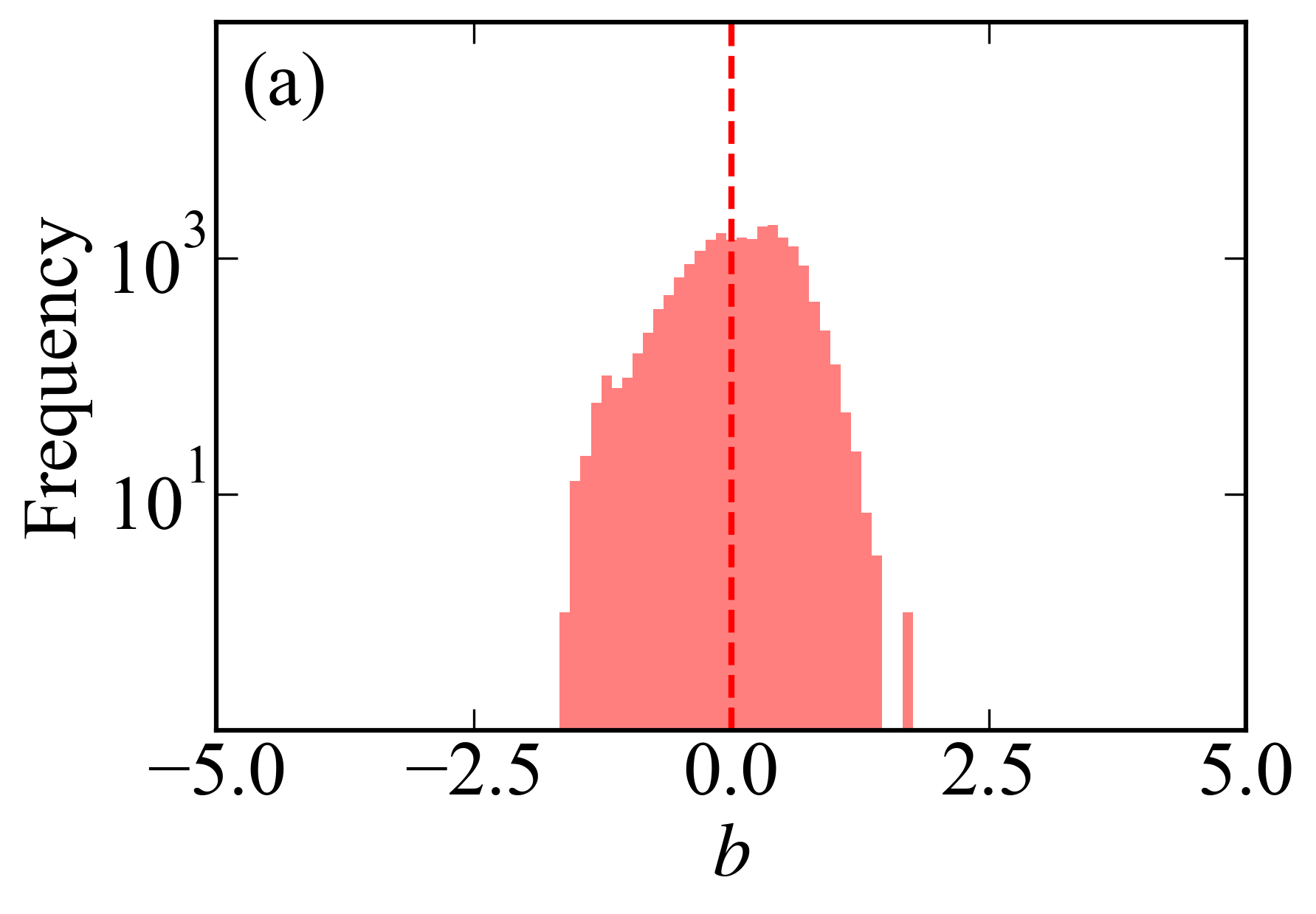}
    \includegraphics*[width = 5.0cm]{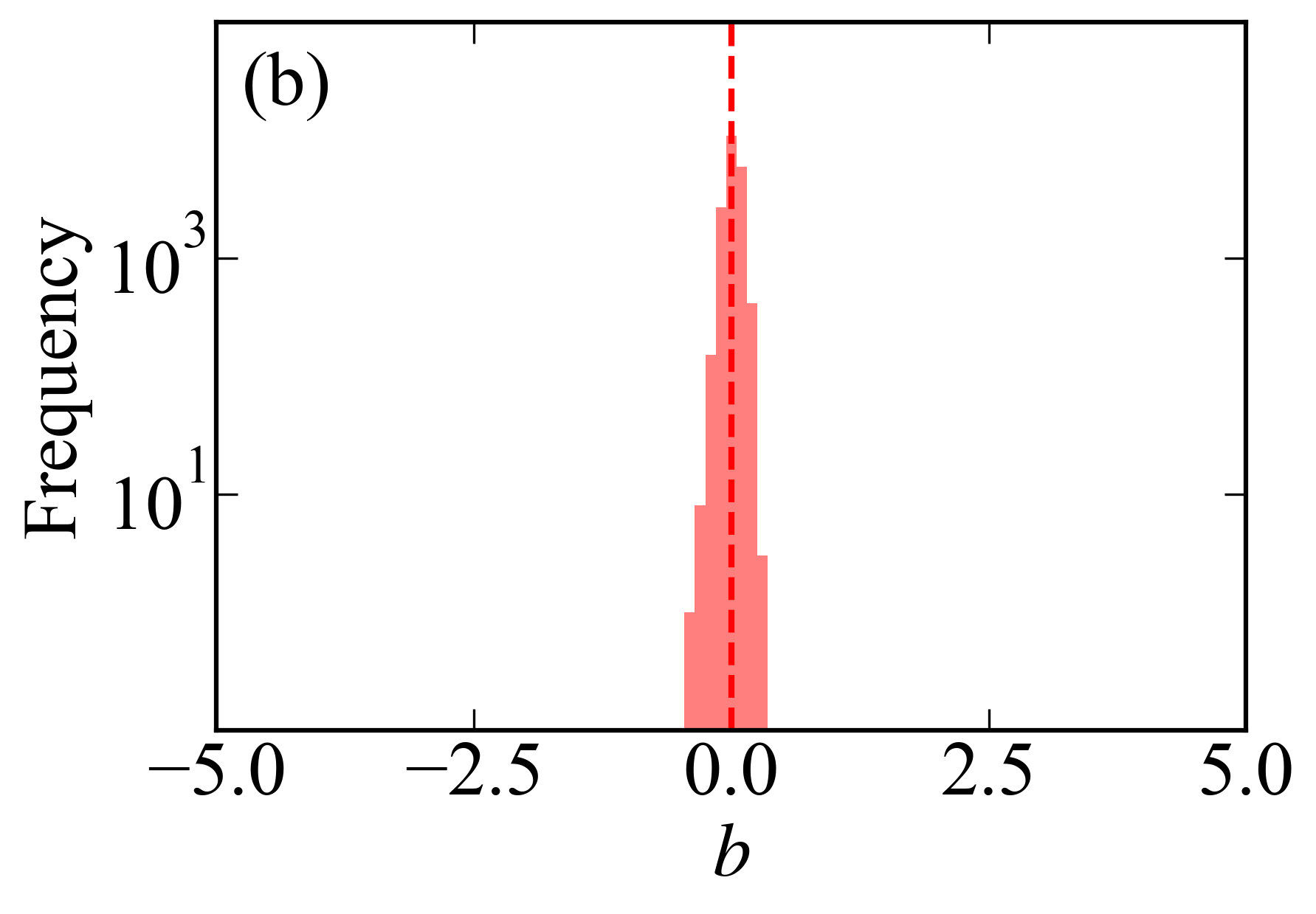}
    \includegraphics*[width = 5.0cm]{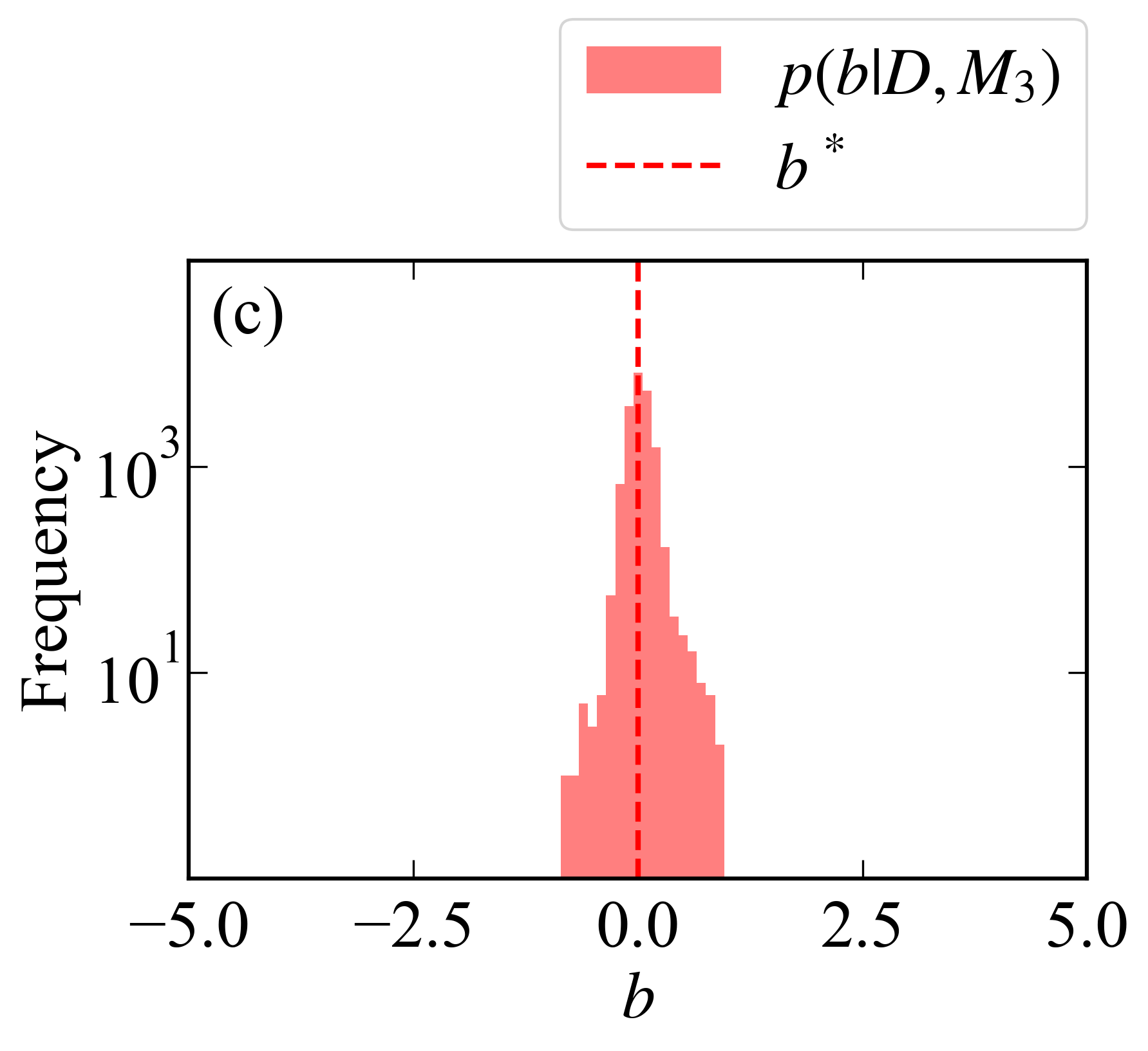}
    \caption{Parameter estimation by the posterior distribution $(p(b|D,M_3))$. The dashed line shows the true parameter values. Panel (a) shows the case of a static experiment with a total measurement time of $T_{sum} = 12,000$. Panel (b) shows the case of a sequential experiment with a total measurement time of $T_{sum} = 12,000$. Panel (c) shows the case of a static experiment with a total measurement time of $T_{sum} = 36,000$.}
    \label{parameter_Hamiltonian_b}
  \end{figure}

  The prior distribution of parameters $V, U_{ff}, \Gamma$, and $b$ are shown in Figs \ref{parameter_Hamiltonian_V}, \ref{parameter_Hamiltonian_Uff}, \ref{parameter_Hamiltonian_Gamma}, and \ref{parameter_Hamiltonian_b}, respectively.
  \par
\begin{figure}[h]
    \centering
    \includegraphics*[width = 17.0cm]{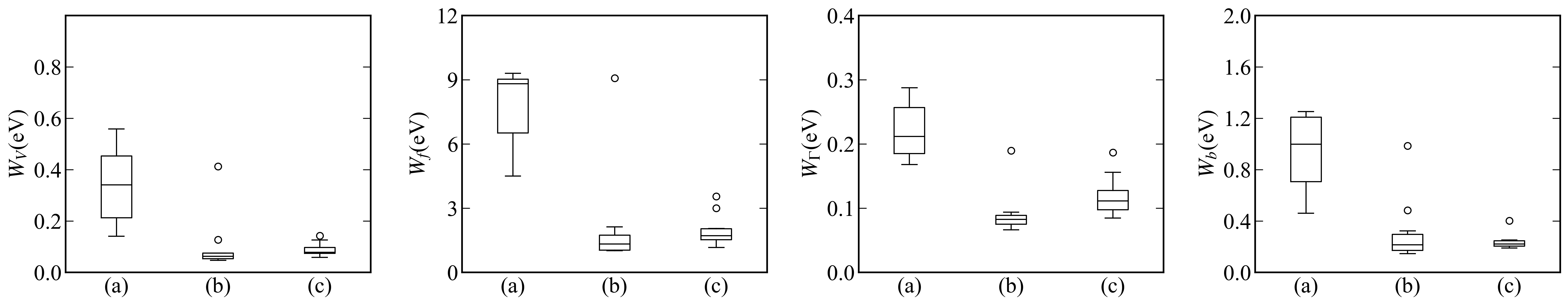}
    \caption{Boxplots representing the parameter estimation accuracy of the peak positions. Label (a) highlights the case of a static experiment with a total measurement time of $T_{sum} = 12,000$. Label (b) highlights the case of a sequential experiment with a total measurement time of $T_{sum} = 12,000$. Label (c) highlights the case of a static experiment with a total measurement time of $T_{sum} = 36,000$.}
    \label{Width_Hamiltonian_other}
  \end{figure}

  We define the indices to evaluate the width of the parameter estimation as follows:
  \begin{align}
    W_{V} = \max_{\alpha \in [0.025,0.975]}|V^* - V_{\alpha}|, W_{f} = \max_{\alpha \in [0.025,0.975]}|U_{f}^* - f_{\alpha}|, W_{\Gamma} = \max_{\alpha \in [0.025,0.975]}|\Gamma^* - \Gamma_{\alpha}|, W_{b} = \max_{\alpha \in [0.025,0.975]}|b^* - b_{\alpha}|,
  \end{align}
  where 
  \begin{align}
    V_{\alpha} &= \min_{V'} \left\{ \left(\int_{V < V'}p(V|D,K)\textup{d}V\right) > \alpha \right\}, \\
    f_{\alpha} &= \min_{U'} \left\{ \left(\int_{U_{ff} < U'}p(U_{ff}|D,K)\textup{d}U_{ff}\right) > \alpha \right\}, \\
    \Gamma_{\alpha} &= \min_{\Gamma'} \left\{ \left(\int_{\Gamma < \Gamma'}p(\Gamma|D,K)\textup{d}\mu_0\right) > \alpha \right\}, \\
    b_{\alpha} &= \min_{b'} \left\{ \left(\int_{b < b'}p(b|D,K)\textup{d}b\right) > \alpha \right\}. \\
  \end{align}
  The boxplots of $W_{V},W_{f},W_{\Gamma}$, and $W_{b}$ are shown in Fig. \ref{Width_Hamiltonian_other}.
  From these figures, it can be established that our method improves the estimation of the other parameters.
%

\end{document}